\documentclass{article}

\usepackage{microtype}
\usepackage{graphicx}
\usepackage{subfig}
\usepackage{booktabs} 

\usepackage{hyperref}

\usepackage[accepted]{icml2023}

\usepackage{amsmath}
\usepackage{amssymb}
\usepackage{mathtools}
\usepackage{amsthm}

\usepackage[capitalize,noabbrev]{cleveref}

\newcommand{\Px}[2]{\mathbb{P}_{#1} \left[ #2 \right]}

\theoremstyle{plain}
\newtheorem{theorem}{Theorem}[section]
\newtheorem{proposition}[theorem]{Proposition}

\theoremstyle{definition}

\theoremstyle{remark}

\usepackage{xspace}
\makeatletter
\DeclareRobustCommand\onedot{\futurelet\@let@token\@onedot}
\def\@onedot{\ifx\@let@token.\else.\null\fi\xspace}

\def\eg{\emph{e.g}\onedot} 
\def\ie{\emph{i.e}\onedot}

\makeatletter

\usepackage[textsize=tiny]{todonotes}

\icmltitlerunning{Emergent Asymmetry of Precision and Recall in High Dimensions}

\begin{document}

\twocolumn[
\icmltitle{Emergent Asymmetry of Precision and Recall for Measuring Fidelity and Diversity of Generative Models in High Dimensions}



\icmlsetsymbol{equal}{*}
\begin{icmlauthorlist}
\icmlauthor{Mahyar Khayatkhoei}{uscisi}
\icmlauthor{Wael AbdAlmageed}{uscisi,uscee}
\end{icmlauthorlist}

\icmlaffiliation{uscisi}{Information Sciences Institute, University of Southern California, Marina del Rey, CA, USA}

\icmlaffiliation{uscee}{Ming Hsieh Department of Electrical and Computer Engineering, University of Southern California, Los Angeles, CA, USA}

\icmlcorrespondingauthor{Mahyar Khayatkhoei}{mkhayat@isi.edu}

\icmlkeywords{Precision, Recall, Coverage, High dimensional, Generative Models, Machine Learning, Curse of Dimensionality, Emergent Asymmetry}

\vskip 0.3in
]



\printAffiliationsAndNotice{}  

\begin{abstract}
Precision and Recall are two prominent metrics of generative performance, which were proposed to separately measure the fidelity and diversity of generative models. Given their central role in comparing and improving generative models, understanding their limitations are crucially important. To that end, in this work, we identify a critical flaw in the common approximation of these metrics using k-nearest-neighbors, namely, that the very interpretations of fidelity and diversity that are assigned to Precision and Recall can fail in high dimensions, resulting in very misleading conclusions. Specifically, we empirically and theoretically show that as the number of dimensions grows, two model distributions with supports at equal point-wise distance from the support of the real distribution, can have vastly different Precision and Recall regardless of their respective distributions, hence an emergent asymmetry in high dimensions. Based on our theoretical insights, we then provide simple yet effective modifications to these metrics to construct symmetric metrics regardless of the number of dimensions. Finally, we provide experiments on real-world datasets to illustrate that the identified flaw is not merely a pathological case, and that our proposed metrics are effective in alleviating its impact.
\end{abstract}

\begin{figure*}[t]
     \centering
     \subfloat[2-Sphere Setting]{
        \centering
        \includegraphics[trim=0 0 0 0, clip, width=0.27\textwidth]{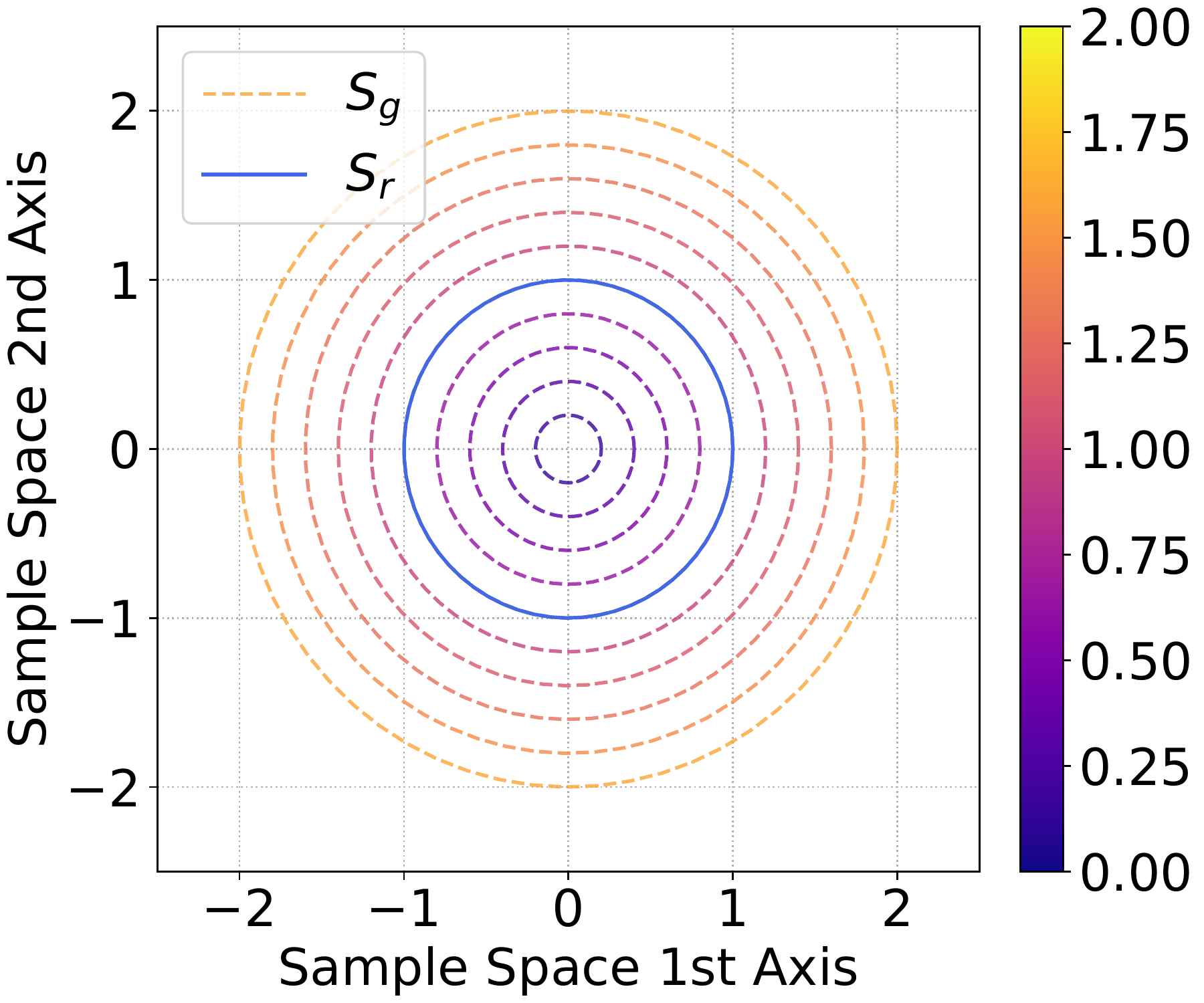}
        \label{fig:sphere_r1_pr_example}}
    \subfloat[Precision]{
        \centering
        \includegraphics[trim=25 0 0 25, clip, width=0.3\textwidth]{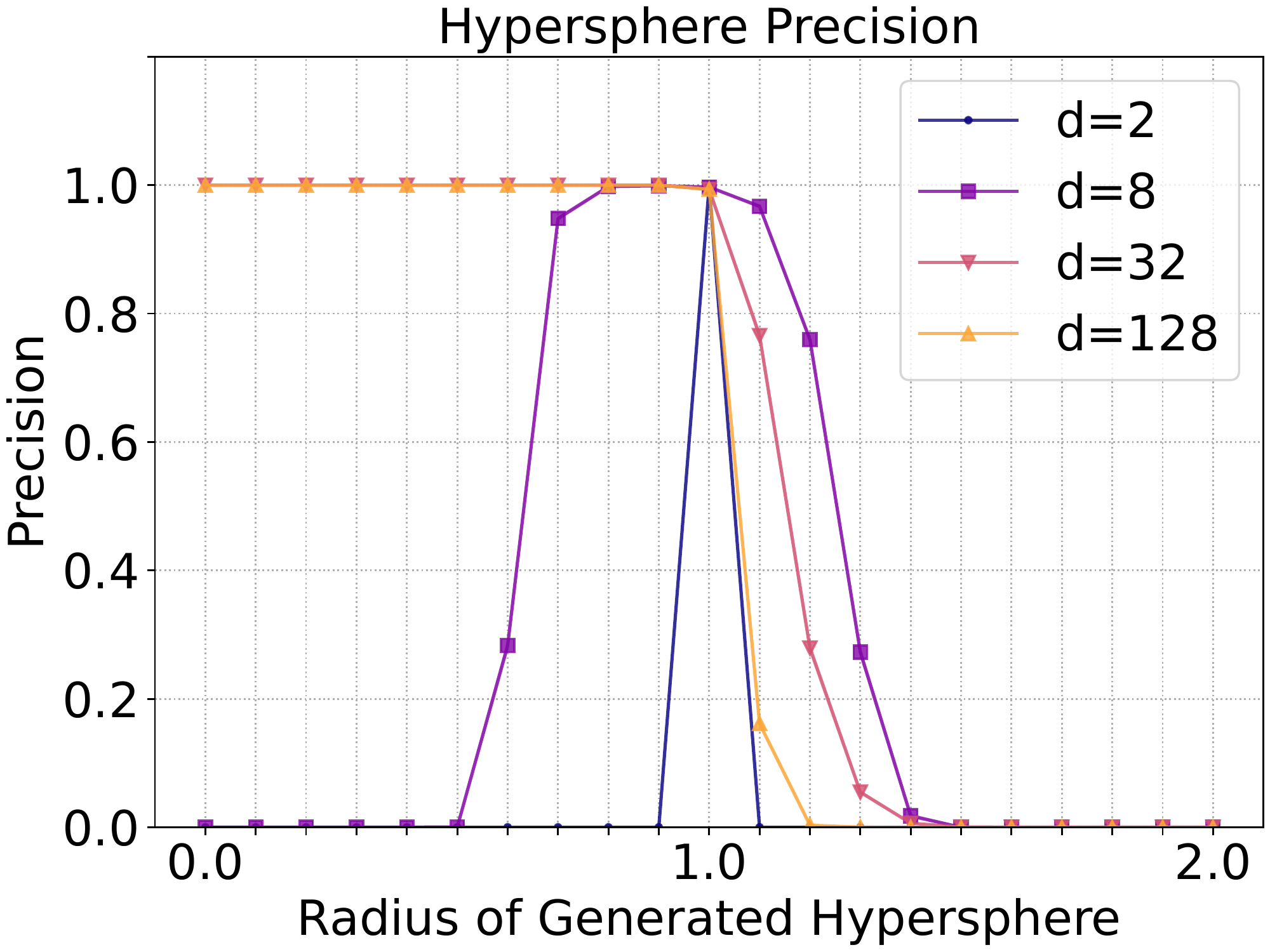}
        \label{fig:sphere_r1_pr_p}}
    \subfloat[Recall]{
        \centering
        \includegraphics[trim=25 0 0 25, clip, width=0.3\textwidth]{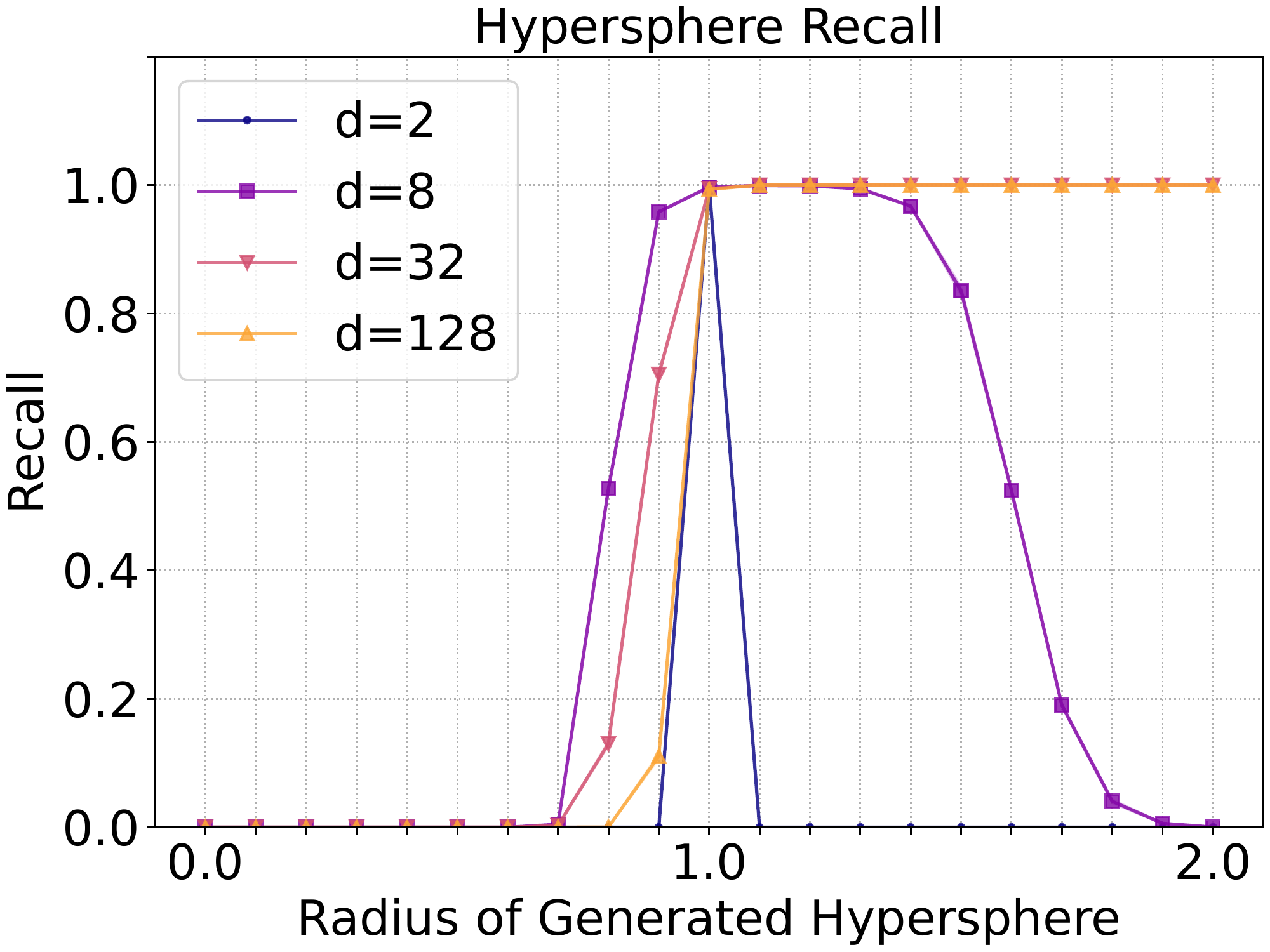}
        \label{fig:sphere_r1_pr_r}}
    \caption{Asymmetry of Precision and Recall with Hyperspherical supports. (a) Illustrates the setup of the experiment in $d=2$, where the solid blue line denotes the reference unit 2-sphere support, and the dashed lines denote generated 2-sphere supports of varying distances from the reference (radius on colorbar). (b, c) The generated support being outside or inside the reference support results in vastly different measures, becoming more asymmetric as the number of dimensions grows. Same behavior is observed with other radii (see~\cref{sec:varying_r}).}
    \label{fig:sphere_r1_pr}
\end{figure*}
\begin{figure*}[t]
    \centering
    \subfloat[Precision (Hypercube)]{
        \centering
        \includegraphics[trim=25 0 0 25, clip, width=0.253\textwidth]{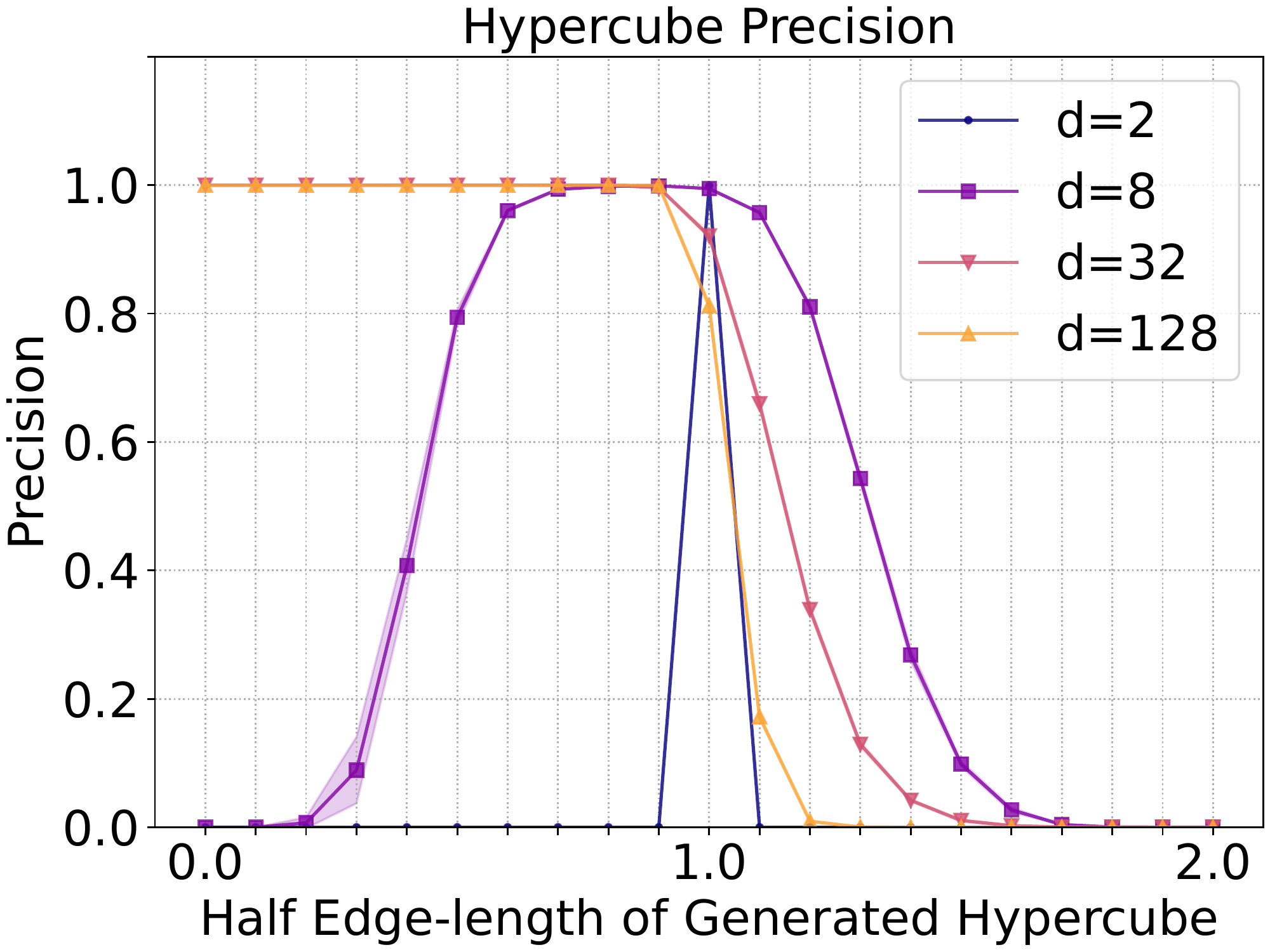}
        \label{fig:cube_r1_pr_p}}
    \subfloat[Recall (Hypercube)]{
        \centering
        \includegraphics[trim=64 0 0 25, clip, width=0.235\textwidth]{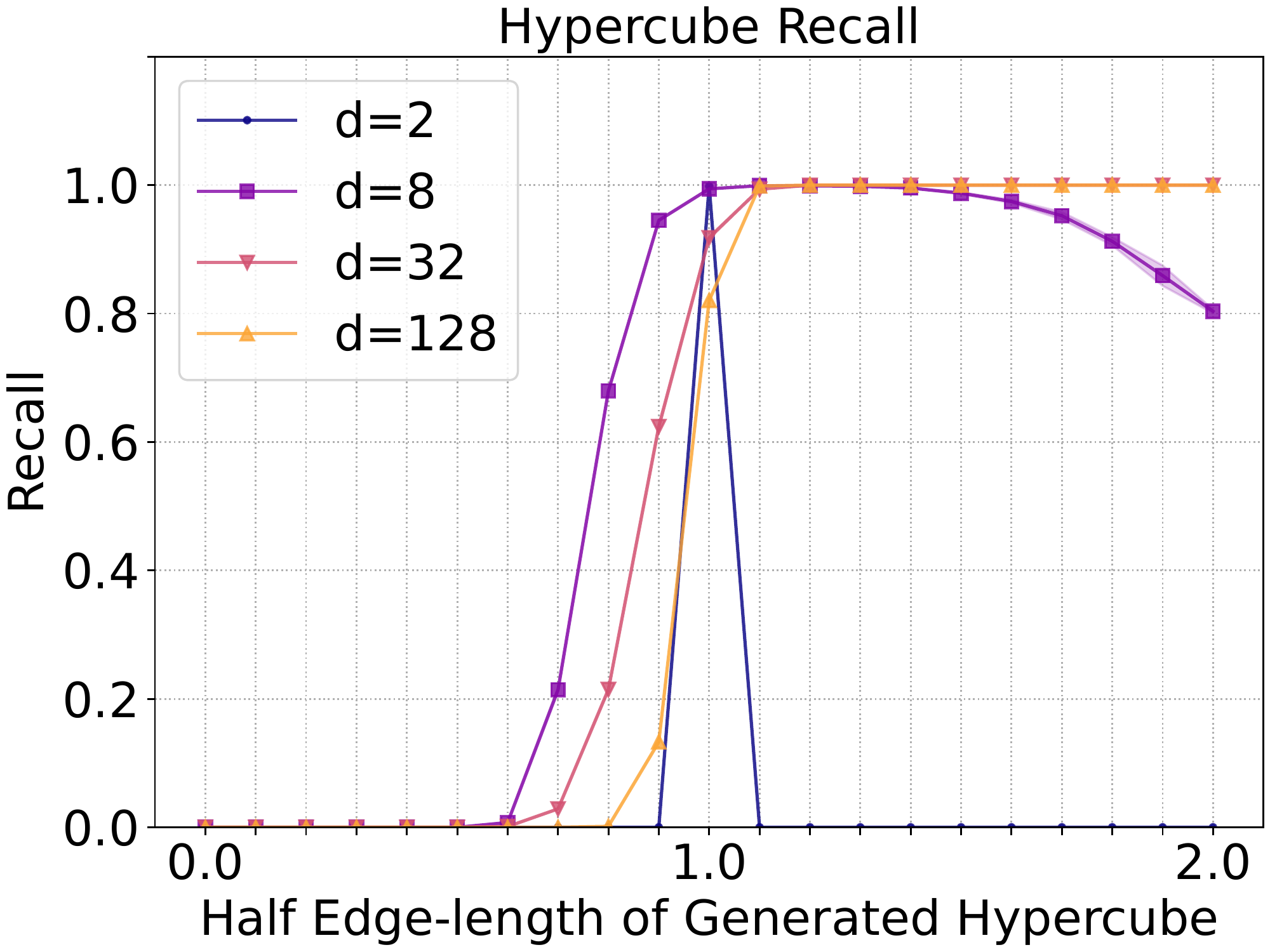}
        \label{fig:cube_r1_pr_r}}
    \subfloat[Precision (Gaussian)]{
        \centering
        \includegraphics[trim=64 0 0 25, clip, width=0.235\textwidth]{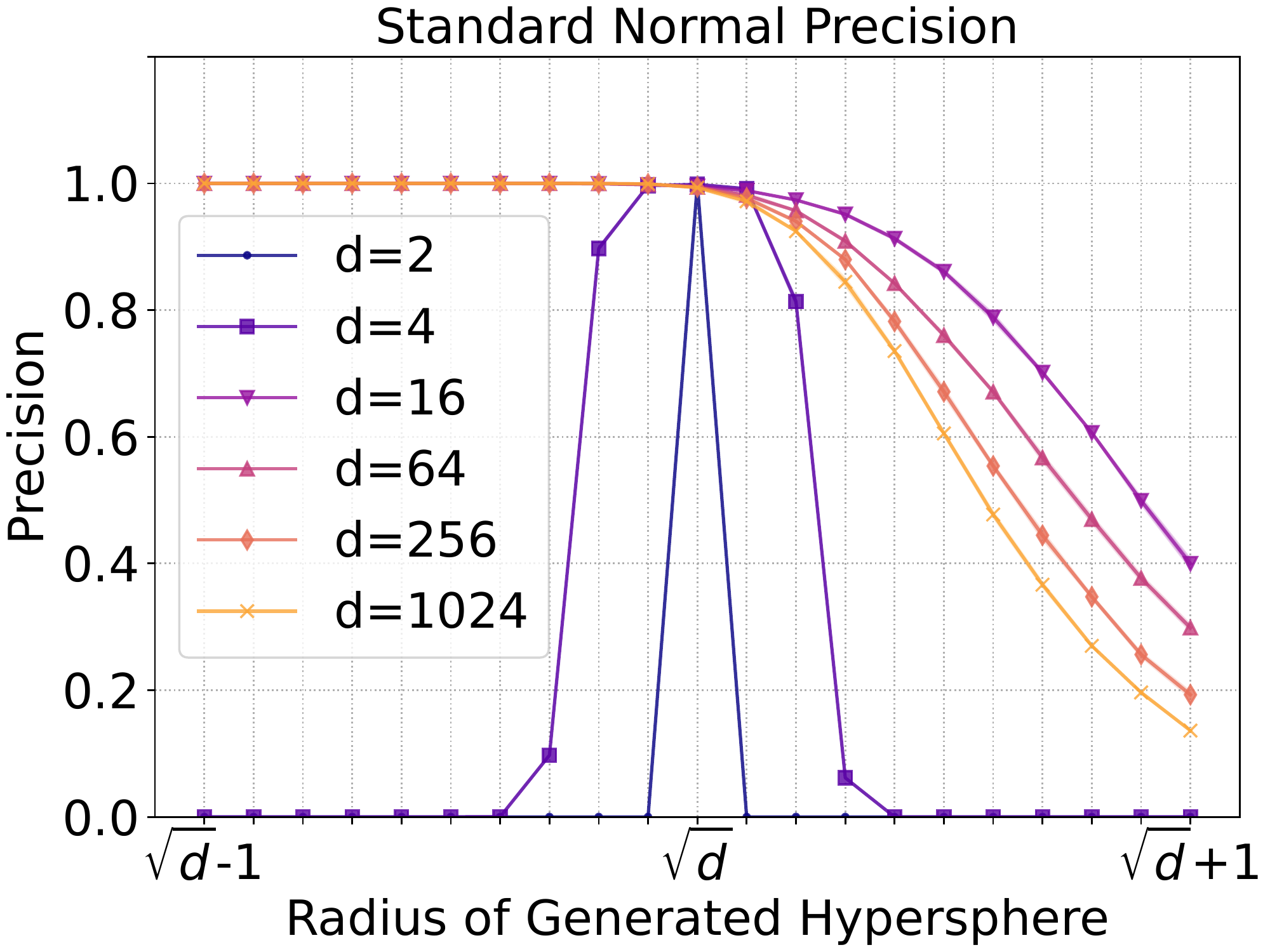}
        \label{fig:normalsphere_r1_pr_p}}
    \subfloat[Recall (Gaussian)]{
        \centering
        \includegraphics[trim=64 0 0 25, clip, width=0.235\textwidth]{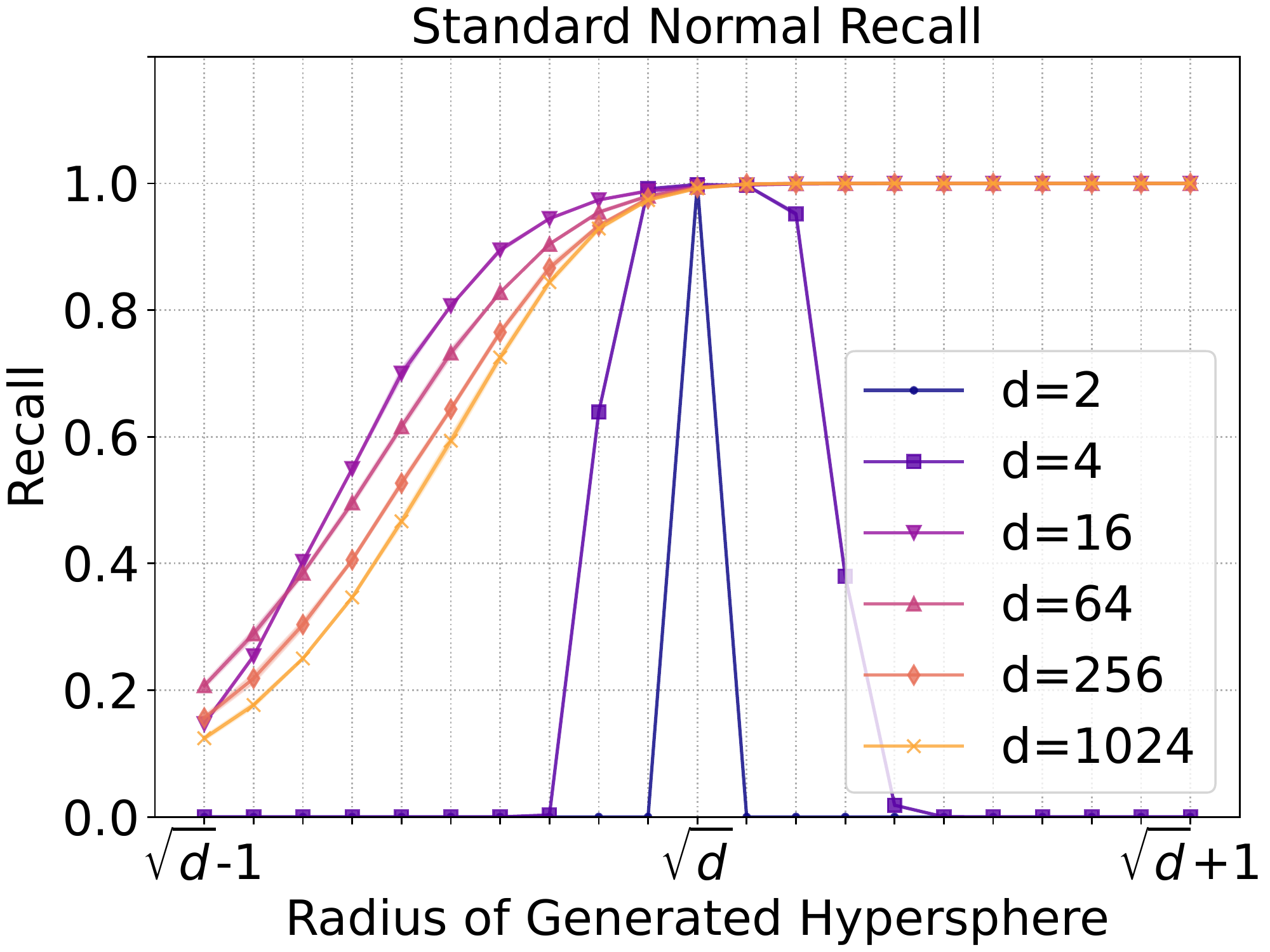}
        \label{fig:normalsphere_r1_pr_r}}
    \caption{Asymmetry of Precision and Recall with Hypercubic supports, and Gaussian spherical supports. (a, b) The reference support is the hypercube of half-edge-length 1. (c, d) The reference support is a hypersphere at $\sqrt{\textrm{d}}$. The generated support being outside or inside the reference support results in vastly different behavior, which becomes more and more asymmetric as dimensions grow.}
    \label{fig:cube_r1_pr}
\end{figure*}
\section{Introduction}
\label{sec:intro}
Accurately measuring the performance of generative models has become a major challenge due to the rapid growth of their application in downstream tasks -- from super-resolution~\cite{pathak2018superres-gan}, to data-augmentation~\cite{sandfort2019data-aug-gan}, and art creation~\cite{rombach2022stable-diffusion}. To address this challenge, initially qualitative and heuristic measures of difference between generated samples and real samples were proposed~\cite{arora2018birthday, salimans2016inception-score, hore2010ssim}, followed by more recent moment-based distances~\cite{bińkowski2018kid, heusel2017fid} and neural network distances~\cite{karras2020stylegan2, ravuri2019cas}, which could provide a more consistent evaluation with human perception~\cite{lucic2018create-equal}. However, these metrics were lacking in one important aspect: separately measuring fidelity and diversity of generated samples. To address this shortcoming, Precision and Recall~\cite{sajjadi2018precision-recall}, and their later improved versions~\cite{kynkaanniemi2019improved-precision-recall}, were proposed (henceforth Precision and Recall refers to the improved versions).

Given finite sets of samples $X_r$ and $X_g$, from a real and a generated distribution, $p_r$ and $p_g$, Precision measures the fraction of generated samples that fall within the support of the real distribution approximated using $K$-nearest-neighbors (fidelity of generated samples), whereas Recall measures the fraction of real samples that fall within the support of the approximated generated distribution (meaningful diversity of generated samples):
\begin{align}
    \begin{split}
        \textrm{Precision} &(p_r, p_g) = \Px{p_g}{\hat{S_r} \cap S_g}\\
        &\approx \frac{1}{|X_g|} \sum_{x_i \in X_g} 1(x_i \in \hat{S_r})
    \end{split}
    \label{eq:precision}\\
    \begin{split}
        \textrm{Recall} &(p_r, p_g)= \Px{p_r}{S_r \cap \hat{S_g}}\\
        &\approx \frac{1}{|X_r|} \sum_{x_i \in X_r} 1(x_i \in \hat{S_g})
    \end{split}
    \label{eq:recall}
\end{align}
where $1(.)$ is the indicator function, $S_r$ and $S_g$ are the supports of $p_r$ and $p_g$, and \textit{hat} denotes approximation of support by $K$-nearest-neighbors (details in~\cref{sec:theory}).

Given the ubiquitous adoption of Precision and Recall in practice, recent works have focused on studying the limitations of these metrics~\citep{naeem2020density-coverage, alaa2022alphabetapr}. In the same spirit, in this work, we identify and formalize a critical flaw in Precision and Recall, namely, that their very interpretations as fidelity and diversity could fail in high dimensions. More specifically, two model distributions with supports $S_g$ and $S_g'$ at equal point-wise distance from the support of the real distribution $S_r$, can have vastly different Precision and Recall regardless of their respective distributions $p_g$ and $p_g'$, hence an emergent asymmetry in high dimensions. Even worse, $S_g'$ can be quite far from $S_r$, while $S_g$ is very close to $S_r$, and yet any $p_g'$ can achieve much higher Precision (or Recall) compared to any $p_g$. Consequently, comparing distributions in high dimensions in terms of Precision and Recall becomes nearly meaningless.

The main contribution of this paper is to empirically and theoretically show the existence of an emergent asymmetry in Precision and Recall, and its consequences in practice. The rest of this paper is organized as follows: we start by providing a motivating example in~\cref{sec:example} that empirically shows the existence of the asymmetry, and then analytically prove its existence in~\cref{sec:theory}; next, in~\cref{sec:symmetricpr}, we use the insights from the developed theory to design modified versions of Precision and Recall that are symmetric in low and high dimensions; finally, in~\cref{sec:exp}, we provide experiments on two real-world datasets, CelebA~\cite{liu2015celeba} and CIFAR10~\cite{krizhevsky2009cifar}, to show the existence of the asymmetry in practice, and to illustrate the effectiveness of our proposed metrics. We close this paper with a discussion of related works in~\cref{sec:related_works} followed by remaining questions and future directions in~\cref{sec:discussion}.

\section{A Motivating Example}
\label{sec:example}
In this section, we will provide a motivating example that illustrates how in high dimensions the expected interpretation of Precision and Recall as measures of fidelity and diversity fails. Consider a real distribution $p_r$ whose support $S_r$ is the surface of the unit $d-1$ dimensional sphere, and a generated/learnt distribution $p_g$ whose support $S_g$ is the surface of another $d-1$ dimensional sphere with the same center point, both uniformly distributed. In two dimensions, these supports would be two concentric circles (as in~\cref{fig:sphere_r1_pr_example}), in three dimensions two concentric spheres, and so forth. Now imagine we start increasing the radius of $S_g$ from zero to infinity. The common understanding of Precision and Recall suggests that as $S_g$ approaches $S_r$ and passes it, we should observe a bump in both Precision and Recall, and otherwise both should be zero. As~\cref{fig:sphere_r1_pr} shows, this expected behavior is correctly observed in low dimensions, however, as the number of dimensions increases we observe a strikingly different behavior: the generated support being slightly outside or inside the real support results in vastly different Precision and Recall, hence an emergent asymmetry in high dimensions. To make sure this is not just a peculiarity of hyperspheres, we repeat the same experiment with hypercubes in~\cref{fig:cube_r1_pr}, and observe the same behavior. We also repeat the experiment for the support of standard Gaussian distributions, which resembles a sphere of radius $\sqrt{d}$, and again observe the same behavior in~\cref{fig:cube_r1_pr}. Using different radii and K values for the K-nearest-neighbors~\footnote{For radii and K in typical range of $\ll d$ the asymmetry is observed. However, for large values, Precision and Recall appear to saturate everywhere, losing any significance.} results in the same behavior as well (\cref{sec:varying_r,sec:varying_k}).

This behavior breaks the general intuition of fidelity and diversity assigned to Precision and Recall in high dimensions. For example, a generated distribution that is slightly outside the real distribution, will have much lower Precision than one that is far away from the real distribution but inside, resulting in the misleading conclusion that the much farther latter distribution is actually generating samples with higher fidelity. Similarly, a distribution slightly inside the real distribution will be seen as having a much lower diversity compared to a distribution far outside of the real distribution, according to Recall. These issues are not mere pathological cases and can have practical consequences: when an algorithm tries to match its generated support to that of the high dimensional real data -- as is the case in most prominent generative models such as GANs~\cite{goodfellow2014gan}, VAEs~\cite{kingma2014vae}, and Diffusion Models~\cite{ho2020denoising} -- it would approach the real support from different directions and oscillate near it~\cite{khayatkhoei2018disconnected, mescheder2018r1regul}, at which point Precision and Recall are typically used to choose the trade-off between diversity and fidelity~\cite{kynkaanniemi2019improved-precision-recall}, and compare models with one another, however, the phenomenon observed in~\cref{fig:sphere_r1_pr,fig:cube_r1_pr} can render such trade-off decisions based on Precision and Recall meaningless.

\section{Emergent Asymmetry in High Dimensions}
\label{sec:theory}
In this section, our goal is to mathematically explain the asymmetry observed in~\cref{sec:example} and the mechanisms behind it. Our analysis will be restricted to the case of distributions supported on hyperspheres, however, we will comment on its generality later in this section.

We start by stating the setup for our analysis. We consider reference and generated distributions, $p_r$ and $p_g$, that are absolutely continuous on their supports $S_r$ and $S_g$, respectively, and further assume $S_r$ is the surface of a $d-1$ dimensional hypersphere. Given a random set of observations $X_r=\{x_i\}_{i=1}^N$ from $p_r$, we construct a $K$-nearest-neighbors approximation to the reference support, that is $\hat{S_r} = \cup_{i=1}^N N_K(x_i) \approx S_r$ where $N_K(x_i)$ is the $d$ dimensional ball centered at $x_i$ with radius equal to the Euclidean distance of $x_i$ from its $K$-th nearest neighbor in $X_r$. This constitutes the support approximation typically used for the calculation of Precision and Recall. Now, to explain the behavior observed in~\cref{sec:example}, we will analyze the Precision of $p_g$ -- the measure of overlap between the approximated reference support and the generated support with respect to the probability measure defined by $p_g$ -- in two cases: when $p_g$ is contained inside and outside of $S_r$.

To study the first case, we assume $p_g$ has support inside the hypersphere $S_r$. More concretely, $S_g = B$ where $B$ is the $d$-ball whose boundary is $\partial B = S_r$. In this case, the following theorem shows that given a practical number of samples (\eg $N$ being polynomial in $d$), the Precision will approach 1 with high probability asymptotically in the number of dimensions. Intuitively, this means that in high dimensions, any absolutely continuous distribution contained inside the hypersphere will be completely covered by the approximated reference distribution.

\begin{proposition}
\label{proposition:inside}
Given reference and generated distributions $p_r$ and $p_g$, absolutely continuous on their respective supports $S_r$ and $S_g$, where $S_r$ is a $d-1$ dimensional hypersphere and $S_g$ is the $d$ dimensional ball $B$ with boundary $S_r$, and the $K$-nearest-neighbors approximation of $S_r$ using $N$ samples from $p_r$ denoted $\hat{S_r}$, if $\lim_{d\rightarrow\infty} N\epsilon^{-d} = 0$ $\forall \epsilon>1$, then with arbitrarily high probability we have:
\begin{align}
    \lim_{d\rightarrow\infty} \Px{p_g}{\hat{S_r} \cap S_g} = 1
\end{align}
\end{proposition}
{\begin{proof}\renewcommand{\qedsymbol}{}In~\cref{sec:proof_inside}.\end{proof}}
This result is intuitively expected from the fact that the volume of a hypersphere of any radius asymptotically (in the number of dimensions) tends to zero, however, note that this fact on its own cannot explain the saturating Precision: given arbitrarily many samples we have $\hat{S}_r = S_r$ and thus $\hat{S_r} \cap S_g = S_r \cap S_g = \partial B \cap B$ which has measure zero under $p_g$ regardless of the number of dimensions. Therefore, the main reason why Precision saturates in~\cref{proposition:inside} is that the number of samples have a sub-exponential growth in the number of dimensions $d$, which in turn suggests that this behavior is not an intrinsic property of hyperspheres in high dimensions, rather an instance of the curse of dimensionality limiting our approximation ability. Note that since Recall is equal to Precision when swapping the generated and reference distributions (see~\cref{eq:precision,eq:recall}), the result in~\cref{proposition:inside} readily extends to Recall. Specifically, it explains that Recall will approach 1 with high probability as the number of dimensions grows, when the reference distribution is inside the generated distribution.

Next, we study the second case, where we assume $p_g$ has support outside of the hypersphere $S_r$. More concretely, we assume $S_g$ is inside a $d$-ball $B_o$ containing the $d$-ball $B$ whose boundary is $\partial B = S_r$, and outside $B$. In this case, the following theorem shows that given a practical number of samples (\eg $N$ being polynomial in $d$), the Precision will approach 0 with high probability asymptotically in the number of dimensions. Intuitively, this means that in high dimensions, any absolutely continuous distribution contained outside of the hypersphere will have no overlap with the approximated reference distribution.

\begin{proposition}
\label{proposition:outside}
Given a reference distribution $p_r$ whose support $S_r$ is the $d-1$ dimensional hypersphere, a generated distribution $p_g$ absolutely continuous on its support $S_g$ which is outside the $d$ dimensional ball $B$ with boundary $\partial B = S_r$ and inside a $d$ dimensional ball $B_o \supset B$, \ie $S_g = B_o\setminus (B\setminus\partial B)$, and the $K$-nearest-neighbors approximation of $S_r$ using $N$ samples from $p_r$ denoted $\hat{S_r}$, if $\lim_{d\rightarrow\infty} N\epsilon^{-d} = 0$ $\forall \epsilon > 1$, then with arbitrarily high probability we have:
\begin{align}
    \lim_{d\rightarrow\infty} \Px{p_g}{\hat{S_r} \cap S_g} = 0
\end{align}
\end{proposition}
{\begin{proof}\renewcommand{\qedsymbol}{}In~\cref{sec:proof_outside}.\end{proof}}
This result also readily extends to Recall by swapping the generated and reference distributions as explained before, that is, Recall will approach 0 with high probability as the number of dimensions grows, when the reference distribution is outside the generated distribution.

There are two main assumptions in the above results that require justification for why they are sensible in practice. First, the assumption of absolutely continuous distributions is sensible when considering the fact that representing the support of distributions in digital computers is subject to numerical rounding errors, which means we can always assume the presence of an infinitesimal amount of noise in the true distribution such that it becomes absolutely continuous. Additionally, various regularization techniques are often used to explicitly avoid close to measure zero supports for $p_g$~\cite{arjovsky2017towards}. The second assumption is the hyperspherical supports. This is indeed diverging from the real-world situation of complicated manifolds as supports. Nonetheless, we think studying this restricted case provides valuable insights into the behavior of Precision and Recall in practice, because while practical distributions have complicated supports, these supports do share some defining characteristics with hyperspheres, most notably, being closed (compact and without boundary). Considering practical distributions as being supported on manifolds that can be well represented by digital computers, their compactness follows from the set of floating-point numbers being finite, and being without boundary follows from the common assumption that the data manifold is everywhere locally homeomorphic to the same Euclidean space. For a more specific discussion focused on image patches see~\cite{carlsson2008local}. Furthermore, we will see in~\cref{sec:exp} that empirical results on real-world datasets are consistent with the behavior we observed and proved for hyperspheres. We conjecture that this behavior is more generally true for any distribution $p_r$ supported on the boundary of a compact space, however, we were not able to prove this at present.

While the above results reveal the mechanism behind the asymmetry we observed in~\cref{sec:example}, they also suggest a potential solution: in the proof of~\cref{proposition:inside}, we observe that a critical step giving rise to the asymmetric behavior is approximating the support of $p_r$ with $K$-nearest-neighbors. If we were to instead approximate the support of $p_g$, then we would end up with a setup that resembles that of~\cref{proposition:outside} in that the approximated support is now placed inside the other support. Similarly, the setup of~\cref{proposition:outside} would resemble that of~\cref{proposition:inside} by changing the distribution that is being approximated. As such, it seems possible to maintain the diversity and fidelity interpretations of Precision and Recall, while inverting the asymmetry. In the following section, we will take advantage of this observation to modify Precision and Recall such that they become more symmetric in high dimensions.

\begin{figure*}[t]
    \centering
    \subfloat[cPrecision (Hypersphere)]{
        \centering
        \includegraphics[trim=25 0 0 25, clip, width=0.253\textwidth]{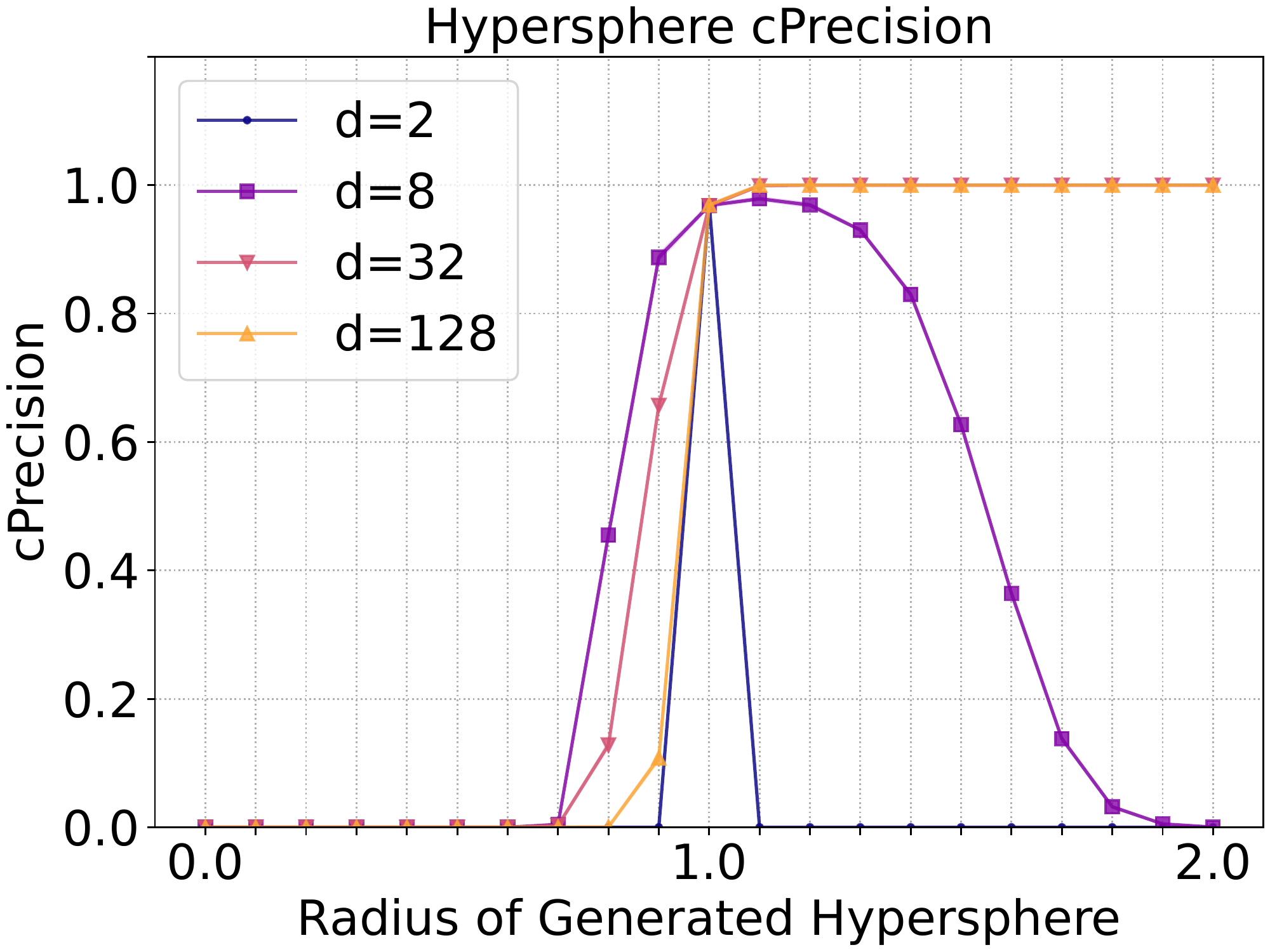}
        \label{fig:sphere_r1_pr_cp}}
    \subfloat[cRecall (Hypersphere)]{
        \centering
        \includegraphics[trim=64 0 0 25, clip, width=0.235\textwidth]{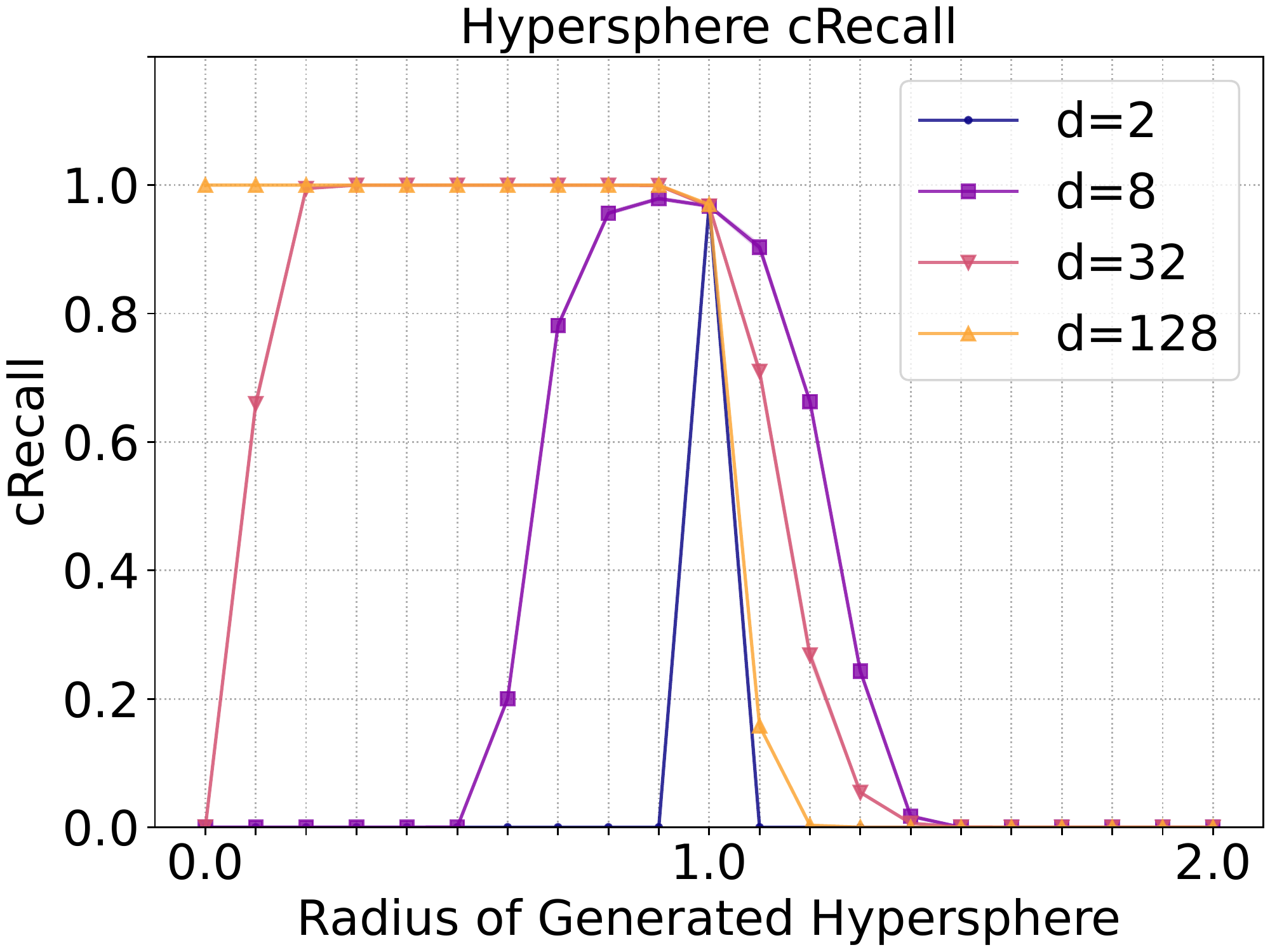}
        \label{fig:sphere_r1_pr_cr}}
    \subfloat[symPrecision (Hypersphere)]{
        \centering
        \includegraphics[trim=64 0 0 25, clip, width=0.235\textwidth]{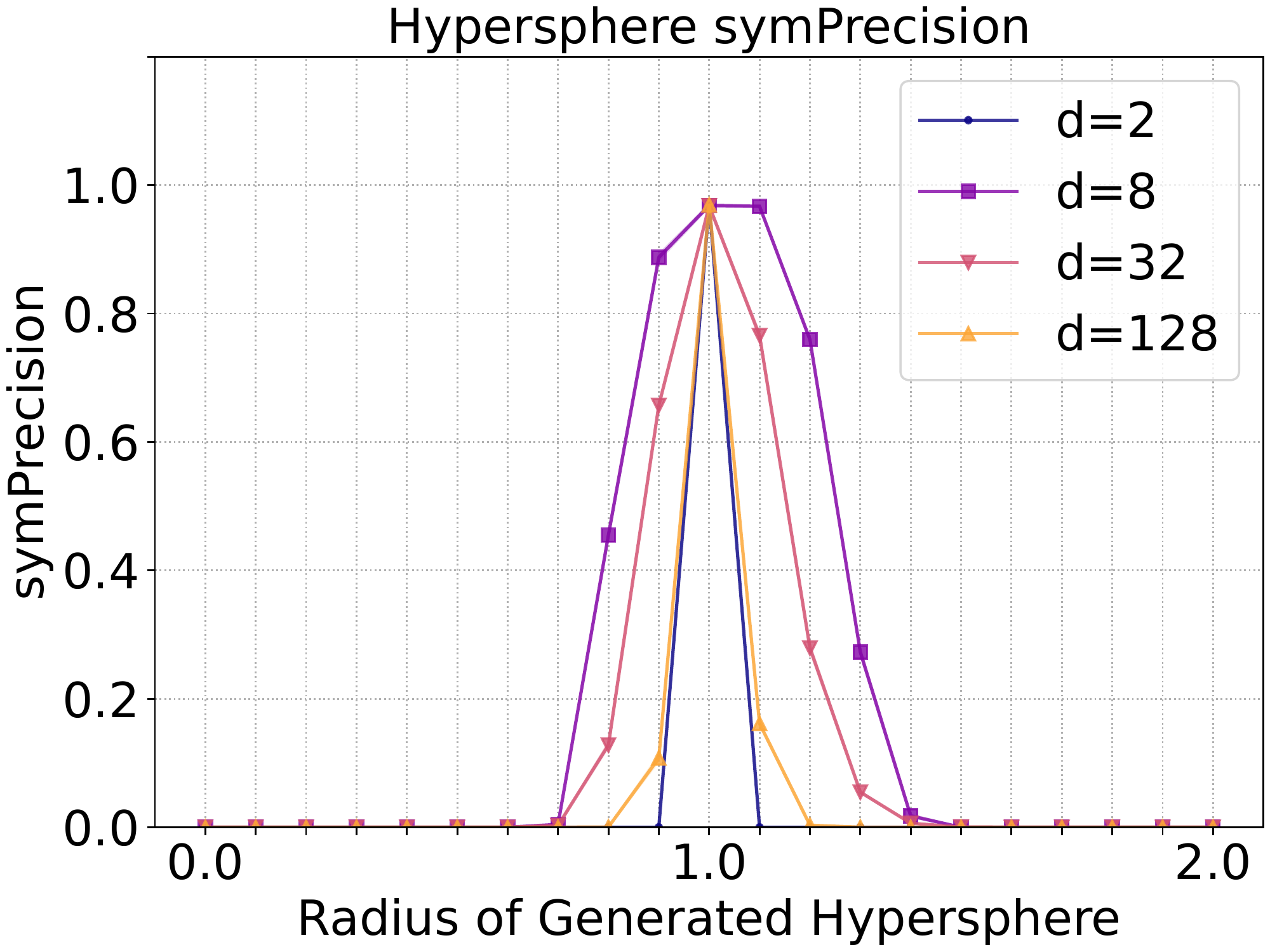}
        \label{fig:sphere_r1_pr_sp}}
    \subfloat[symRecall (Hypersphere)]{
        \centering
        \includegraphics[trim=64 0 0 25, clip, width=0.235\textwidth]{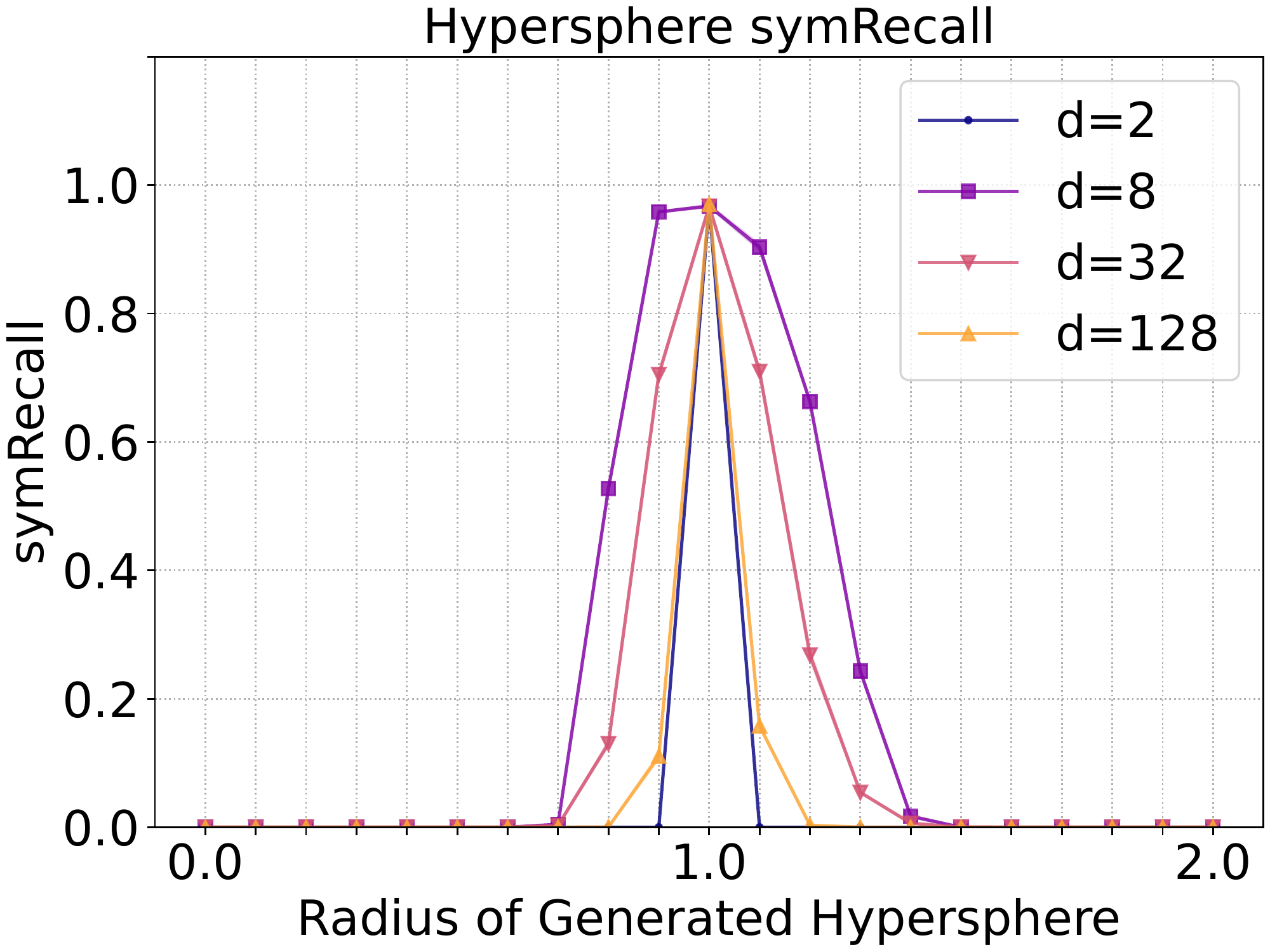}
        \label{fig:sphere_r1_pr_sr}}\\
    \subfloat[cPrecision (Hypercube)]{
        \centering
        \includegraphics[trim=25 0 0 25, clip, width=0.253\textwidth]{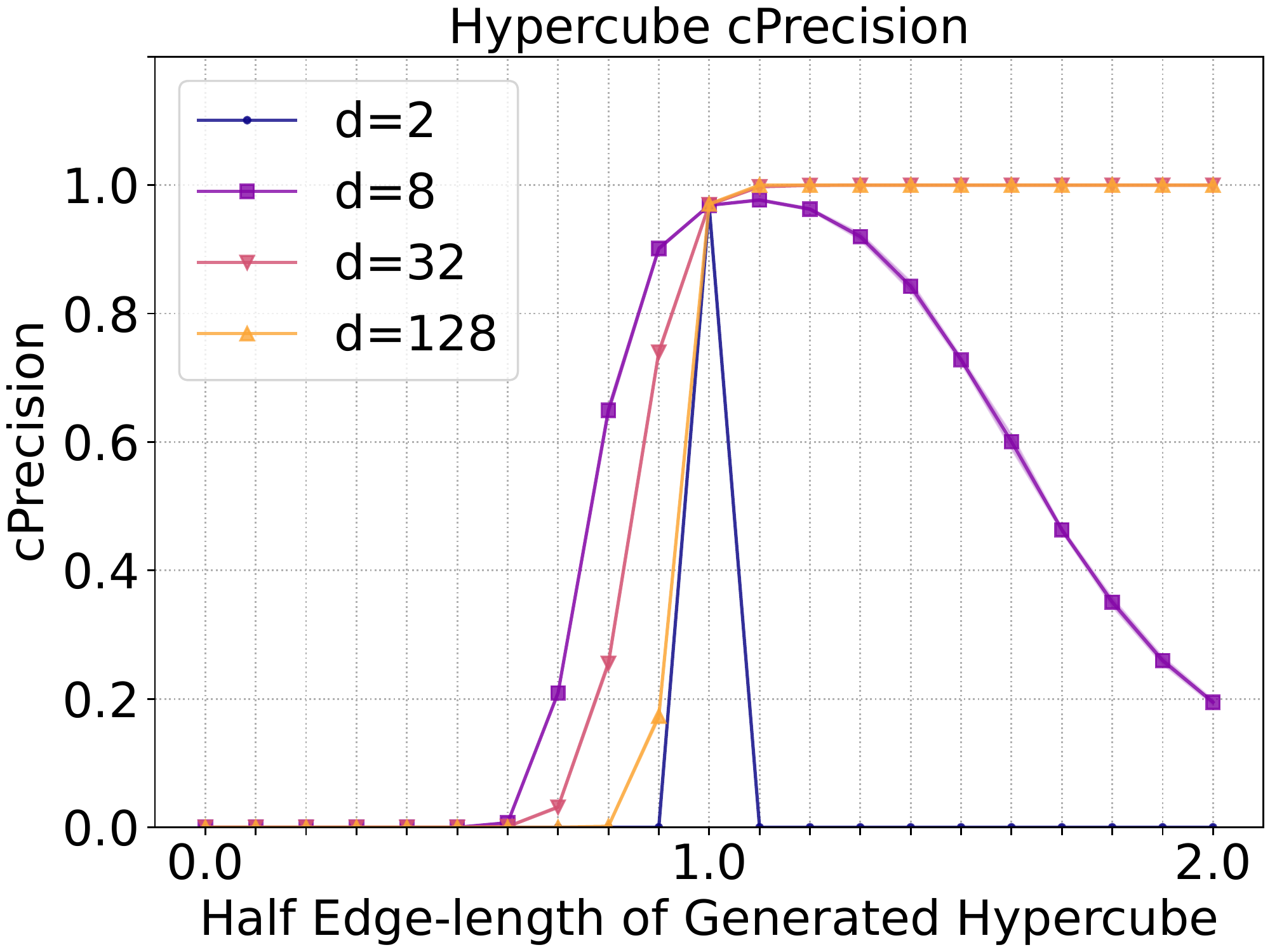}
        \label{fig:cube_r1_pr_cp}}
    \subfloat[cRecall (Hypercube)]{
        \centering
        \includegraphics[trim=64 0 0 25, clip, width=0.235\textwidth]{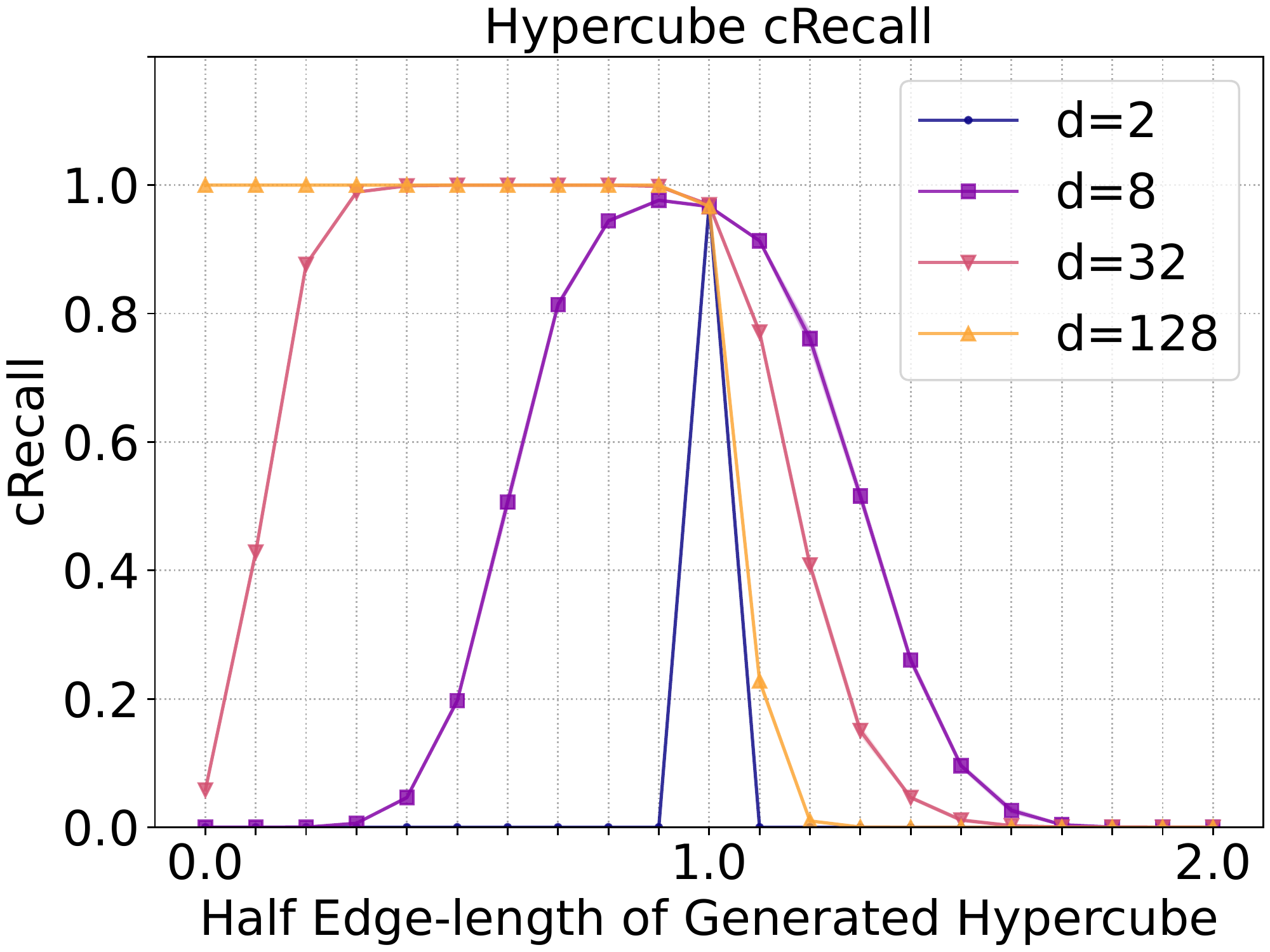}
        \label{fig:cube_r1_pr_cr}}
    \subfloat[symPrecision (Hypercube)]{
        \centering
        \includegraphics[trim=64 0 0 25, clip, width=0.235\textwidth]{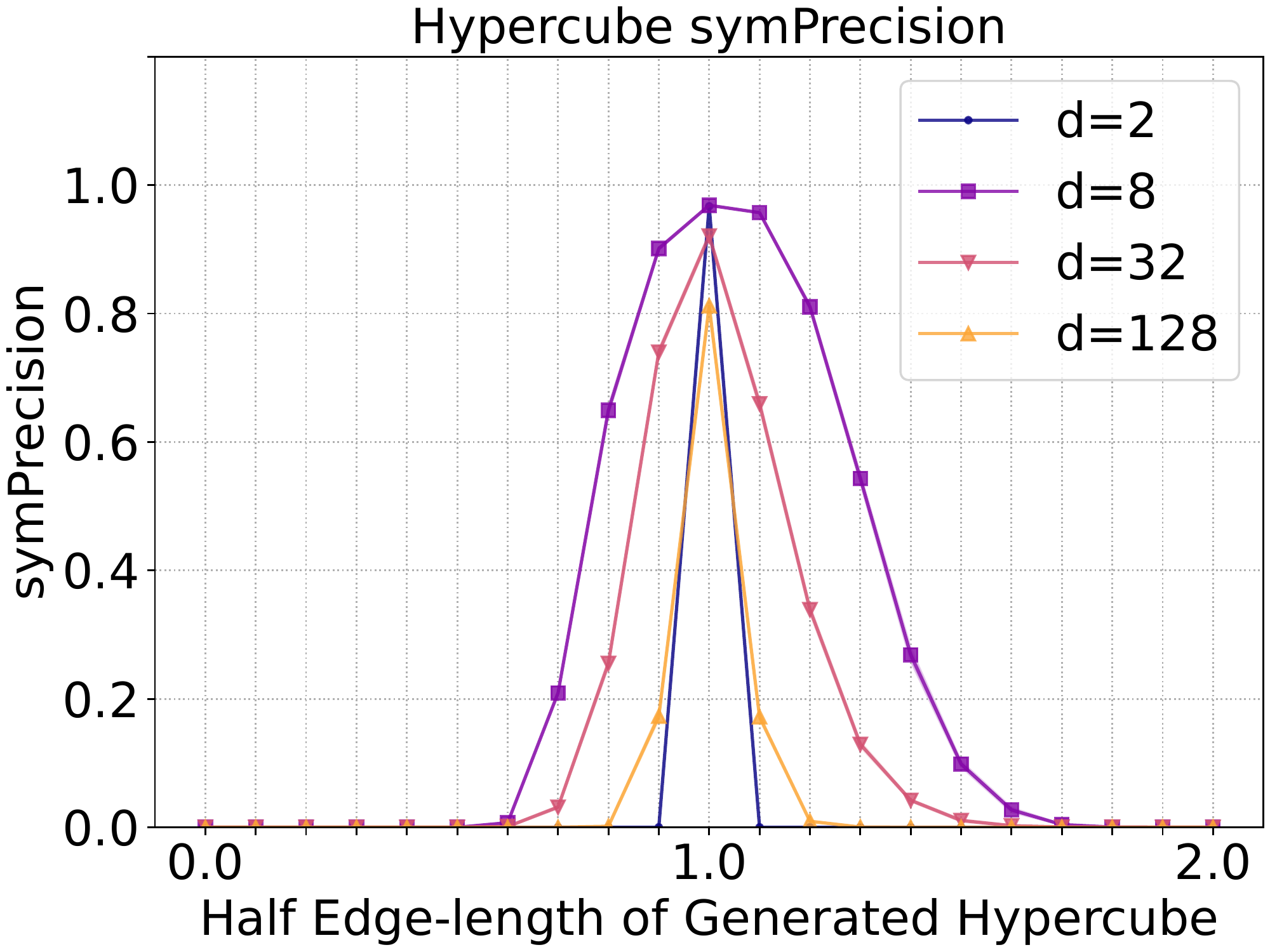}
        \label{fig:cube_r1_pr_sp}}
    \subfloat[symRecall (Hypercube)]{
        \centering
        \includegraphics[trim=64 0 0 25, clip, width=0.235\textwidth]{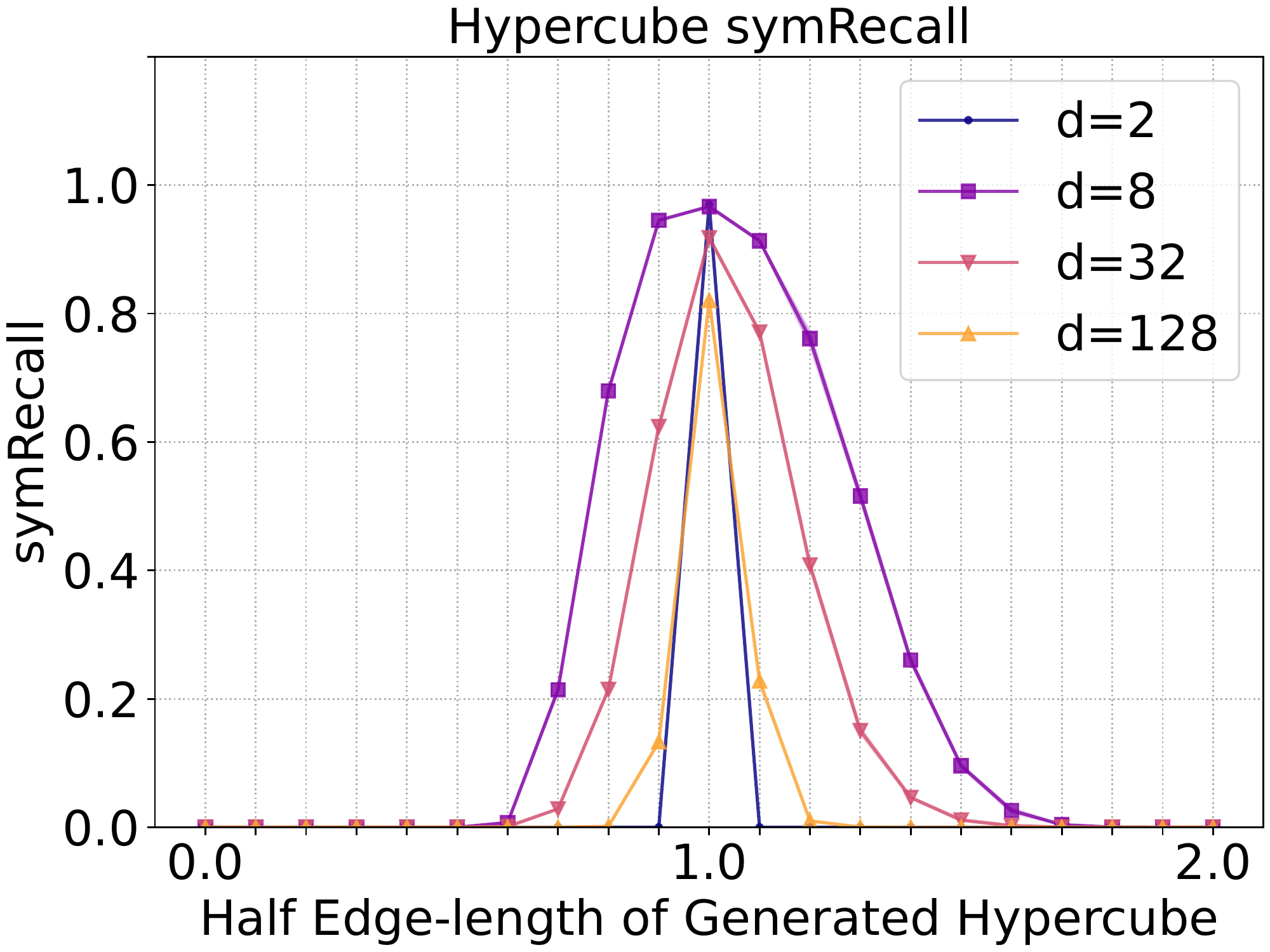}
        \label{fig:cube_r1_pr_sr}}\\
    \subfloat[cPrecision (Gaussian)]{
        \centering
        \includegraphics[trim=25 0 0 25, clip, width=0.253\textwidth]{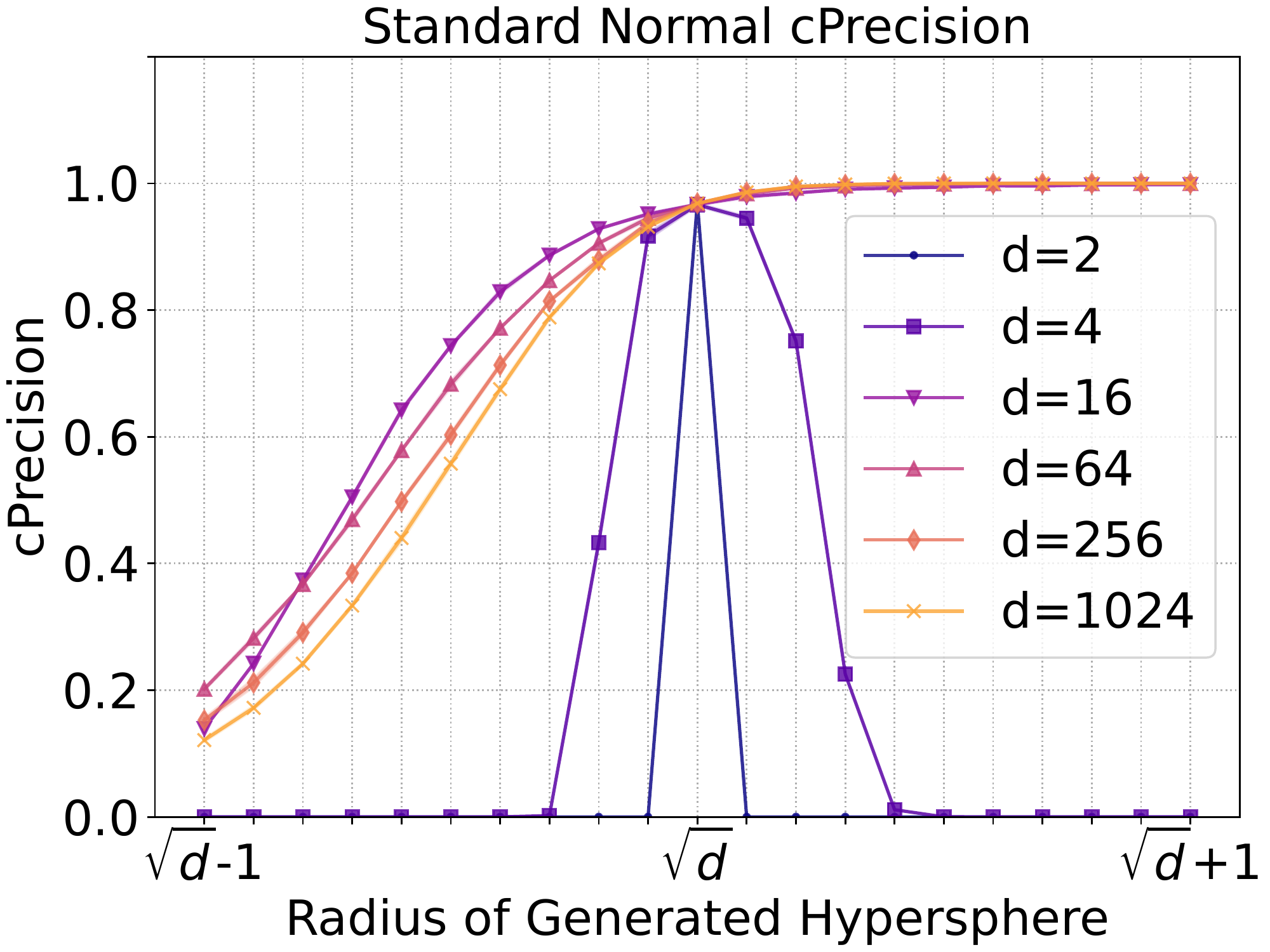}
        \label{fig:normalsphere_r1_pr_cp}}
    \subfloat[cRecall (Gaussian)]{
        \centering
        \includegraphics[trim=64 0 0 25, clip, width=0.235\textwidth]{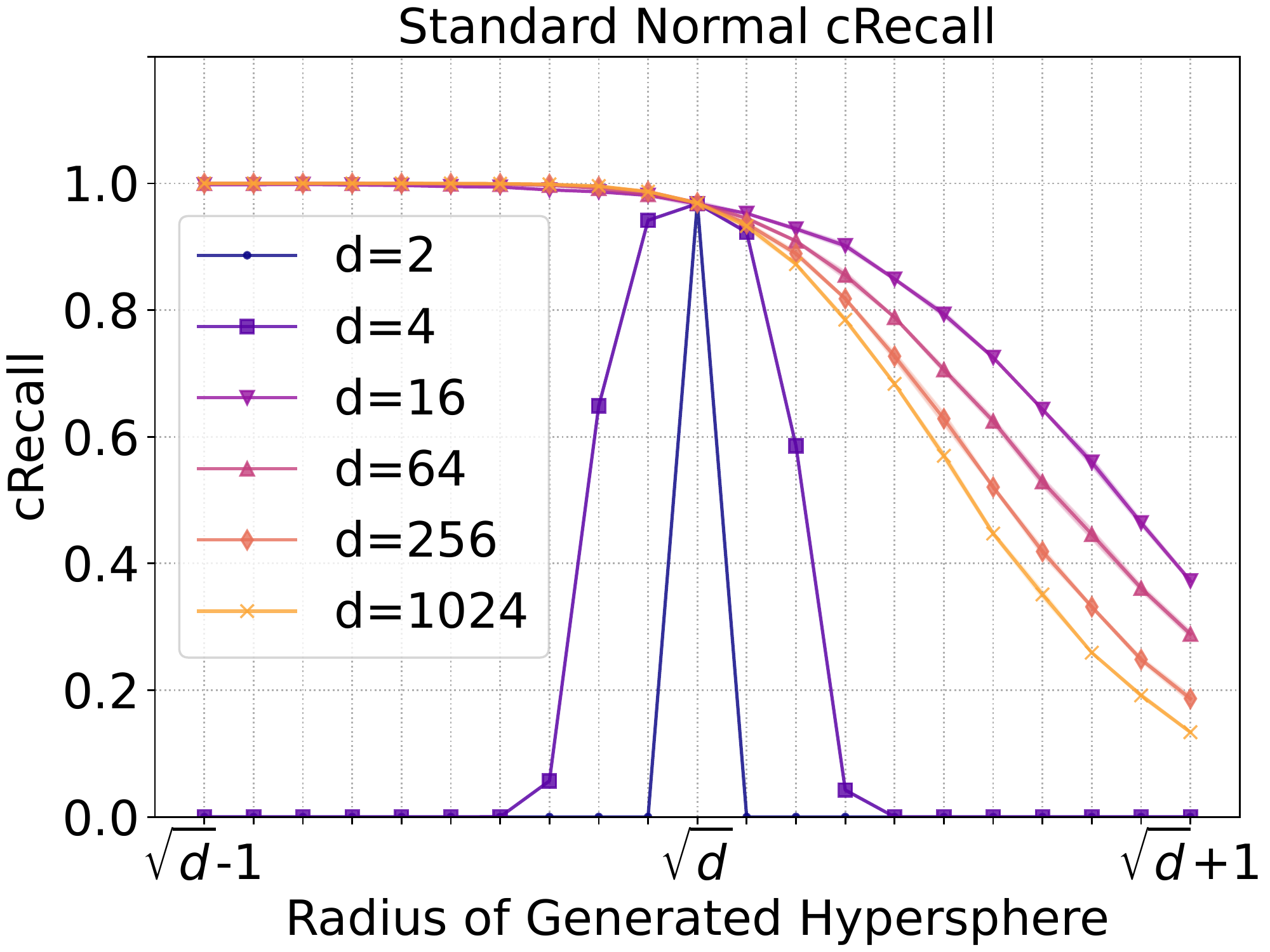}
        \label{fig:normalsphere_r1_pr_cr}}
    \subfloat[symPrecision (Gaussian)]{
        \centering
        \includegraphics[trim=64 0 0 25, clip, width=0.235\textwidth]{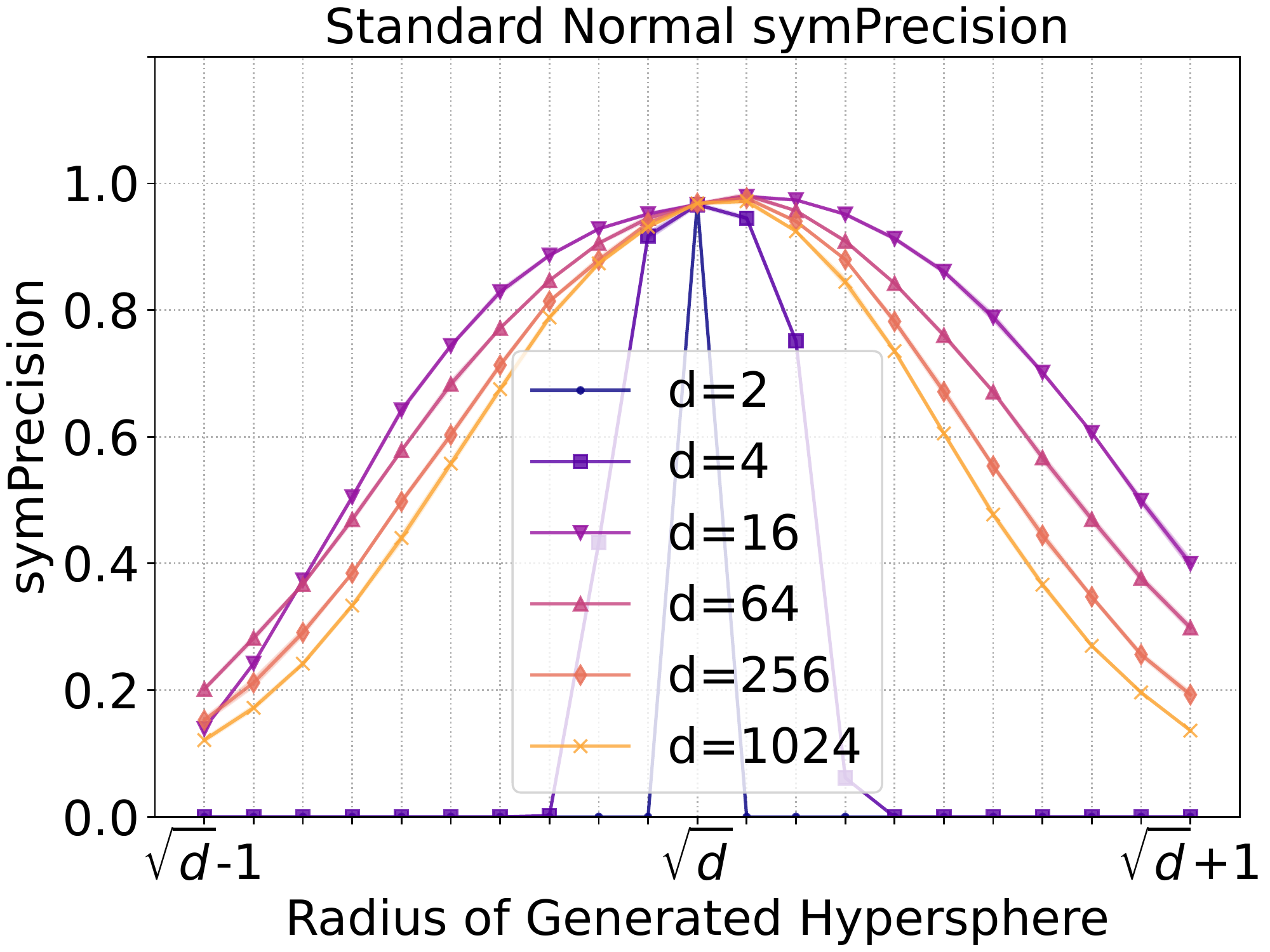}
        \label{fig:normalsphere_r1_pr_sp}}
    \subfloat[symRecall (Gaussian)]{
        \centering
        \includegraphics[trim=64 0 0 25, clip, width=0.235\textwidth]{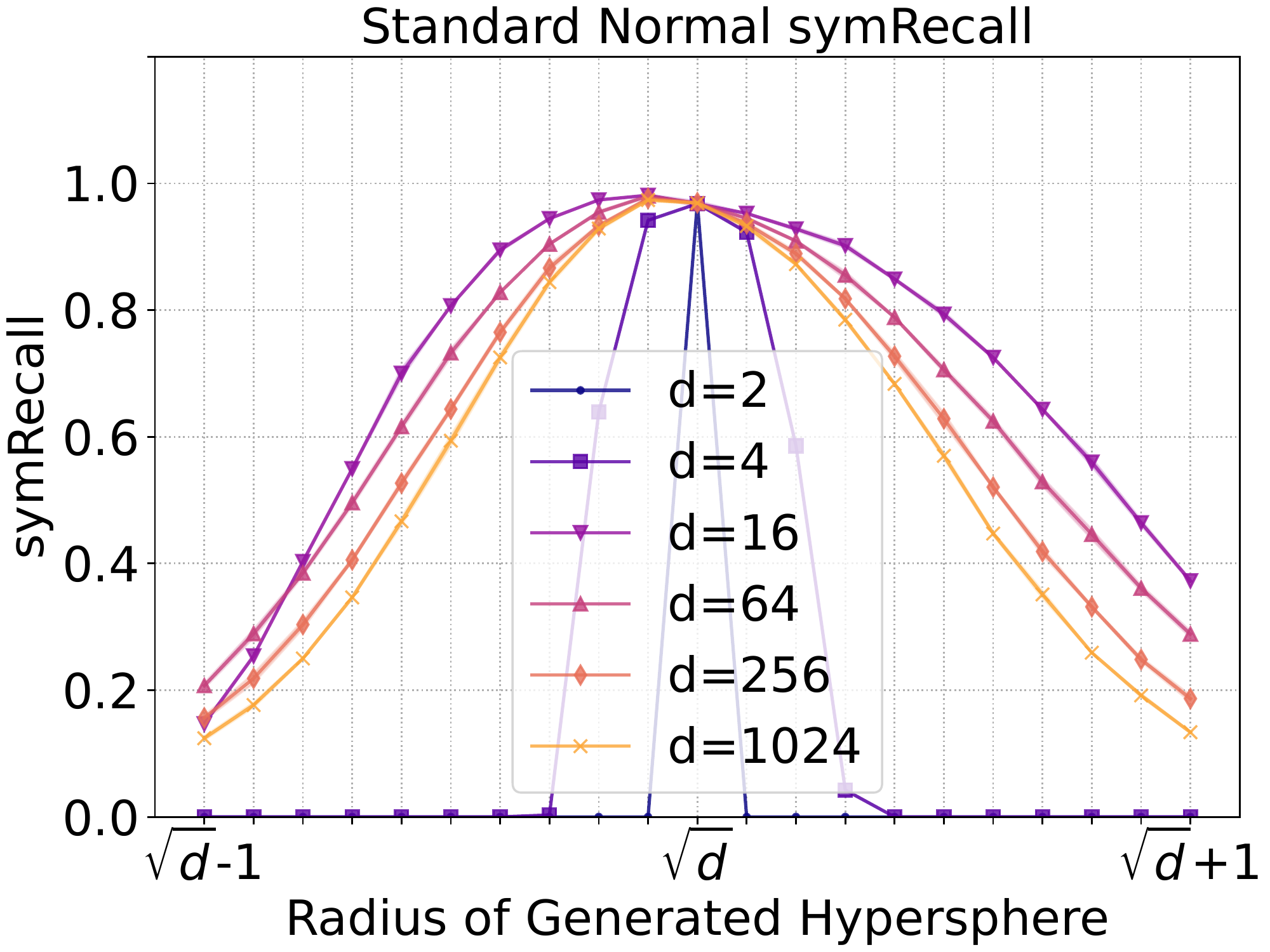}
        \label{fig:normalsphere_r1_pr_sr}}
    \caption{symPrecision and symRecall exhibit symmetric behavior inside or outside the reference support, regardless of the number of dimensions, in contrast to the asymmetry observed in Precision and Recall, and their complements (cPrecision and cRecall). (a, b, c, d) The reference support is the hypersphere of radius 1. (e, f, g, h) The reference support is the hypercube of half-edge-length 1. (i, j, k, l) The reference support is hypersphere at $\sqrt{\textrm{d}}$, following the support of Gaussian distributions.}
    \label{fig:sym_metrics}
\end{figure*}
\section{Symmetric Precision and Recall}
\label{sec:symmetricpr}
As we observed in Section~\ref{sec:theory}, the direction of the asymmetry in Precision and Recall is connected to which distribution's support is approximated with the $K$-nearest-neighbors. Following up on this observation, we can consider \textit{complement} versions of Precision and Recall, denoted \textit{cPrecision} and \textit{cRecall}, where we keep everything in the respective formulas the same except for which distribution is approximated, arriving at the following definitions:
\begin{align}
    \begin{split}
        \textrm{cPrecision} &(p_r, p_g) = \Px{p_g}{S_r \cap \hat{S_g}} \\
        &\approx \frac{1}{|X_g|} \sum_{x_i \in X_g} 1(N_K(x_i) \ni\joinrel\in X_r)
    \end{split}\\
    \begin{split}
        \textrm{cRecall} &(p_r, p_g) = \Px{p_r}{\hat{S_r} \cap S_g}\\
        &\approx \frac{1}{|X_r|} \sum_{x_i \in X_r} 1(N_K(x_i) \ni\joinrel\in X_g)
    \end{split}
\end{align}
where $\ni\joinrel\in$ denotes non-empty intersection, $1(.)$ is the indicator function, $X_r$ and $X_g$ are sets of samples from $p_r$ and $p_g$, $S_r$ and $S_g$ are their respective supports, $N_K(x_i)$ denotes the $K$-nearest-neighbors neighborhood of $x_i$, and hat denotes approximation of support by $K$-nearest-neighbors. In other words, $\textrm{cPrecision}$ measures the fraction of generated samples whose neighborhoods each contains at least one real sample, and $\textrm{cRecall}$ measures the fraction of real samples whose neighborhoods each contains at least one generated sample. The latter has been previously proposed by~\citet{naeem2020density-coverage} under the name Coverage~\footnote{We use cRecall instead of Coverage in our discussions to emphasize its complement nature to Recall.}, to address the problem of sensitivity of Recall to outliers in generated data, while we are not aware of the former being presented in prior works. However, in this discussion, we are interested not in their ability to overcome outliers, rather their asymmetry in high dimensions: as shown in~\cref{fig:sym_metrics}, both cPrecision and cRecall exhibit asymmetry in high dimensions similar to Precision and Recall, but with a crucial twist: their asymmetrical behavior is inverted. cPrecision vanishes when $p_g$ is inside $p_r$ and saturates when outside (opposite the trend in Precision we observed in~\cref{fig:cube_r1_pr}). Conversely, cRecall saturates when $p_g$ is inside $p_r$ and vanishes when outside (opposite the trend in Recall we observed in~\cref{fig:cube_r1_pr}).

As such, while cRecall and cPrecision cannot fix the asymmetry in high dimensions, their combination with Precision and Recall might. To that end, we can naturally define a more symmetric Precision and Recall by taking the minimum of the corresponding pairs of metrics, which we denote \textit{symPrecision} and \textit{symRecall}:
\begin{align}
    \textrm{symPrecision} &= \min(\textrm{cPrecision}, \textrm{Precision})\\
    \textrm{symRecall} &= \min(\textrm{cRecall}, \textrm{Recall})
\end{align}

The reasoning behind the choice of $\min$ is as follows. We want to convert an asymmetric metric, say $f(x)$, into a symmetric one, say $h(x)$, while maintaining its semantics (here $x$ is a scalar representing the expansion/contraction of the generated support such that at $x=0$ it is equal to the real manifold). To do so, we designed a complement metric $g(x)$ with two properties: first, having the same semantics of diversity or fidelity as $f$ (\eg Precision and cPrecision both measure fidelity); second, having the inverted asymmetry of $f$, that is, $g(x) = f(-x)$. Now, choosing $h(x)=\min(g(x), f(x))$ makes $h$ readily symmetric, $h(x)=h(-x)$. Instead of $\min$ we could use any function that is invariant under permutations of its variables, however, we chose $\min$ because it can maintain the semantics of $f$ and $g$ by gating between them.

As expected, in~\cref{fig:sym_metrics} we observe that these metrics behave more symmetrically in all experiments, regardless of the number of dimensions. Note that since the proposed metrics simply take the minimum, they do not violate the intended intuition of fidelity and diversity assigned to Precision and Recall, rather extend it to when the supports do not exactly match in high dimensions. In particular, the insensitivity to outliers is maintained since outliers can only artificially inflate Precision or Recall, in which case the minimum will result in the use of cPrecision and/or cRecall which are more robust to outliers~\cite{naeem2020density-coverage}. Additionally, when the generated and real supports match, since asymptotically (in the number of samples) Precision and Recall are equal to their complements, the symmetric Precision and Recall will also converge to the same asymptotic values. So far, all our empirical evidence have been restricted to synthetic data. In the next section, we will explore whether the asymmetry also emerges in practice.

\begin{figure*}[h!]
    \centering
    \subfloat[Fidelity (CelebA)]{
        \centering
        \includegraphics[trim=25 0 0 25, clip, width=0.253\textwidth]{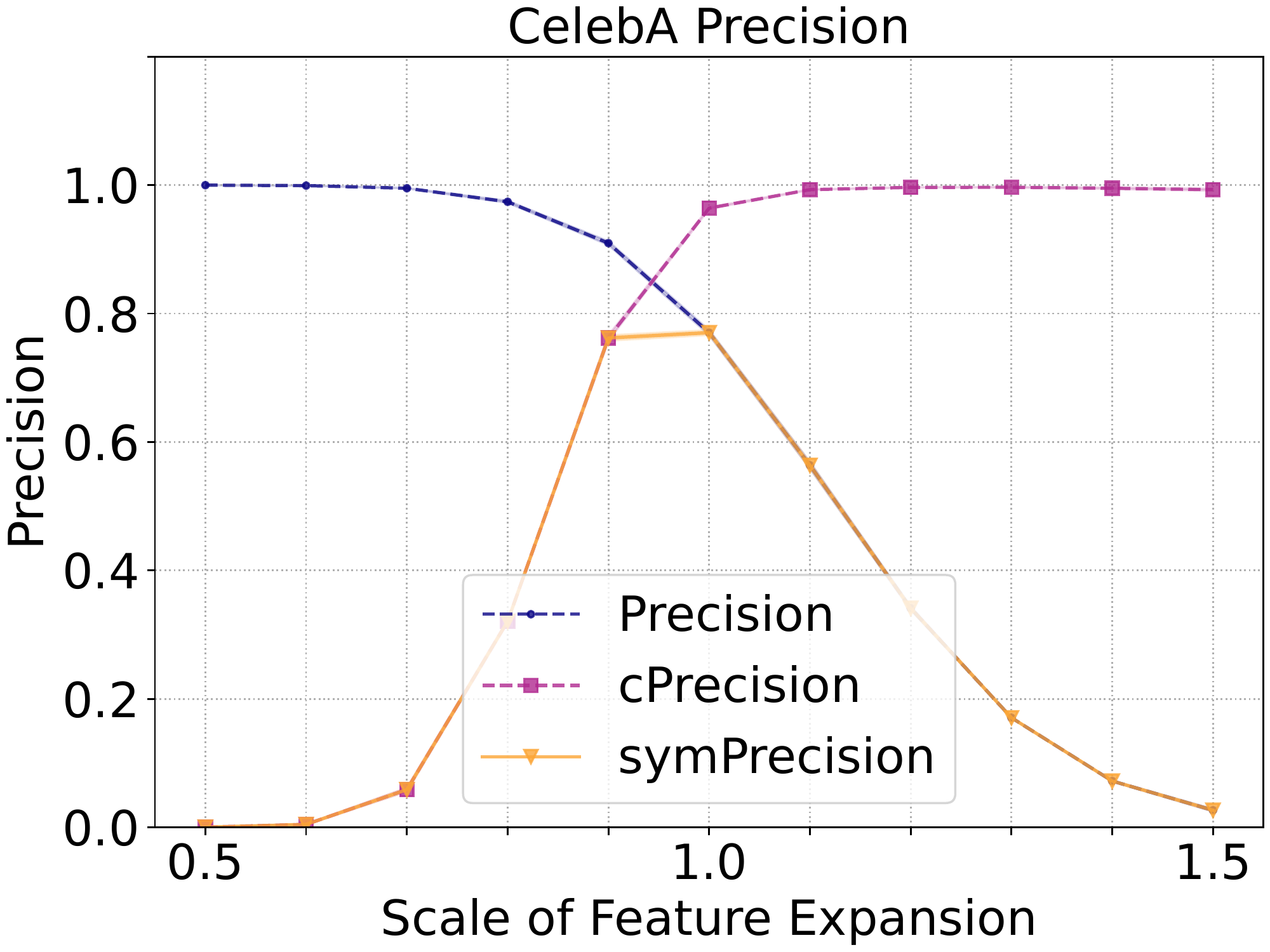}
        \label{fig:scale_celeba_p}}
    \subfloat[Diversity (CelebA)]{
        \centering
        \includegraphics[trim=64 0 0 25, clip, width=0.235\textwidth]{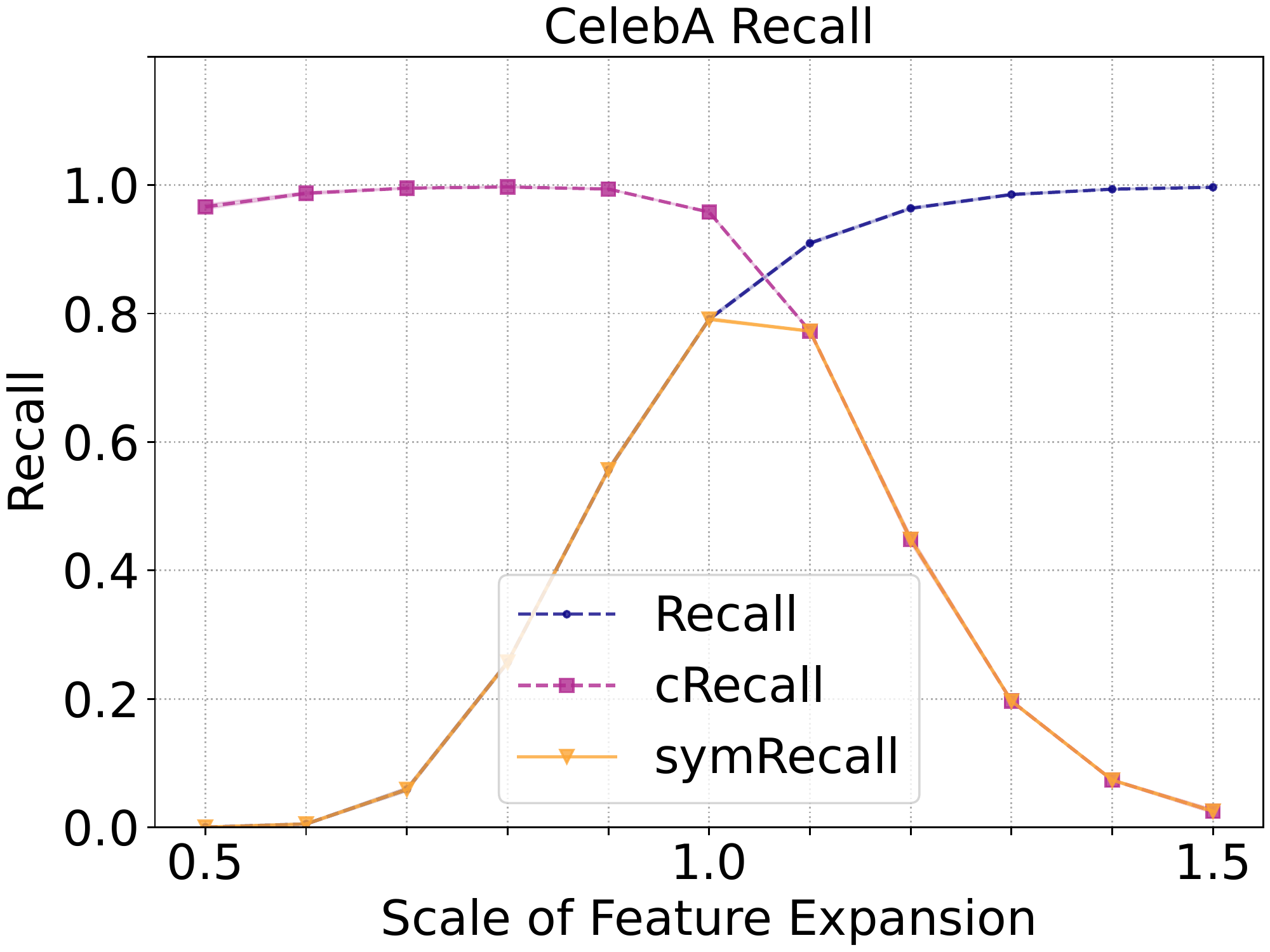}
        \label{fig:scale_celeba_r}}
        \subfloat[Fidelity (CIFAR10)]{
        \centering
        \includegraphics[trim=64 0 0 25, clip, width=0.235\textwidth]{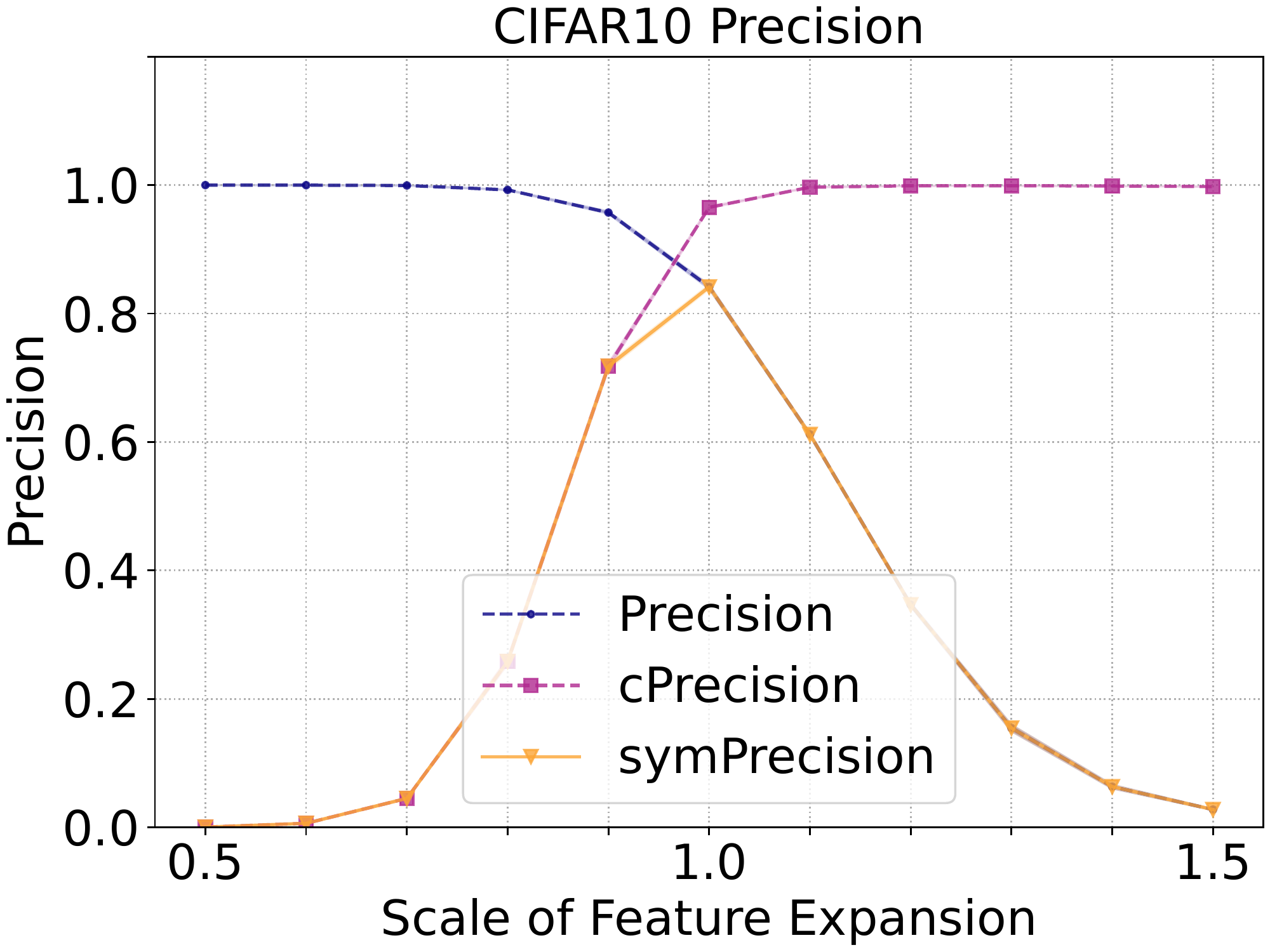}
        \label{fig:scale_cifar_p}}
    \subfloat[Diversity (CIFAR10)]{
        \centering
        \includegraphics[trim=64 0 0 25, clip, width=0.235\textwidth]{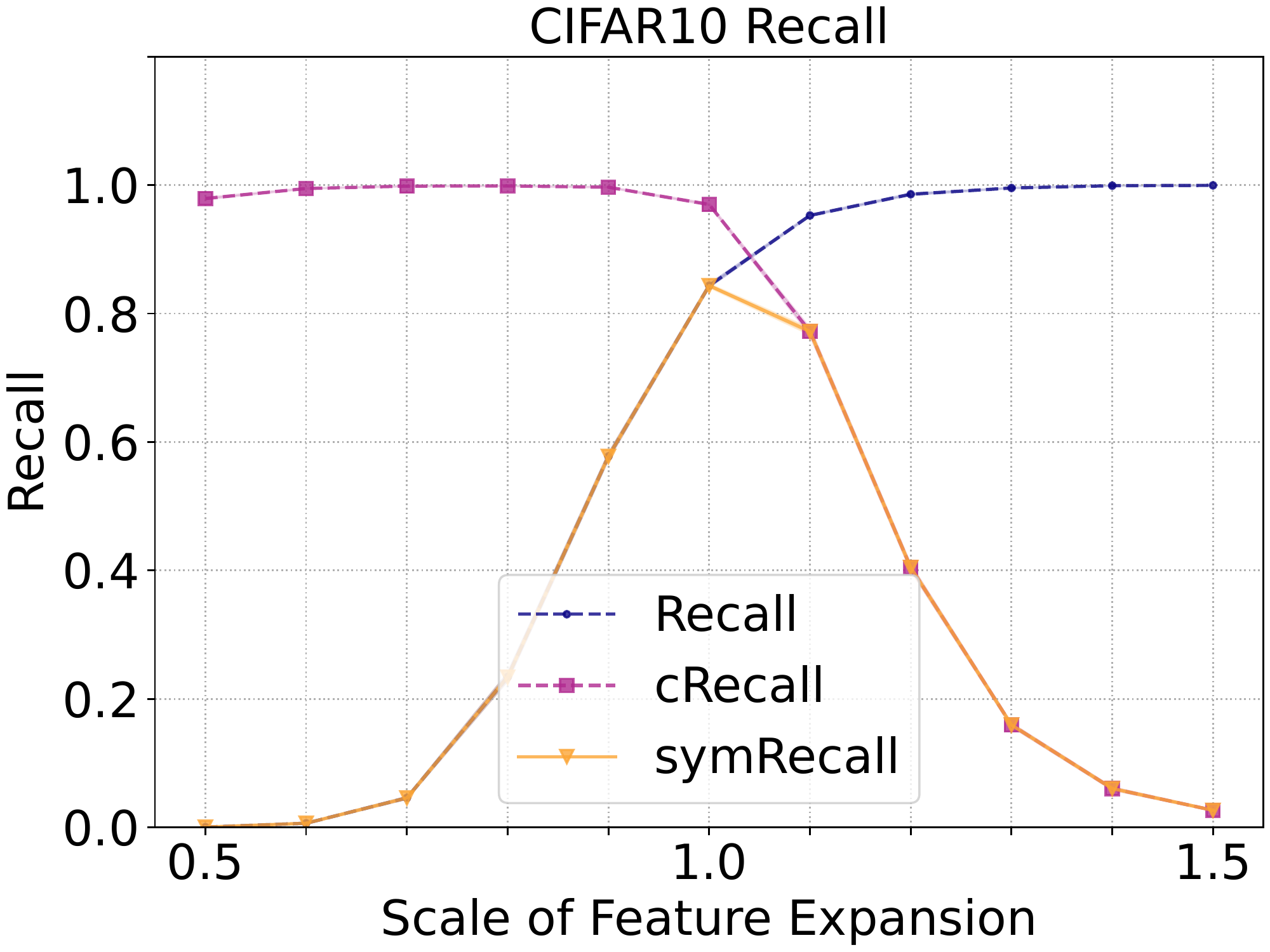}
        \label{fig:scale_cifar_r}}
    \caption{The effect of scaling the feature space of CelebA and CIFAR10 images with various scaling factors, where scaling of 1 will be the same as no scaling (\ie the generated and reference supports become equal). While Precision and Recall, and their complements, all exhibit asymmetric behavior, symPrecision and symRecall can achieve symmetric behavior.}
    \label{fig:scale_metrics}
\end{figure*}
\begin{figure*}[h!]
    \centering
    \subfloat[Fidelity (CelebA)]{
        \centering
        \includegraphics[trim=25 0 0 25, clip, width=0.253\textwidth]{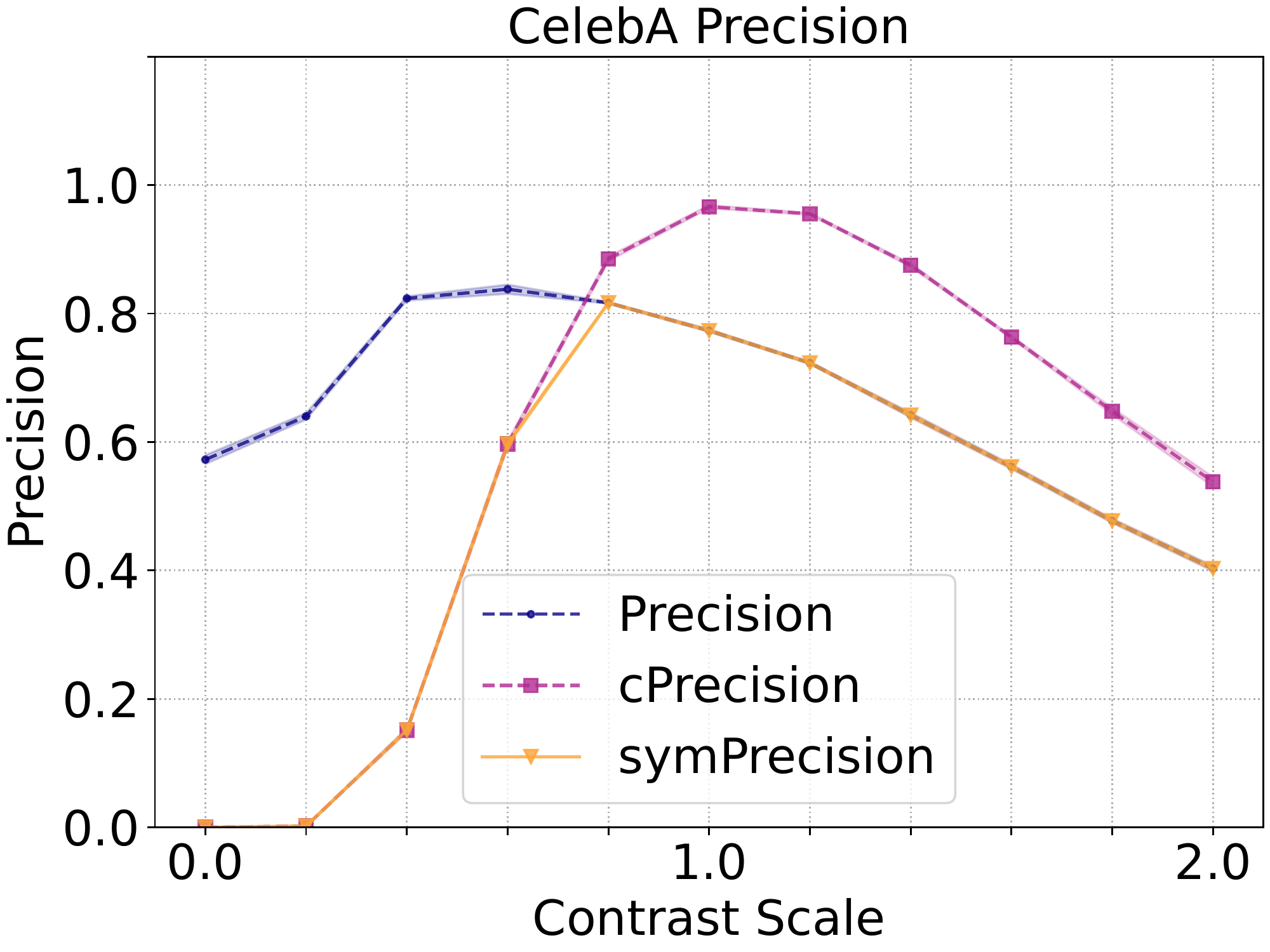}
        \label{fig:contrast_celeba_p}}
    \subfloat[Diversity (CelebA)]{
        \centering
        \includegraphics[trim=64 0 0 25, clip, width=0.235\textwidth]{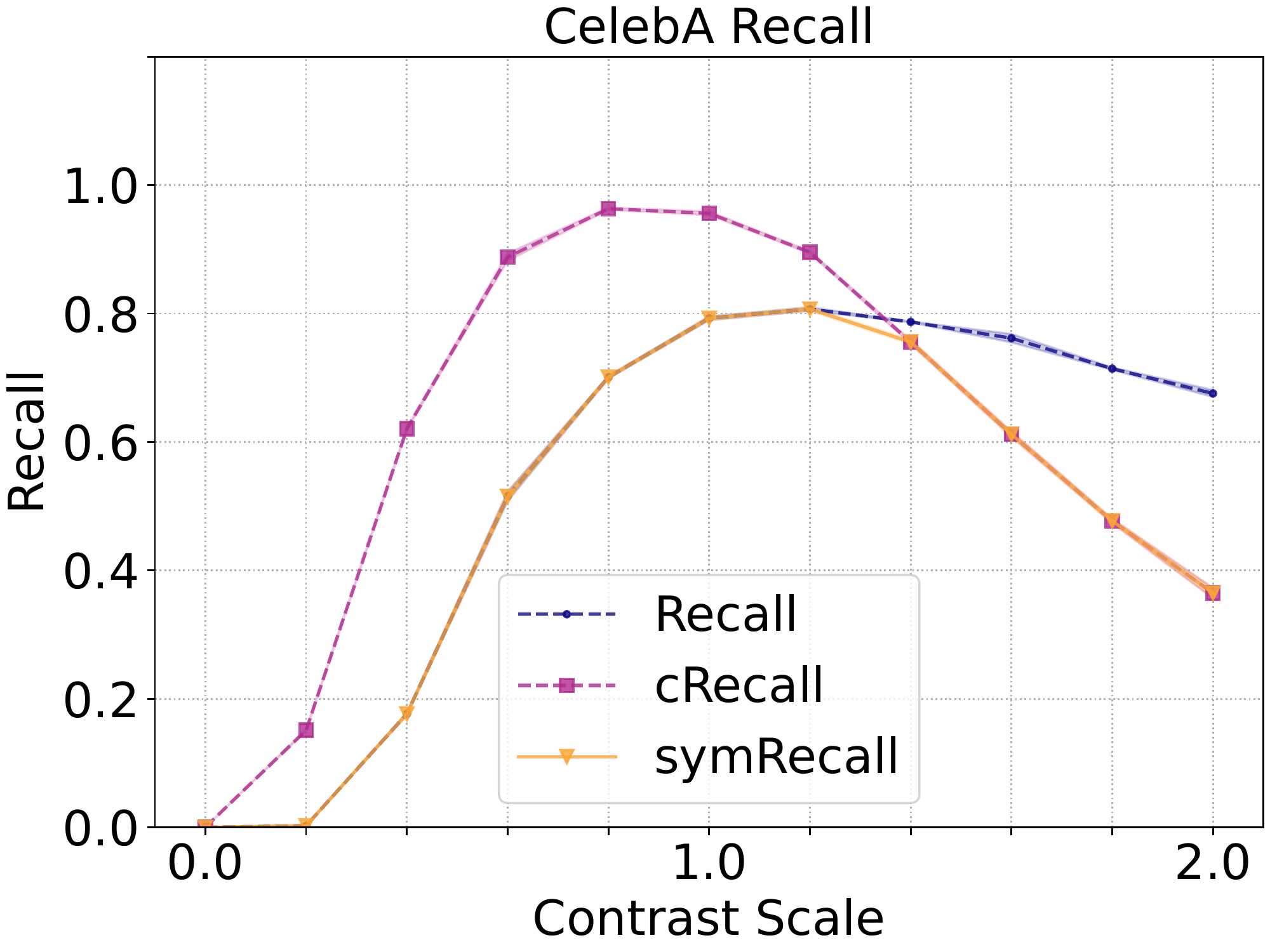}
        \label{fig:contrast_celeba_r}}
        \subfloat[Fidelity (CIFAR10)]{
        \centering
        \includegraphics[trim=64 0 0 25, clip, width=0.235\textwidth]{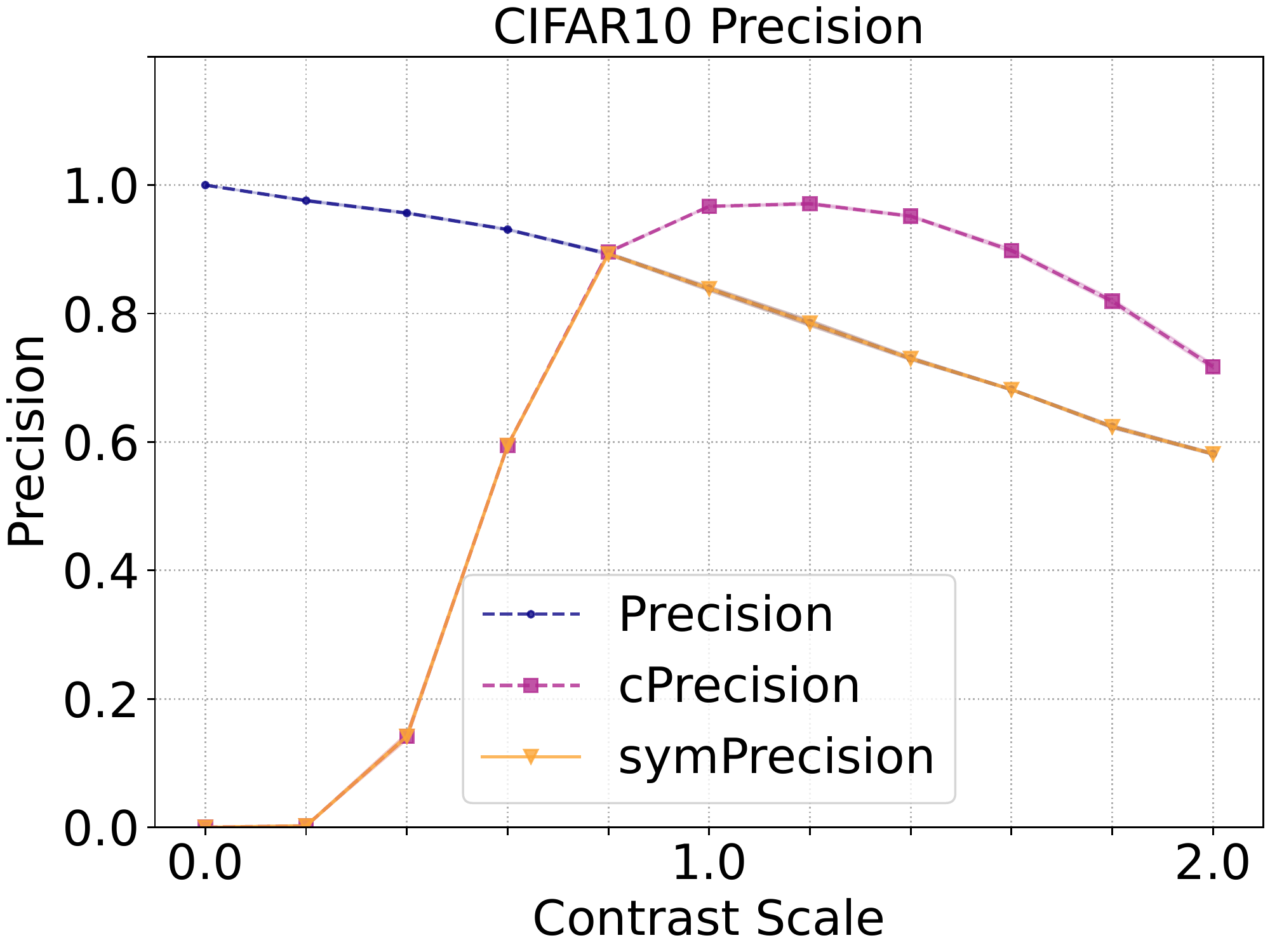}
        \label{fig:contrast_cifar_p}}
    \subfloat[Diversity (CIFAR10)]{
        \centering
        \includegraphics[trim=64 0 0 25, clip, width=0.235\textwidth]{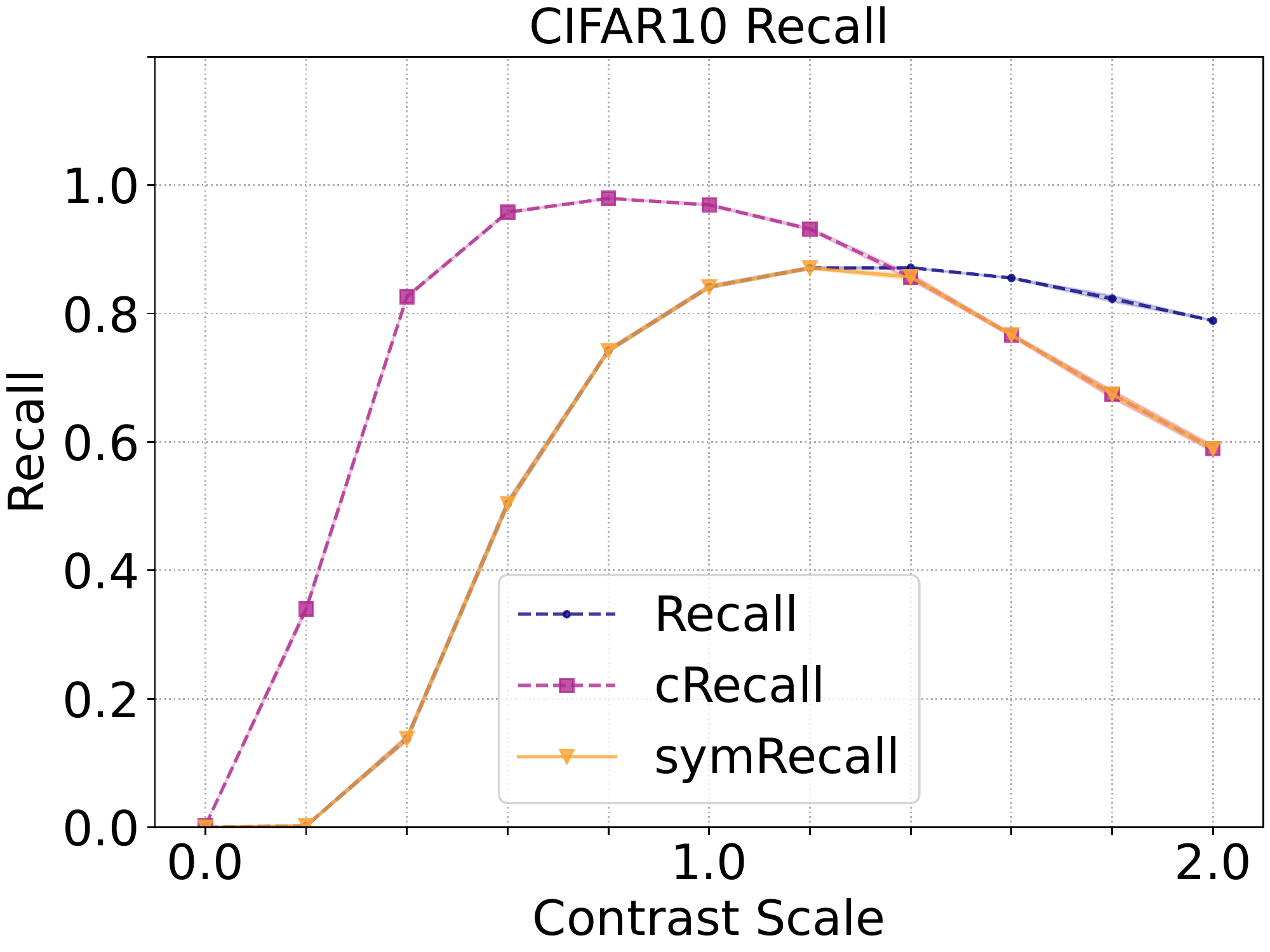}
        \label{fig:contrast_cifar_r}}
    \caption{The effect of increasing the contrast of images of CelebA and CIFAR10 with various scaling factors, where scaling of 1 will be the same as no scaling (\ie the generated and reference supports become equal). While Precision and Recall, and their complements, all exhibit asymmetric behavior, symPrecision and symRecall can achieve more symmetric behavior (notice the faster decaying tails on both sides for symPrecision and symRecall). The symmetry not being ideal is due to the fact that increasing/decreasing contrast might not move the embedded image manifold with equal rates outward and inward.}
    \label{fig:contrast_metrics}
\end{figure*}
\section{Real Data Experiments}
\label{sec:exp}
In this section, we provide two experiments to study the existence of the emergent asymmetry of Precision and Recall in real-world datasets. In these experiments, we use images from CelebA~\cite{liu2015celeba} at $128\times128$ resolution, and images from CIFAR10~\cite{krizhevsky2009cifar} at $32\times32$ resolution, and the VGG16~\cite{simonyan2014vgg} pretrained on ImageNet~\cite{deng2009imagenet} to encode the images into embedding space as suggested by~\citet{kynkaanniemi2019improved-precision-recall}. We also consider an alternative random embedding space in~\cref{sec:random_feat}. In all experiments, both with synthetic data and real data, we use 10,000 random samples from each of the real and generated distributions to compute the metrics, and repeat all experiments five times and report average values with one standard deviation above and below the average (the standard deviation is regularly very small and not discernible in the figures). Additionally, we note that the authors of Precision and Recall suggested $K=3$ in~\cite{kynkaanniemi2019improved-precision-recall}, whereas follow up work~\cite{naeem2020density-coverage} suggested $K=5$ for Coverage (cRecall). We experimented with both values, and observed no significant change in the reported behavior, and therefore chose to report all experiments using $K=5$ in favor of consistency. The code and experiments will be available at~\url{https://github.com/mahyarkoy/emergent_asymmetry_pr}.

\subsection{Scaling the Feature Space}
\label{sec:exp_scale}
In this experiment, we directly scale the feature space of embedded images, in order to contract and expand the image manifold in the feature space. More specifically, for each set of embedded images using $\phi$, $\{\phi(x_i)\}_{i=1}^N$, we compute the sample mean $\hat{\phi}$, and scale each sample along the direction of the mean, that is, $\phi_s(x) = s (\phi(x) - \hat{\phi}) + \hat{\phi}$, where $s\in[0.5, 1.5]$ is the scaling factor. Each value of $s$, together with a random subset from the training set of the respective datasets embedded with VGG16, is treated as a distinct generative model, whereas the real distribution is the embedded testing set of the respective datasets.

In~\cref{fig:scale_metrics}, as we change the scale from 0.5 to 1.5, passing the real distribution's support at $s=1$, we again observe the asymmetric behavior in Precision and Recall, that is, being inside the real support makes a generative model appear as if it is generating samples of much higher fidelity compared to being on the outside, and vice versa for Recall. In contrast, our proposed metrics, symPrecision and symRecall, exhibit the expected symmetrical behavior, consistent with the interpretation of fidelity and diversity. Note that the image distributions in embedding space have much more complicated supports compared to the synthetic distributions we considered in~\cref{sec:example}, which shows the existence of the asymmetry in real-world distributions.

\subsection{Varying Image Contrast}
\label{sec:exp_contrast}
In the previous experiment, we directly scaled the image manifold in the embedding space and observed the asymmetry in metrics, however, the question remains whether the feature space scaling can be realized by actual changes in the image space, or it is a pathological modification unlikely to occur during image generation. The challenge here is that if we use trained generative models, it is unclear how to rigorously determine when models are moving the image support outside/inside the real image support, versus when models are genuinely generating samples that are of high/low fidelity and diversity. To make sure images are actually being moved away (inwards or outwards) from the real image support, we consider a family of generative models that generate by applying a fixed contrast scale $s\in[0, 2]$ to randomly chosen images from the training set of real datasets. The real dataset is considered as the testing set of the dataset under study (with no modification to the contrast). With this family of models (each model applying a distinct contrast scale $s$), we can be sure that by increasing and decreasing contrast away from $1$, the generative models' supports are moving outward and inward from the real image support, respectively.

\begin{figure}[t!]
    \centering
    \includegraphics[trim=0 0 0 0, clip, width=0.45\textwidth]{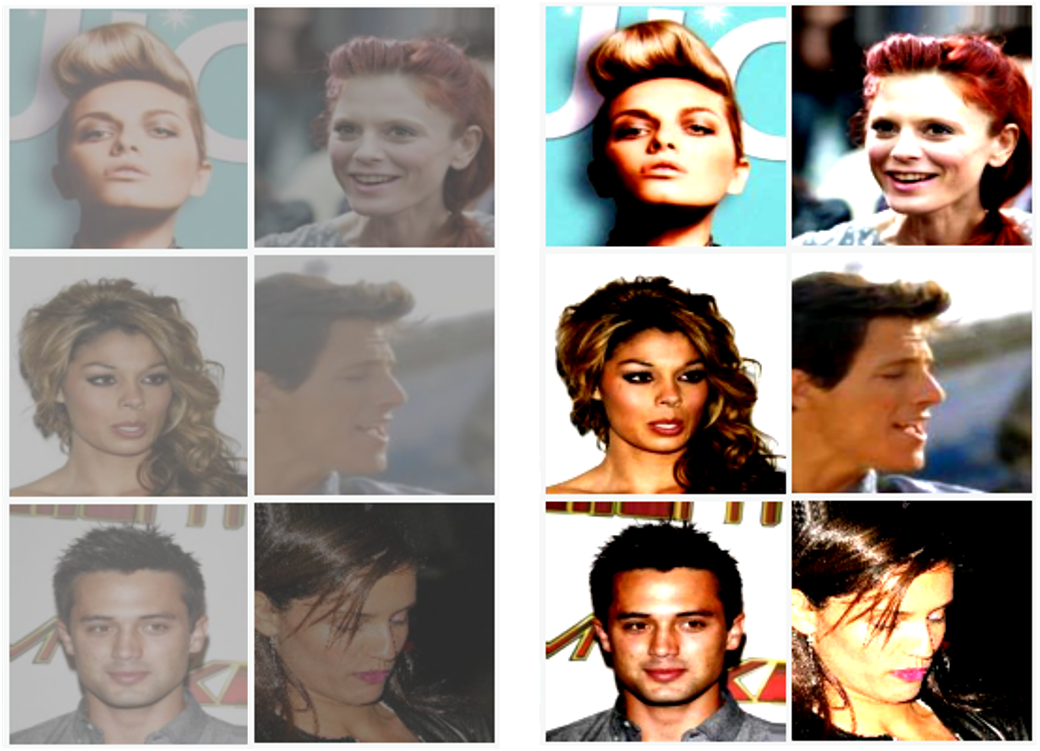}
    \vskip 0.5em
    \begin{tabular}{rl@{ $\rightarrow$ }l@{\hspace{0.9in}}}
        \textbf{Contrast} & $0.4$ & $2.0$ \\
        \cmidrule[1pt](l{0.5em}r{1em}){1-3}
        Precision & $0.82$ {\tiny$\pm0.00$} & $0.40$ {\tiny$\pm0.00$} \\
        Recall & $0.18$ {\tiny$\pm0.00$} & $0.68$ {\tiny$\pm0.01$} \\
        \cmidrule[0.5pt](l{0.5em}r{1em}){1-3}
        symPrecision & $0.15$ {\tiny$\pm0.00$} & $0.40$ {\tiny$\pm0.00$} \\
        symRecall & $0.18$ {\tiny$\pm0.00$} & $0.36$ {\tiny$\pm0.01$}\\
        \cmidrule[1pt](l{0.5em}r{1em}){1-3}
    \end{tabular}
    \caption{Visualizing examples from the contrast varying experiment, where changing the contrast results in a rapid saturation/vanishing in Precision and Recall, consistent with the emergence of asymmetry in high dimensions. The images on the left are deemed to be of much higher fidelity (higher Precision) than the images to the right, which is misleading. Also, The images on the left are considered as capturing less of the diversity of real images than the ones on the right, which again is misleading. Proposed symmetric metrics substantially reduce this unexpected gap, keeping both metrics under 0.5 for both left and right images.
    }
    \label{fig:celeba_contrast}
\end{figure}

In~\cref{fig:contrast_metrics}, we observe the asymmetric behavior in Precision and Recall, showing that it is not only possible, but practical, for a family of generative models to generate samples that have unfairly high/low Precision or Recall. For example, in~\cref{fig:contrast_celeba_p,fig:contrast_cifar_p}, samples of $s=0.4$ contrast, which are washed out, always achieve higher Precision than samples of $s\geq 1$ which are of much higher fidelity (see samples in~\cref{fig:celeba_contrast}). Similarly, in~\cref{fig:contrast_celeba_r,fig:contrast_cifar_r}, samples of $s=2$, which are extremely over-exposed, always receive higher Recall than samples of $s=0.4$, although both have the same amount of meaningful diversity. In all cases, the use of our proposed metrics alleviate these issues and results in a more symmetric behavior (\ie faster decaying tails as $s$ approach $0$ and $2$). The remaining asymmetry can be explained by noting that the sensitivity of the embedding network (VGG16) to contrast increase and decrease is not necessarily exactly the same.



\section{Related Works}
\label{sec:related_works}
\textbf{Heuristic Metrics of Generative Performance}. Comparing two sets of images in terms of quality has long been of interest due to its application to compression. Most notable classical methods for this task are PSNR and SSIM which compare images directly in terms of differences in pixel values~\cite{hore2010ssim}, and remain useful to date, in particular to detect whether a generative model is memorizing training samples~\cite{karras2020stylegan-ada}. Divergences between distributions have also been traditionally used to compare generative models, however, since computing likelihood is often intractable in high dimensions, direct use of divergences remain useful mostly in low dimensional setting. A notable example is Inception Score~\cite{salimans2016inception-score} which uses a classifier to construct a tractable likelihood for a set of images over a finite number of class categories, which can then be compared with another set of images using KL divergence. Several other heuristics have also proven useful in probing specific aspects of generative performance: comparing the support size of generative models based on human-guided detection of duplicates~\cite{arora2018birthday}, measuring the amount of high frequency artifacts in generative models by comparing their average spectra~\cite{dzanic2019fourier, khayatkhoei2022spatial}, comparing the linear separability of generative models' latent spaces and the smoothness of the mapping to image space -- denoted Perceptual Path Length -- as a surrogate for having learnt the correct generative model~\cite{karras2018style, karras2020stylegan2}, computing the accuracy of a classifier trained to distinguish generated samples from real ones as a measure of distribution mismatch, denoted Neural Net Divergence~\cite{arora2017nnd}, and using improvements from adding generated samples to downstream tasks as a measure of generalization~\cite{ravuri2019cas}. Despite each method having its own particular use cases, a shared limitation of them is not providing a direct way of disentangling the differences in diversity and fidelity.

\textbf{Moment-based Metrics}. These metrics compare two sets of images by estimating and comparing their moments in a predetermined feature space. Most notably, Frechet Inception Distance~\cite{heusel2017fid}, maps the two image sets into the latent space of an Inception Net~\cite{szegedy2016inceptionv3} pretrained on ImageNet~\cite{deng2009imagenet}, and compare their first and second moments. Kernel Inception Distance~\cite{bińkowski2018kid} allows for comparing higher moments in the same latent space. Moment-based methods, similar to heuristic methods, cannot distinguish lack of diversity from lack of fidelity. In order to address this drawback, manifold-based metrics were introduced.

\textbf{Manifold-based Metrics}. These metrics compare two sets of images by estimating and comparing their support manifold in an embedding space, typically using VGG16~\cite{simonyan2014vgg} pretrained on ImageNet~\cite{deng2009imagenet}. The manifold of each set is estimated as the union of the $K$-nearest-neighbors balls centered at each data point, in a similar construction as Isomap~\cite{tenenbaum2000isomap}. How to compare the two estimated manifolds results in various metrics. Improved Precision and Recall, compute the fraction of model samples that fall in the data manifold and the fraction of data samples that fall in the model manifold, respectively -- these methods were originally proposed to improve the estimation of similar concepts proposed by~\citet{sajjadi2018precision-recall}.~\citet{naeem2020density-coverage} discovered a sensitivity to outliers in Precision and Recall, and therefore proposed modifications denoted Density and Coverage, where Density measures the average number of real data neighborhoods that cover any generated data sample normalized by the neighborhood's expected size ($K$ of $K$-nearest-neighbors), and Coverage measures the fraction of generated data samples whose neighborhood contain any real data sample. More recently,~\citet{alaa2022alphabetapr} proposed $\alpha$/$\beta$ Precision and Recall, which aimed to generalize these metrics such that instead of comparing the whole supports of real and generated data, their supports are first partitioned into different levels of likelihood, and then Precision and Recall are computed for each pair of the partition. This would allow a more granular comparison between distributions, and also addresses the problem of sensitivity to outliers. However, to compute the partitions, they use a trained embedding network to maximally squeeze the data manifold into a sphere, which raises the possibility of distorting the semantic meaning of distances in the data manifold, that is, some distances could arbitrarily collapse.

\textbf{The Curse of Dimensionality}. Our analysis is a particular manifestation of the curse of dimensionality. The effect of growing dimensions on distance concentration and meaningfulness of nearest neighbors has been extensively explored in classical settings, specially in the context of kernels~\cite{franccois2007concentration, evangelista2006taming, aggarwal2001surprising, beyer1999nearest}. In particular, the fact that the ratio of distance variance to distance to mean of i.i.d. random variables vanishes with increasing number of dimensions, and consequently distances between all points appear similar, which will break down the utility of many kernel based estimations. A phenomenon very closely related to our analysis in this paper, is the emergence of hubs in high dimensional Gaussian distributions. Hubs refer to points that are close to a very large number of other samples from the Gaussian distribution under a $K$-nearest-neighbors notion of closeness, much larger than the average number of neighborhoods that contain any point~\cite{radovanovic2010hubs}. Our theoretical analysis extends the known reach of the curse of dimensionality, by showing how it affects the overlap between distributions supported on hyperspheres.
\section{Conclusion}
\label{sec:discussion}
In this work, we identified a critical flaw in the common approximation of Precision and Recall using $K$-nearest-neighbors, denoted emergent asymmetry: in high dimensions, moving the generated distribution slightly outwards or inwards away from the real distribution's support can lead to vastly different values of Precision and Recall in each direction. We proved this asymmetry for distributions supported on hyperspheres, and empirically showed its emergence in synthetic and real-world datasets. We also proposed modifications to Precision and Recall to reduce the effect of the asymmetry in high dimensions. Our findings suggest several interesting directions for future research. First, as we conjectured in~\cref{sec:theory} and observed in the experiments of~\cref{sec:exp}, the asymmetry is not restricted to distributions with hyperspherical supports; identifying the necessary assumptions on a space for the emergence of the asymmetry is a valuable future direction. Second, while we proved the asymmetry asymptotically, deriving bounds in the finite case would provide more granular insights. Finally, while we showed the existence of the emergent asymmetry in the widely-used Improved Precision and Recall metrics, the extent to which it affects other metrics of generative performance remains to be explored.

\section*{Acknowledgements}
We wish to thank Joe Mathai for maintaining our compute cluster, and the anonymous reviewers for their helpful comments and suggestions.
This research is based upon work supported in part by the Office of the Director of National Intelligence (ODNI), Intelligence Advanced Research Projects Activity (IARPA), via [2022-21102100007]. The views and conclusions contained herein are those of the authors and should not be interpreted as necessarily representing the official policies, either expressed or implied, of ODNI, IARPA, or the U.S. Government. The U.S. Government is authorized to reproduce and distribute reprints for governmental purposes notwithstanding any copyright annotation therein.

\bibliography{refs}
\bibliographystyle{icml2023}

\newpage
\appendix
\onecolumn
\section{Proof of~\cref{proposition:inside}}
\label{sec:proof_inside}
\begin{proof}
We start by making two redundant assumptions which we will relax later: 1) the distributions $p_g$ and $p_r$ are uniformly distributed over their respective supports $S_g$ and $S_r$; and 2) we are constructing the approximate support using first nearest neighbor ($K=1$). Note that $S_g$ is the $d$ dimensional ball $B$ of radius $R$, $S_r$ is the $d-1$ dimensional hypersphere at its boundary $\partial B = S_r$, and $\hat{S}_r$ denotes the $K$-nearest-neighbors approximation of $S_r$ using a set of $N$ i.i.d. observations $X_r=\{x_i\}_{i=1}^N$ from $p_r$, that is, $\hat{S_r} = \cup_{i=1}^N N_K(x_i)$ where $N_K(x_i)$ is the $d$-ball centered at $x_i$ with radius equal to the Euclidean distance of $x_i$ from its $K$-th nearest neighbor in $X_r$.

The direction of the proof is as follows: we consider the event that there exists a sample from $p_r$ with its Euclidean nearest neighbor distance greater or equal to $R\sqrt{2}$, that is:
\begin{align}
    \{ \exists x^* \in X_r: \min_{x\neq x^*, x\in X_r} ||x^*-x|| \geq R\sqrt{2} \}
\end{align}
and show that under this event, the support of $p_g$ will be completely covered by the approximate support of $p_r$ as $d \rightarrow \infty$. Then, we compute the probability of this event under $p_r$ and observe that it saturates with $d \rightarrow \infty$, concluding the proof.

At a distance $R\sqrt{2}$, the neighborhood of $x^*$ intersects $\partial B$ at its equator, therefore the volume of $B$ not covered by $N_K(x^*)$ is less than or equal to the volume of the hyperspherical cap of $B$ at $\phi<\frac{\pi}{2}$ (strictly less than $\frac{\pi}{2}$ because the overlap volume is strictly larger than the hemisphere's volume):
\begin{align}
    V(B \setminus N_K(x^*)) = V(B) - V(N_K(x^*) \cap B) \leq V(B^{cap}_{\phi<\frac{\pi}{2}})
\end{align}
where hyperspherical cap is defined as the smaller part when a hyperplane cuts a hypersphere of the same number of dimensions into two parts at a colatitude angle $0 < \phi \leq \frac{\pi}{2}$ (\eg $\phi=\frac{\pi}{2}$ would give the two hemispheres). The volume of this hyperspherical cap has the following closed form~\cite{li2011hypercap}:
\begin{align}
\label{eq:cap_volume}
     {V(B^{cap}_\phi}) = \frac{1}{2} V_d(R) I_{\sin^2 \phi}(\frac{d+1}{2}, \frac{1}{2})
\end{align}
where $I$ is the regularized incomplete beta function, and $V_d(R)$ is the volume of the $d$ dimensional ball of radius $R$. Additionally, we can approximate $I$ as follows for large $d$ (omitting higher order terms in $\sin \phi < 1$):
\begin{align}
\label{eq:beta_approx}
    I_{\sin^2 \phi}(\frac{d+1}{2}, \frac{1}{2}) = \frac{\beta(\sin^2\phi; \frac{d+1}{2}, \frac{1}{2})}{\beta(\frac{d+1}{2}, \frac{1}{2})} \approx \frac{\frac{(\sin^2\phi)^{\frac{d+1}{2}}}{\frac{d+1}{2}}}{\Gamma(\frac{1}{2}) (\frac{d+1}{2})^{-\frac{1}{2}}} = C \sqrt{\frac{(\sin \phi)^{2(d+1)}}{d+1}}
\end{align}
where $\beta$, $\Gamma$ are the beta (incomplete and complete) and gamma functions, respectively, the numerator is due to series expansion, and the denominator is due to Stirling's approximation. We use C to represent a non-unique constant value in our discussions (which might change between two different formulas). Now, we can compute the probability of $B \setminus N_K(x^*)$ with respect to $p_g$ as the ratio of volumes (since we assumed $p_g$ to be uniformly distributed on $B$):
\begin{align}
\label{eq:sr_cup_sg_bound}
    \Px{p_g}{B \setminus N_K(x^*)} = \frac{V(B) - V(N_K(x^*) \cap B)}{V(B)} \leq C \sqrt{\frac{(\sin \phi)^{2(d+1)}}{d+1}}
\end{align}

At this point we can relax the requirement that $p_g$ is uniform, by noting that any absolutely continuous probability measure on $B$ must be within a constant factor of the measure produced by the uniformly distributed measure (due to Radon-Nikodym theorem), therefore the above inequality holds regardless of the uniform assumption (albeit with different constants). Next, since $x^*$ is not the only sample used to construct the approximate manifold $\hat{S_r}$, and from~\cref{eq:sr_cup_sg_bound}, we have:
\begin{align}
    \Px{p_g}{\hat{S_r} \cap S_g} = \Px{p_g}{\hat{S_r} \cap B} \geq \Px{p_g}{N_K(x^*) \cap B} \geq 1 - C \sqrt{\frac{(\sin \phi)^{2(d+1)}}{d+1}}
\end{align}
where the right hand side tends to 1 as $d\rightarrow\infty$, so we arrive at:
\begin{align}
    \lim_{d\rightarrow\infty} \Px{p_g}{\hat{S_r} \cap S_g} = 1
\end{align}

So far we showed that under the event $\{\exists x^* \in X_r: \min_{x\neq x^*, x\in X_r} ||x^*-x|| \geq R\sqrt{2}\}$, the support of $p_g$ will be completely covered by the approximate support of $p_r$. What remains is to compute the probability of this event and observe how it behaves with $d$. First, note that to have one sample whose nearest neighbor is more than $R\sqrt{2}$ away, we must have at least one sample from which all the other samples are at least $R\sqrt{2}$ away, hence the following:
\begin{align}
    &\Px{X_r\sim p_r}{\exists x^* \in X_r: \min_{x\neq x^*, x\in X_r} ||x^*-x|| \geq R\sqrt{2}}\\
 &= \Px{X_r\sim p_r}{\exists x^* \in X_r, \forall x\neq x^* \in X_r : ||x^*-x|| \geq R\sqrt{2}}\\
 &= (1 - \Px{p_r}{||x^*-x|| <R\sqrt{2} })^{N-1}
\label{eq:event_prob}
\end{align}

Then for the inner event we have:
\begin{align}
\label{eq:proof_inside_event_pr_capratio}
    \Px{p_r}{||x^*-x|| <R\sqrt{2} }
    = \frac{A(B^{cap}_{\phi<\frac{\pi}{2}})}{A(S_r)}
\end{align}
since $p_r$ is uniformly distributed on $\partial B$, and $A(B^{cap}_{\phi<\frac{\pi}{2}})$ denotes the area of the hyperspherical cap of $B$ at colatitute angle $\phi<\frac{\pi}{2}$. This area has the following closed form~\cite{li2011hypercap}:
\begin{align}
    {A(B^{cap}_\phi}) = \frac{1}{2} A_d(R) I_{\sin^2 \phi}(\frac{d-1}{2}, \frac{1}{2})
\end{align}
where $I$ is the regularized incomplete beta function, and $A_d(R)$ is the area of the $d$ dimensional ball of radius $R$. Using the approximation mentioned in~\cref{eq:beta_approx}, we arrive at (omitting higher order terms in $\sin \phi < 1$):
\begin{align}
    \Px{p_r}{||x^*-x|| <R\sqrt{2} } \approx C \sqrt{\frac{(\sin \phi)^{2(d-1)}}{d-1}}
\end{align}

At this point we can also relax the requirement that $p_r$ is uniform, by once again noting that any absolutely continuous probability measure on $\partial B$ must be within a constant factor of the measure produced by the uniformly distributed measure, therefore the above equality becomes an inequality ($\leq$) and holds regardless of the uniform assumption (albeit with different constants). What remains is to apply series expansion to~\cref{eq:event_prob} (omitting higher order terms in $p_r$):
\begin{align}
    \Px{X_r\sim p_r}{\exists x^* \in X_r: \min_{x\neq x^*, x\in X_r} ||x^*-x|| \geq R\sqrt{2}} &= (1 - \Px{p_r}{||x^*-x|| <R\sqrt{2} })^{N-1}\\
    &\approx 1 - (N-1)\Px{p_r}{||x^*-x|| <R\sqrt{2}}\\
    &\geq 1 - C N \sqrt{\frac{(\sin \phi)^{2(d-1)}}{d-1}}
\end{align}

Now, we note that for $\lim_{d\rightarrow\infty} N\epsilon^{-d} = 0$ $\forall \epsilon > 1$, \ie number of samples less than exponential in $d$, the above tends to 1 as $d\rightarrow\infty$:
\begin{align}
\label{eq:proof_inside_event_pr_1}
    \lim_{d\rightarrow\infty} \Px{X_r\sim p_r}{\exists x^* \in X_r: \min_{x\neq x^*, x\in X_r} ||x^*-x|| \geq R\sqrt{2}} = 1
\end{align}

Finally, we relax the assumption of $K=1$ for the $K$-nearest-neighbors approximation of $S_r$, since increasing $K$ will strictly increase the overlap between the supports. 
\end{proof}
\clearpage

\section{Proof of~\cref{proposition:outside}}
\label{sec:proof_outside}

\begin{proof}
We start by making a redundant assumption which we will later relax: that the distributions $p_g$ and $p_r$ are uniformly distributed over their respective supports $S_g$ and $S_r$. Note that we have assumed $S_g = B_o\setminus (B \setminus \partial B)$, where $B$ and $B_o$ are $d$ dimensional balls such that $B\subset B_o$ and $\partial B = S_r$. We denote by $R_o$ and $R$ the respective radii of $B_o$ and $B$, and $\hat{S}_r$ denotes the $K$-nearest-neighbors approximation of $S_r$ using a set of i.i.d. observations $X_r=\{x_i\}_{i=1}^N$ from $p_r$, that is, $\hat{S}_r = \cup_{i=1}^N N_K(x_i)$ where $N_K(x_i)$ is the $d$-ball centered at $x_i$ with radius equal to the Euclidean distance of $x_i$ from its $K$-th nearest neighbor in $X_r$. Throughout the proof, we consider the largest $K$ (farthest neighbor) since decreasing $K$ strictly reduces the overlap between the supports.

The direction of the proof is as follows: we consider the event that the farthest neighbor for any sample in $X_r$ is within an Euclidean distance of $R\sqrt{2}$, that is:
\begin{align}
\label{eq:proof_outside_event}
    \{ \forall x^* \in X_r: \max_{x\neq x^*, x\in X_r} \lVert x^*-x\rVert \leq R\sqrt{2} \}
\end{align}
and show that under this event, the support of $p_g$ will have almost no overlap with the approximate support of $p_r$ as $d\rightarrow \infty$. Then, we compute the probability of this event under $p_r$ and observe that it saturates with $d\rightarrow \infty$, concluding the proof.

Under the above event, since the farthest neighbor of any sample $x^* \in X_r$ is within $R\sqrt{2}$, the approximate support is covered by the union of $d$-balls of radius $R\sqrt{2}$ placed at the observed samples in $X_r$, denoted $B_{R\sqrt{2}}(x^*)$, and using the assumption of uniform $p_g$ we have:
\begin{align}
    \Px{p_g}{\hat{S}_r \cap S_g} \leq N \Px{p_g}{B_{R\sqrt{2}}(x^*) \cap S_g} \leq N \Px{p_g}{B_{R\sqrt{2}}(x^*) \cap B_o} = N \frac{V(B_{R\sqrt{2}}(x^*) \cap B_o)}{V(B_o) - V(B)}
\end{align}

Now we consider two cases. First, when $R_o > R\sqrt{2}$, we can write:
\begin{align}
    \Px{p_g}{\hat{S}_r \cap S_g} \leq N \frac{V(B_{R\sqrt{2}})}{V(B_o) - V(B)} = \frac{N}{(\frac{R_o}{R\sqrt{2}})^d - (\frac{1}{\sqrt{2}})^d}
\end{align}
and then under the assumption of $\frac{R_o}{R\sqrt{2}} > 1$, if $\lim_{d\rightarrow\infty} N\epsilon^{-d} = 0$ $\forall \epsilon > 1$, \ie the number of samples less than exponential in $d$, the above tends to 0 as $d\rightarrow\infty$.

Second, we consider the case where $R_o \leq R\sqrt{2}$ and compute an upper bound for the intersection volume $V(B_{R\sqrt{2}}(x^*) \cap B_o)$. To do so, we consider the triangle formed by the centers of the three balls $B_o, B, B_{R\sqrt{2}}(x^*)$, and denote the length of the respective sides as $\delta_{ob}, \delta_{b*}, \delta_{o*}$ corresponding to the distances between the centers of the balls. The intersection volume $V(B_{R\sqrt{2}}(x^*) \cap B_o)$ is maximized when the length $\delta_{o*}$ is minimized. From the triangle inequality we have $\delta_{o*} \geq |\delta_{b*}-\delta_{ob}|$, and from $x^* \in S_r = \partial B$ we have $\delta_{b*} = R$, so it follows that $\delta_{o*} \geq |R-\delta_{ob}|$. Furthermore, from $B\subset B_o$ we have $\delta_{ob} \leq R_o - R$ and thus $R-\delta_{ob} \geq 2R-R_o$, and then from the second case assumption of $R_o \leq R\sqrt{2} < 2R$ it follows that $\delta_{o*} \geq 2R-R_o$. Therefore, we can compute the maximum of $V(B_{R\sqrt{2}}(x^*) \cap B_o)$ as the volume of intersection between two $d$-balls of radii $R\sqrt{2}$ and $R_o$ with centers at a distance of $2R-R_o = \min \delta_{o*}$.

This intersection is consisted of two $d$ dimensional hyperspherical caps, one on $B_{R\sqrt{2}}(x^*)$ at colatitude angle $\phi_*$, and another on $B_o$ at colatitude angle $\phi_o$. We first show that both angles are acute, by showing that the intersection triangle formed by the centers of the two intersecting balls and a point of intersection, whose sides are of length $R_o, R\sqrt{2}, 2R-R_o$, is an acute triangle. The largest side of this triangle has length $R\sqrt{2} \geq R_o > 2R-R_o$, and $2R^2 < R_o^2 + (2R-R_o)^2$, so its acuteness follows from the Pythagorean theorem. The two acute colatitude angles can be computed using Pythagorean equality on the triangle's altitude:
\begin{align}
    \sin^2 \phi_o &= 1 - \frac{(R_o-R)^4}{(2R-R_o)^2R_o^2}\\
    \sin^2 \phi_* &= 1 - \frac{(3R-2R_o)^2}{2(2R-R_o)^2}
\end{align}

Now we can use the formula in~\cref{eq:cap_volume} for the volume of hyperspherical cap at collatitude angle $\phi < \frac{\pi}{2}$ to compute the volume of intersection:
\begin{align}
    \Px{p_g}{\hat{S_r} \cap S_g} \leq N\frac{V(B_{R\sqrt{2}}(x^*) \cap B_o)}{V(B_o) - V(B)} \leq \frac{NV(B_{\phi_o < \frac{\pi}{2}}^{cap})}{V(B_o) - V(B)} + \frac{NV(B_{\phi_* < \frac{\pi}{2}}^{cap})}{V(B_o) - V(B)}
\end{align}

For the first term, using the approximation in~\cref{eq:beta_approx} (omitting higher order terms in $\sin \phi_o < 1$), we have:
\begin{align}
\label{eq:proof_outside_capo_volume}
    \frac{NV(B_{\phi_o}^{cap})}{V(B_o) - V(B)} \approx
    \frac{CN}{2\sqrt{d+1}} \frac{(\frac{R_o}{R})^d (\sin^2 \phi_o)^{\frac{d+1}{2}}}{(\frac{R_o}{R})^d - 1}
\end{align}

For the second term, again using the approximation in~\cref{eq:beta_approx} (omitting higher order terms in $\sin \phi_* < 1$), we have:
\begin{align}
\label{eq:proof_outside_caps_volume}
    \frac{NV(B_{\phi_*}^{cap})}{V(B_o) - V(B)} \approx
    \frac{CN}{2\sqrt{d+1}} \frac{(\sqrt{2})^d (\sin^2 \phi_*)^{\frac{d+1}{2}}}{(\frac{R_o}{R})^d - 1}
    = \frac{CN}{2\sqrt{2}\sqrt{d+1}} \frac{(2\sin^2 \phi_*)^{\frac{d+1}{2}}}{(\frac{R_o}{R})^d - 1}
\end{align}
where for the numerator we can show that $\sqrt{2\sin^2 \phi_*} < \frac{R_o}{R}$ as follows:
\begin{align}
    2\sin^2 \phi_* - \frac{R_o^2}{R^2} = 2 - \frac{(3R-2R_o)^2}{(2R-R_o)^2} - \frac{R_o^2}{R^2} = \frac{-(R_o-R)^4}{(2R-R_o)^2R^2} < 0
\end{align}

When $\lim_{d\rightarrow\infty} N\epsilon^{-d} = 0$ $\forall \epsilon > 1$, \ie the number of samples less than exponential in $d$, both~\cref{eq:proof_outside_capo_volume,eq:proof_outside_caps_volume} tend to 0 as $d\rightarrow\infty$, because $\sin \phi_o < 1$ and $\sqrt{2\sin^2 \phi_*} < \frac{R_o}{R}$, respectively.

At this point, we can relax the requirement that $p_g$ is uniform, by noting that any absolutely continuous probability measure on $S_g$ must be within a constant factor of the measure produced by the uniformly distributed measure (due to Radon-Nikodym theorem). This concludes the first part of the proof that $\Px{p_g}{\hat{S_r} \cap S_g} = 0$ as $d \rightarrow \infty$ under the event in~\cref{eq:proof_outside_event}.

What remains is to compute the probability of the event in~\cref{eq:proof_outside_event} under $p_r$. This can be computed as the probability that given any sample $x^*\in X_r$, all other samples in $X_r$ fall within a distance of $R\sqrt{2}$ from $x^*$:
\begin{align}
    \Px{X_r \sim p_r}{\forall x^* \in X_r: \max_{x\neq x^*, x\in X_r} \lVert x^*-x\rVert \leq R\sqrt{2}} = (1-\Px{p_r}{\lVert x^*-x\rVert > R\sqrt{2}})^{N-1}
\end{align}
where for the inner event, given the uniform assumption on $p_r$, we have:
\begin{align}
    \Px{p_r}{\lVert x^*-x \rVert > R\sqrt{2} }
    = \frac{A(B^{cap}_{\phi<\frac{\pi}{2}})}{A(S_r)}
\end{align}
and the rest of the proof follows the exact steps that connect~\cref{eq:proof_inside_event_pr_capratio,eq:proof_inside_event_pr_1}, concluding that for any absolutely continuous $p_r$ and the number of samples less than exponential in $d$, we have $\Px{X_r \sim p_r}{\forall x^* \in X_r: \max_{x\neq x^*, x\in X_r} \lVert x^*-x\rVert \leq R\sqrt{2}} = 1$ as $d\rightarrow \infty$.

\end{proof}

\clearpage
\section{Experiments with Varying Radii}
\label{sec:varying_r}
\begin{figure*}[h!]
    \centering
    \subfloat[Precision (R=0.5)]{
        \centering
        \includegraphics[trim=25 0 0 25, clip, width=0.253\textwidth]{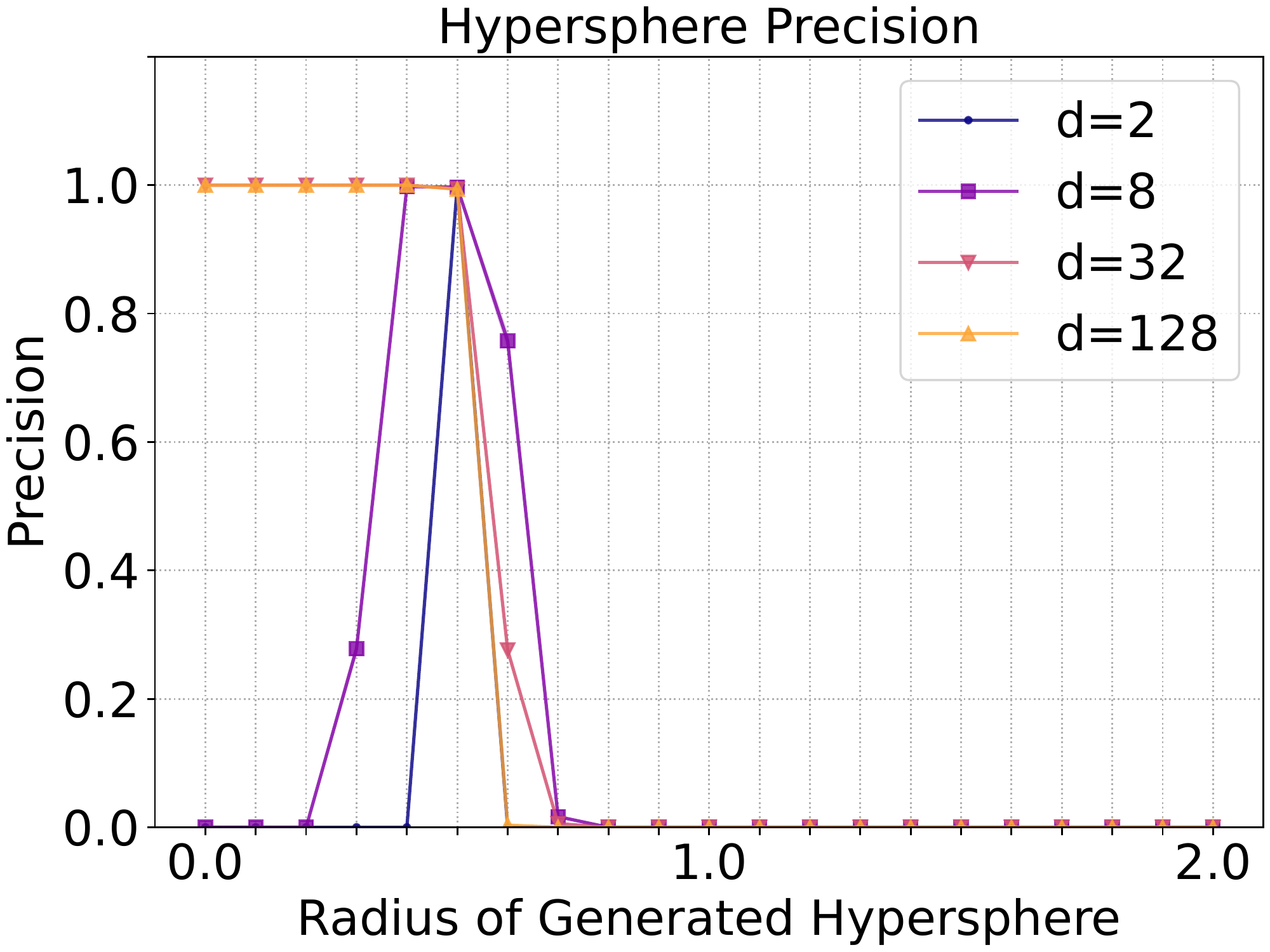}
        \label{fig:sphere_r05_pr_p}}
    \subfloat[Recall (R=0.5)]{
        \centering
        \includegraphics[trim=64 0 0 25, clip, width=0.235\textwidth]{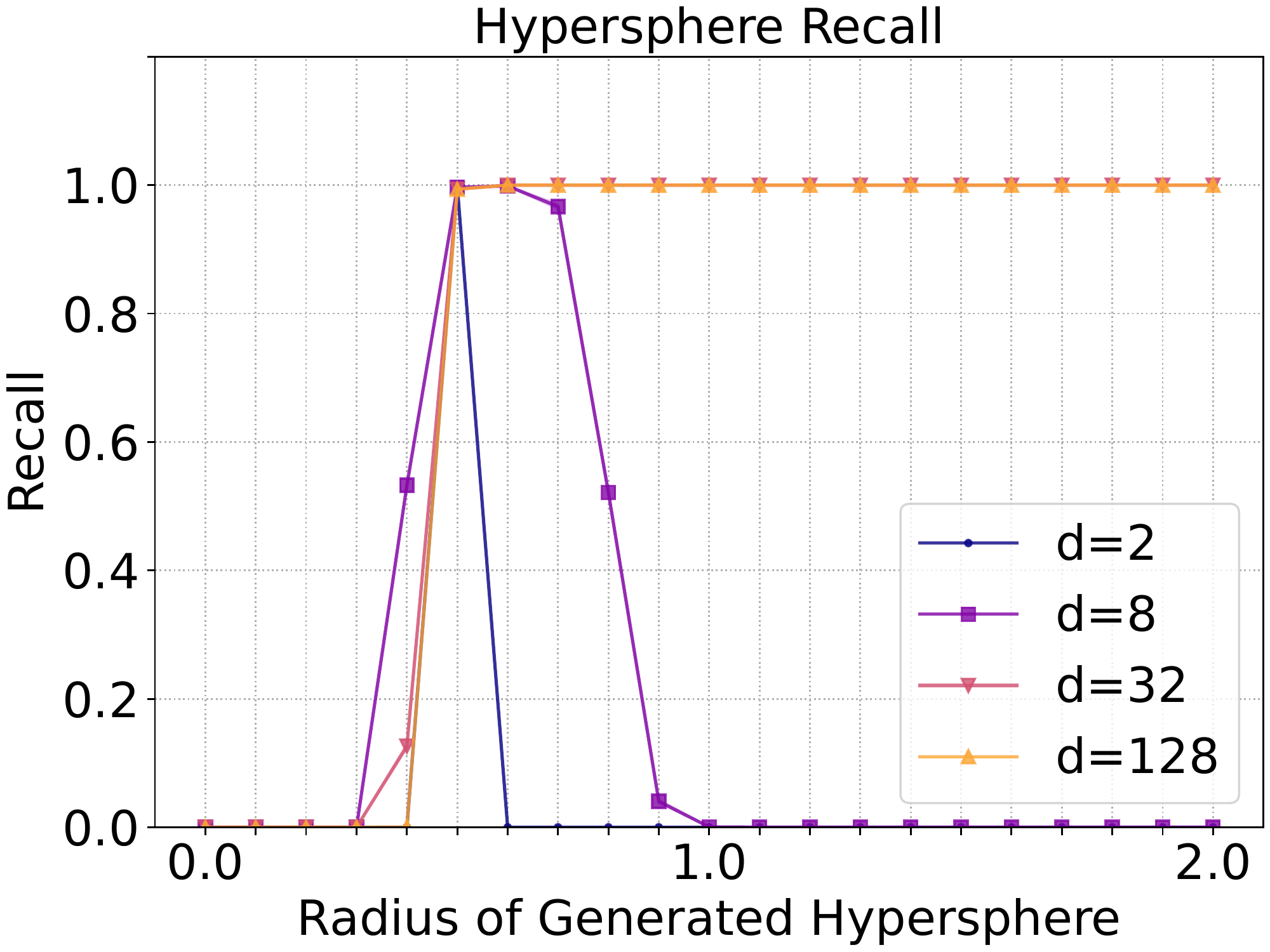}
        \label{fig:sphere_r05_pr_r}}
    \subfloat[symPrecision (R=0.5)]{
        \centering
        \includegraphics[trim=64 0 0 25, clip, width=0.235\textwidth]{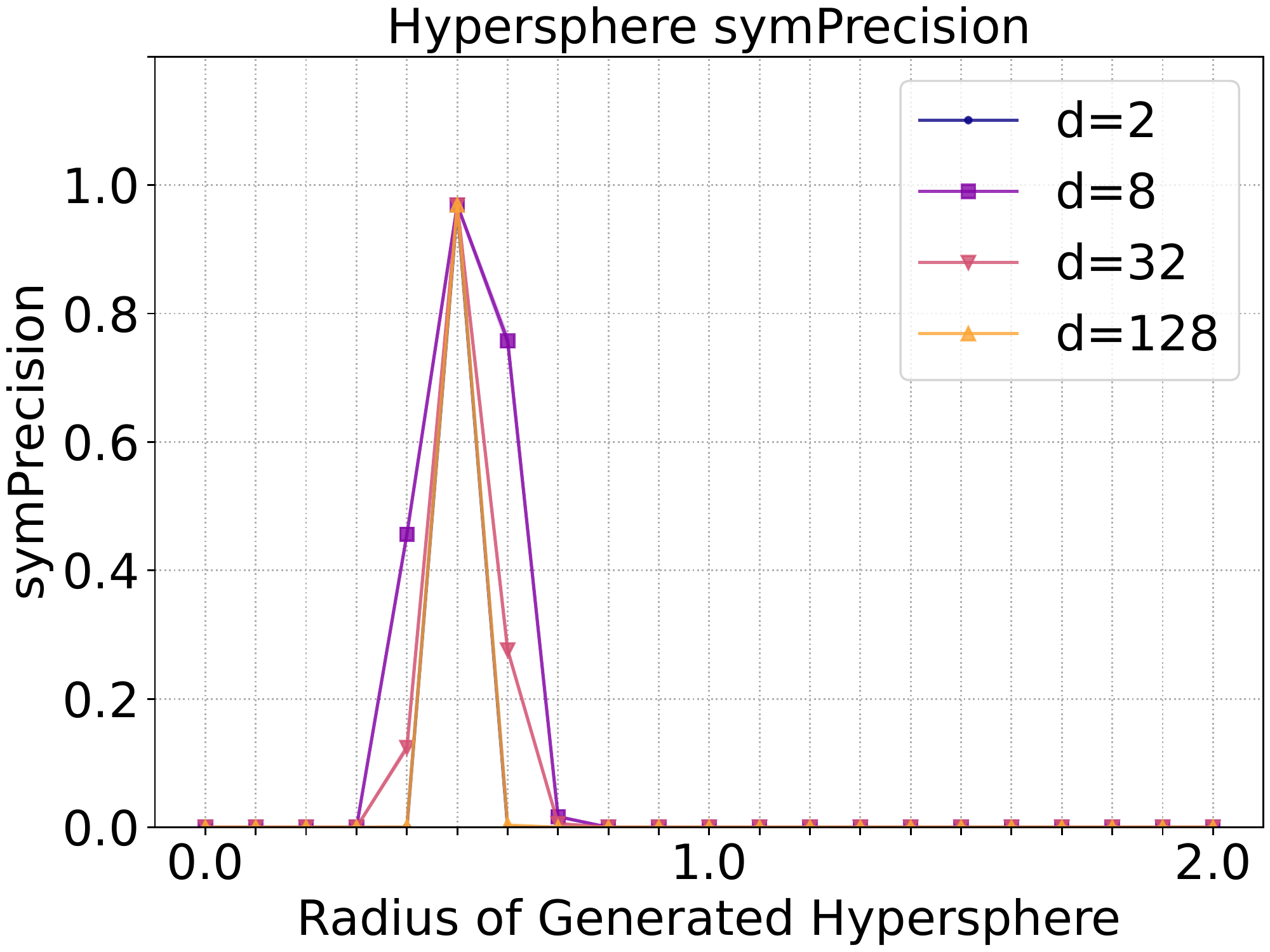}
        \label{fig:sphere_r05_pr_sp}}
    \subfloat[symRecall (R=0.5)]{
        \centering
        \includegraphics[trim=64 0 0 25, clip, width=0.235\textwidth]{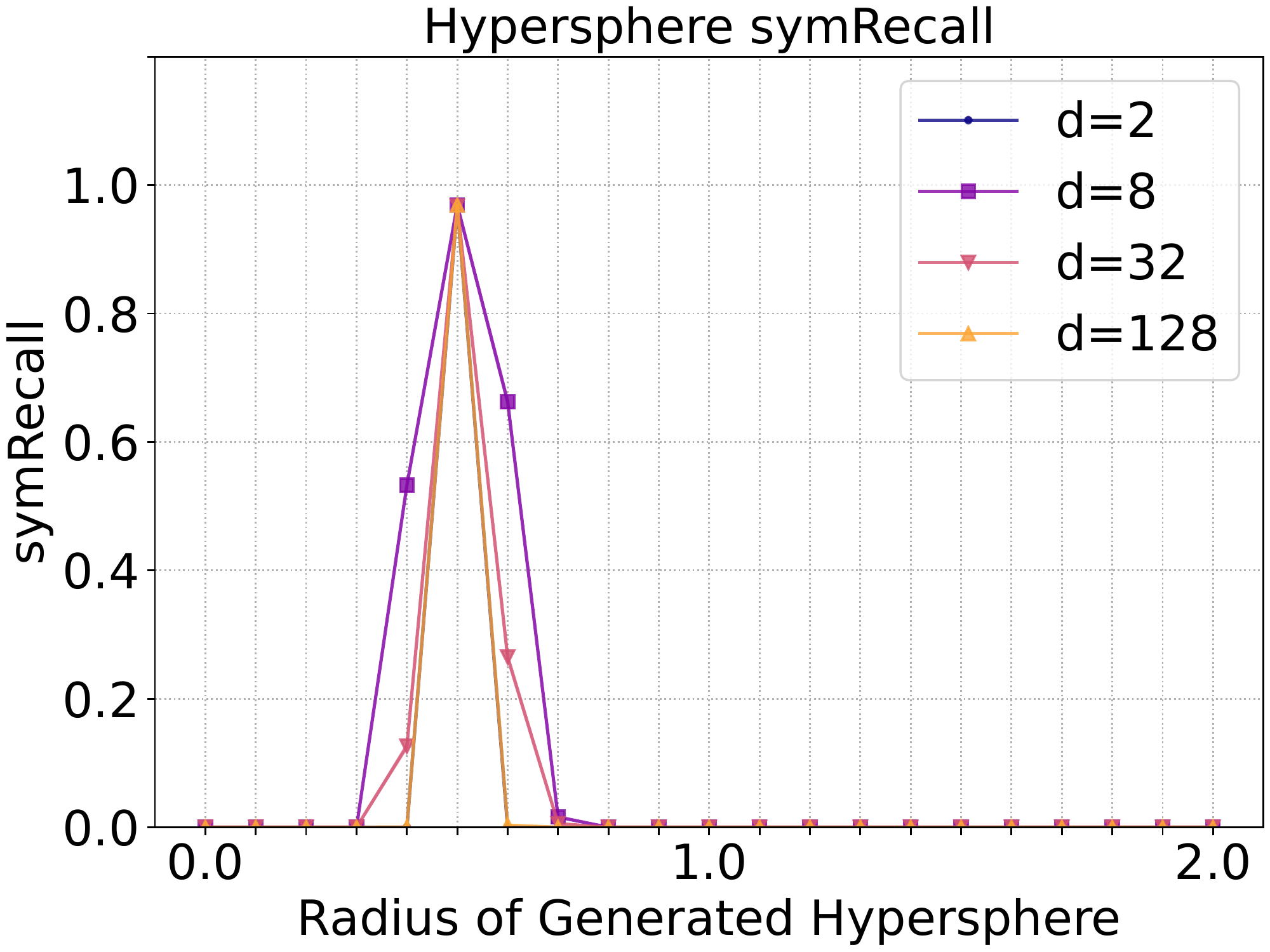}
        \label{fig:sphere_r05_pr_sr}}\\
    \subfloat[Precision (R=2)]{
        \centering
        \includegraphics[trim=25 0 0 25, clip, width=0.253\textwidth]{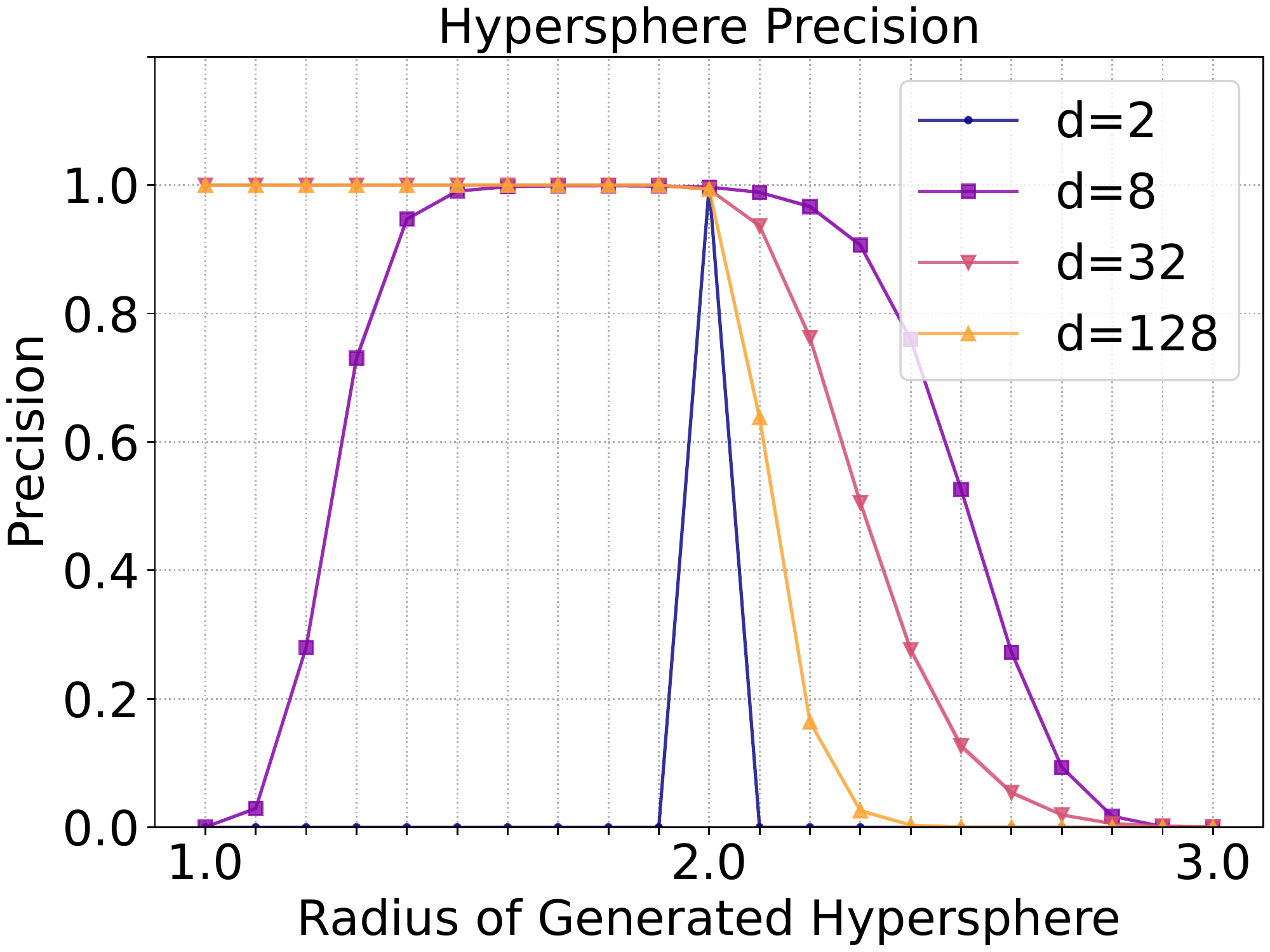}
        \label{fig:sphere_r2_pr_p}}
    \subfloat[Recall (R=2)]{
        \centering
        \includegraphics[trim=64 0 0 25, clip, width=0.235\textwidth]{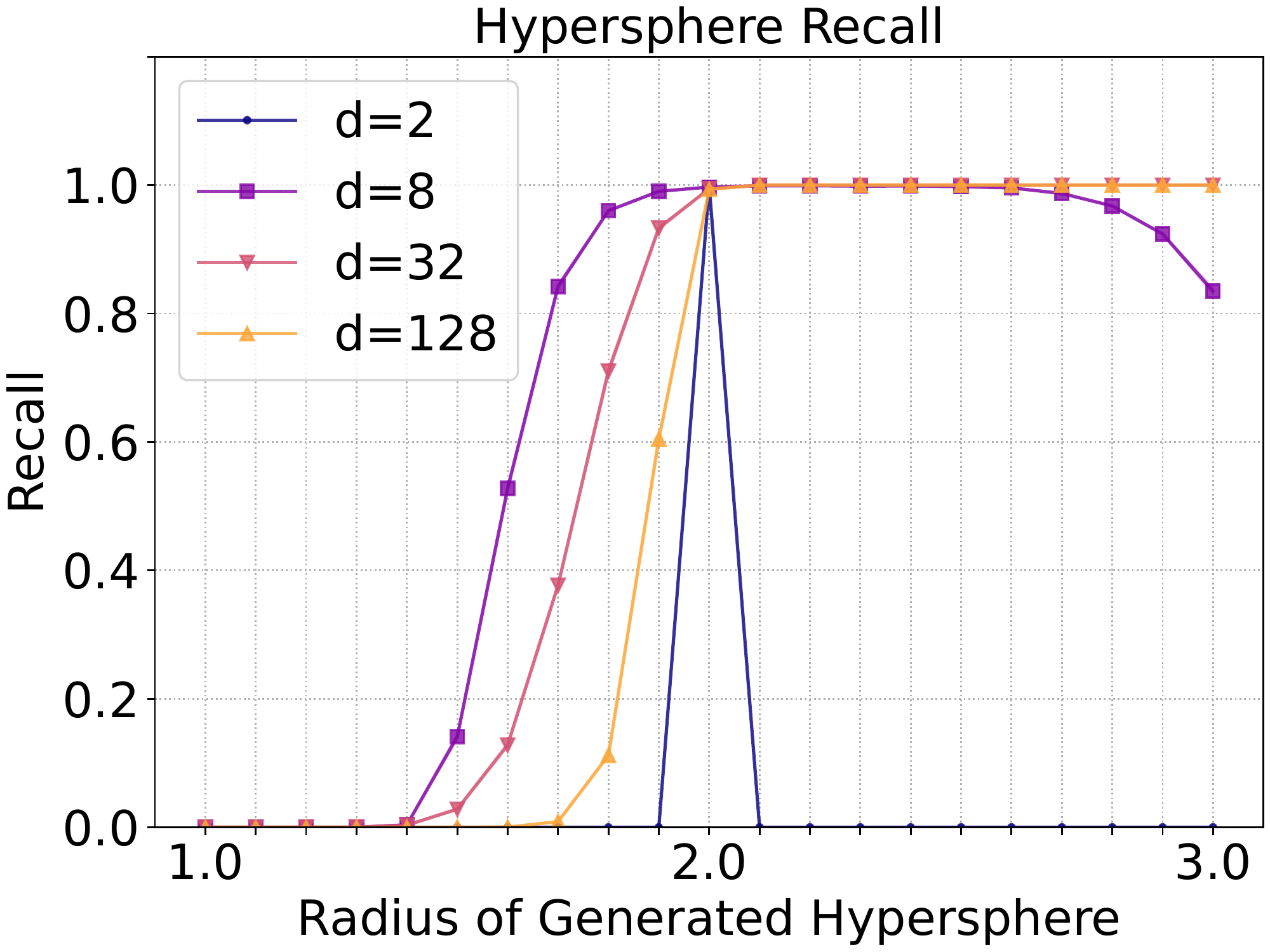}
        \label{fig:sphere_r2_pr_r}}
    \subfloat[symPrecision (R=2)]{
        \centering
        \includegraphics[trim=64 0 0 25, clip, width=0.235\textwidth]{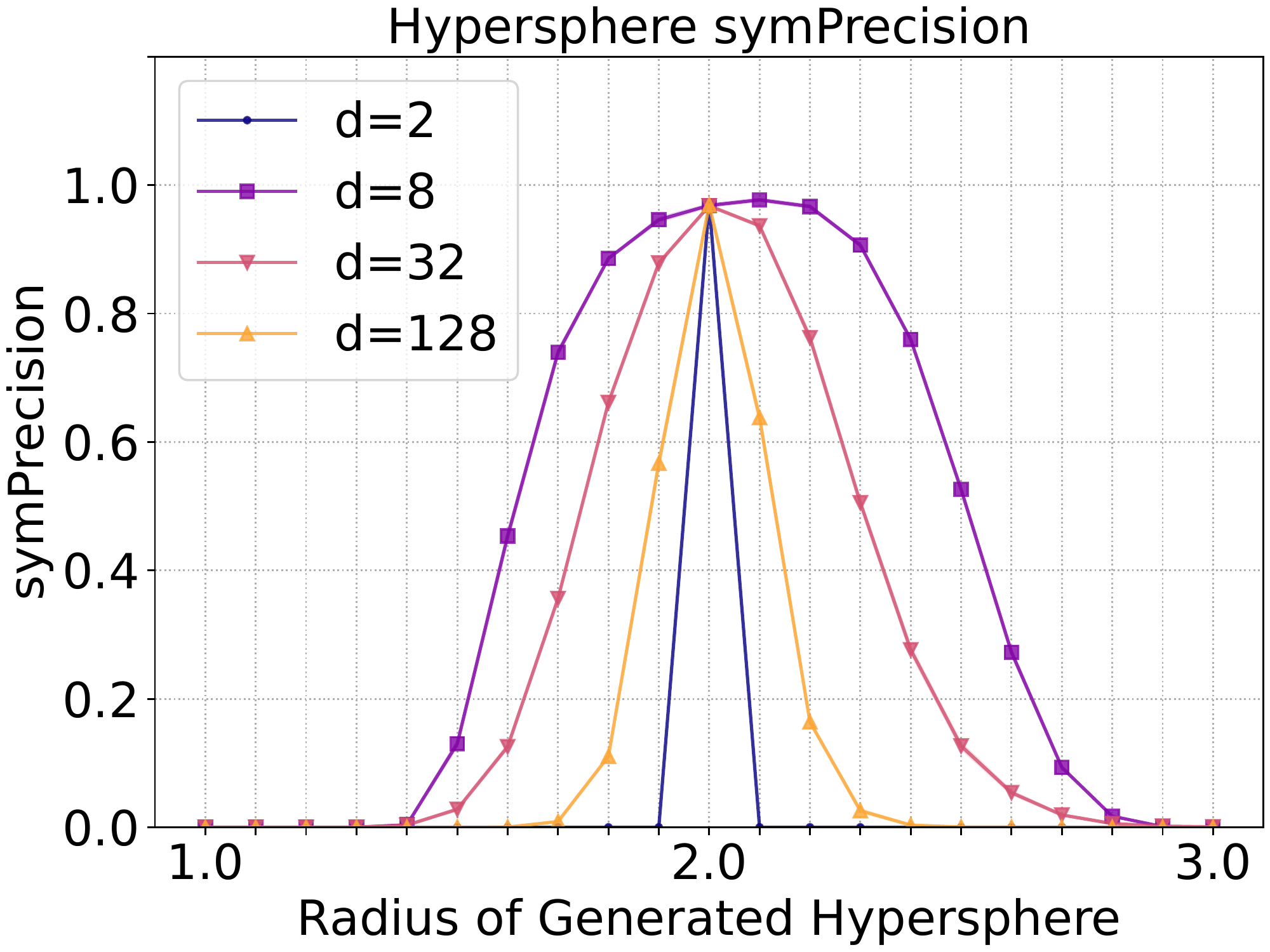}
        \label{fig:sphere_r2_pr_sp}}
    \subfloat[symRecall (R=2)]{
        \centering
        \includegraphics[trim=64 0 0 25, clip, width=0.235\textwidth]{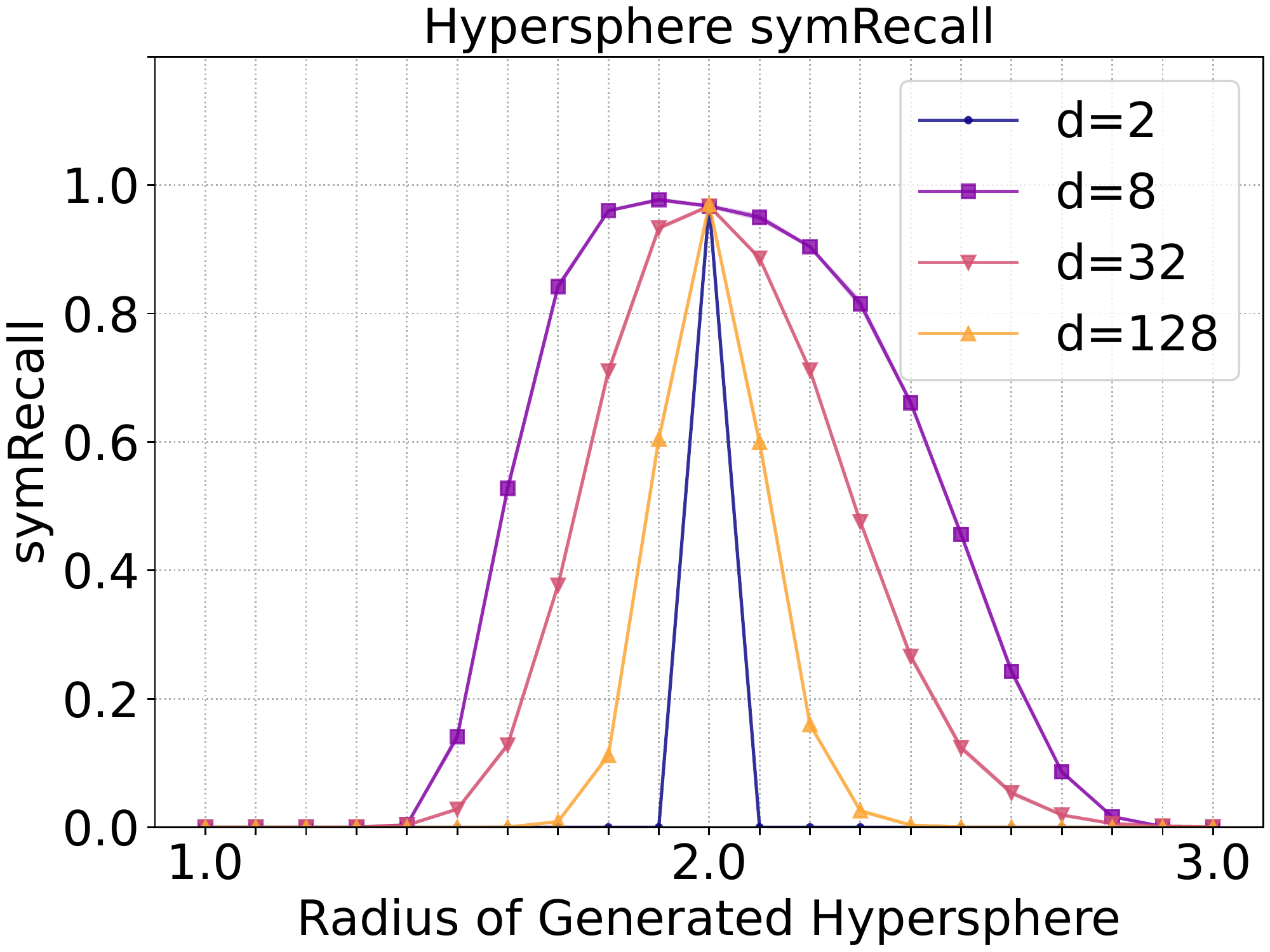}
        \label{fig:sphere_r2_pr_sr}}\\
    \subfloat[Precision (R=8)]{
        \centering
        \includegraphics[trim=25 0 0 25, clip, width=0.253\textwidth]{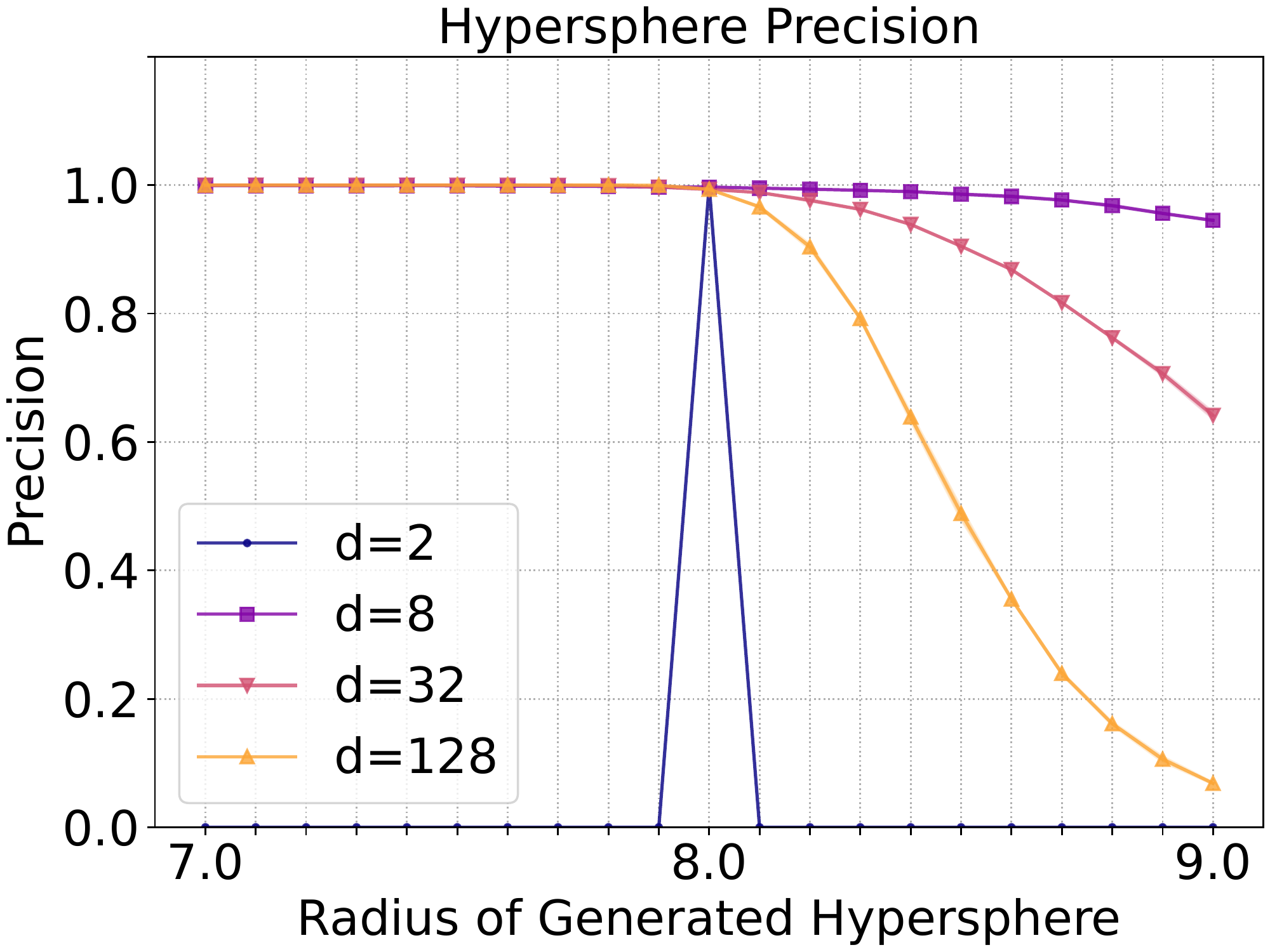}
        \label{fig:sphere_r8_pr_p}}
    \subfloat[Recall (R=8)]{
        \centering
        \includegraphics[trim=64 0 0 25, clip, width=0.235\textwidth]{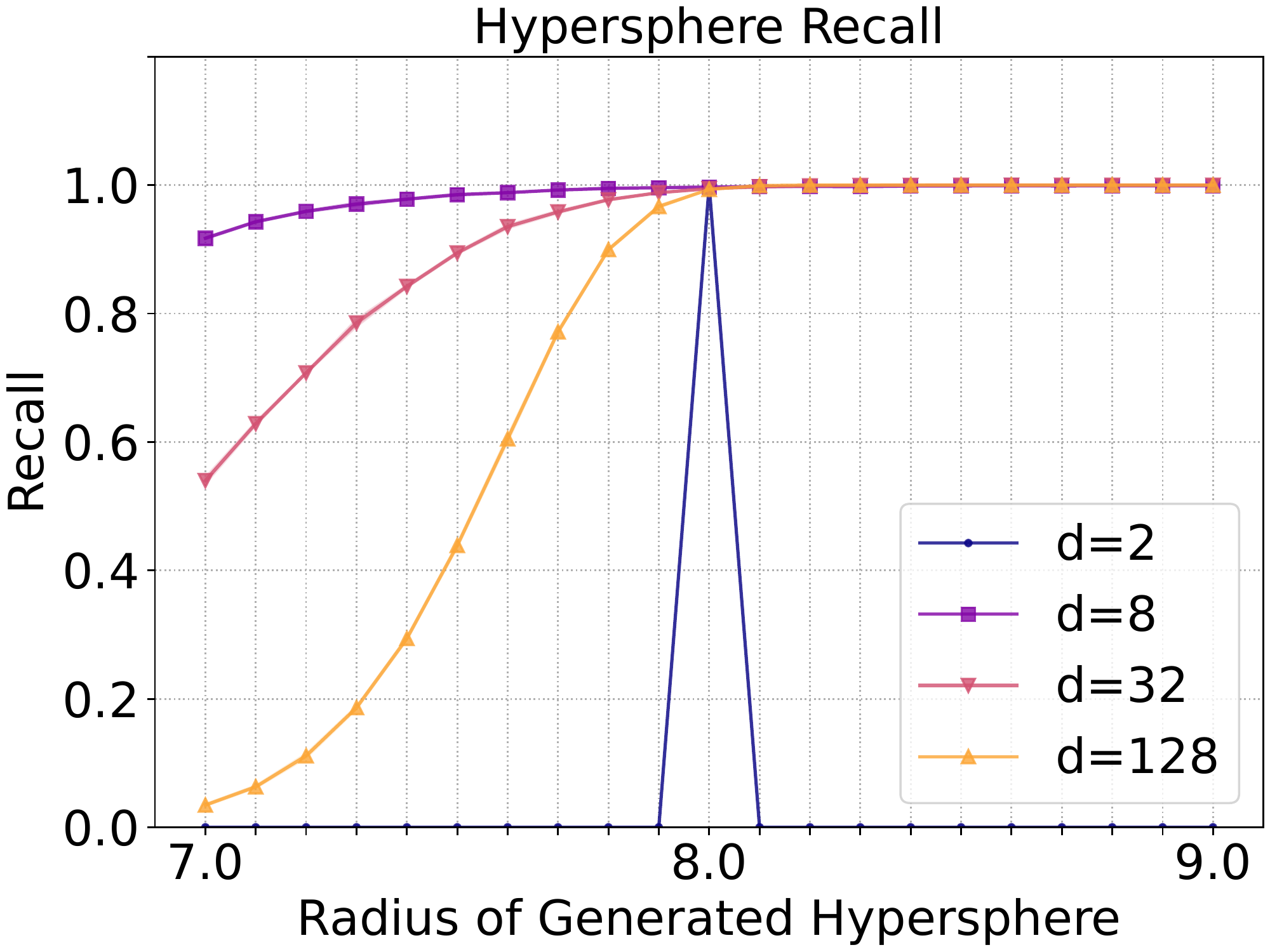}
        \label{fig:sphere_r8_pr_r}}
    \subfloat[symPrecision (R=8)]{
        \centering
        \includegraphics[trim=64 0 0 25, clip, width=0.235\textwidth]{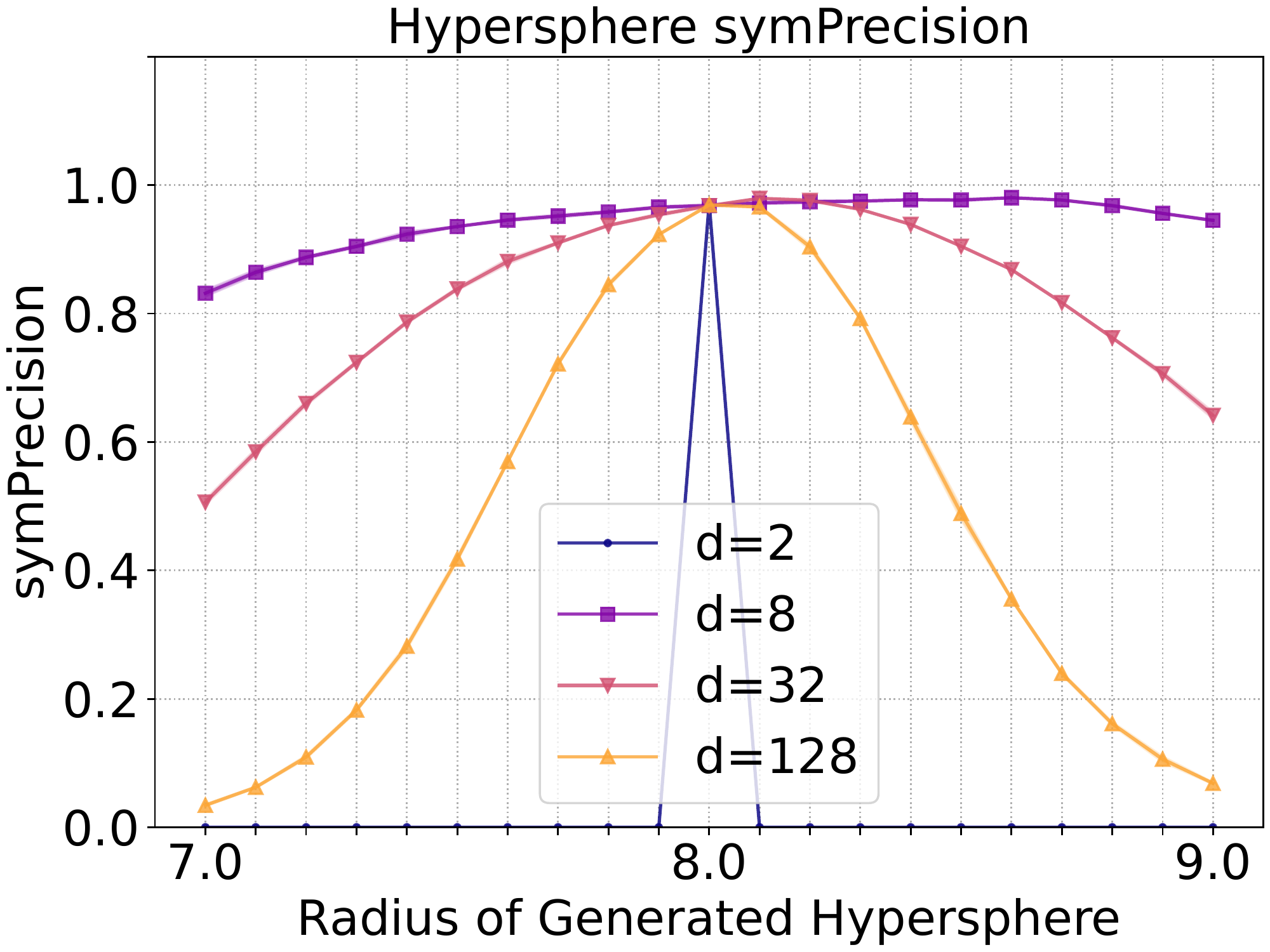}
        \label{fig:sphere_r8_pr_sp}}
    \subfloat[symRecall (R=8)]{
        \centering
        \includegraphics[trim=64 0 0 25, clip, width=0.235\textwidth]{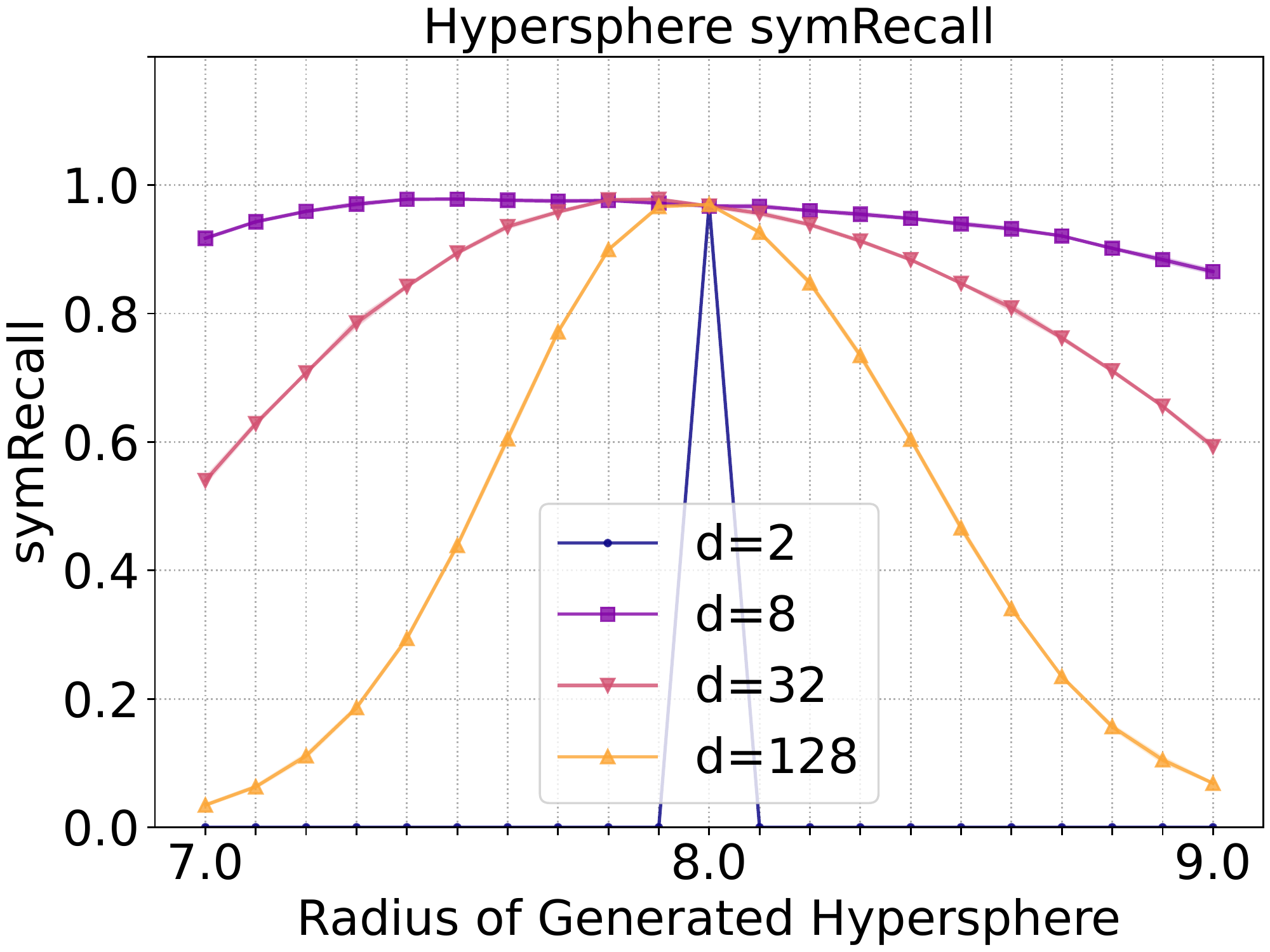}
        \label{fig:sphere_r8_pr_sr}}
    \caption{Hyperspherical reference and generated supports of varying radii at $K=5$ of $K$-nearest-neighbors.}
    \label{fig:sphere_rm}
\end{figure*}
\begin{figure*}[h!]
    \centering
    \subfloat[Precision (R=0.5)]{
        \centering
        \includegraphics[trim=25 0 0 25, clip, width=0.253\textwidth]{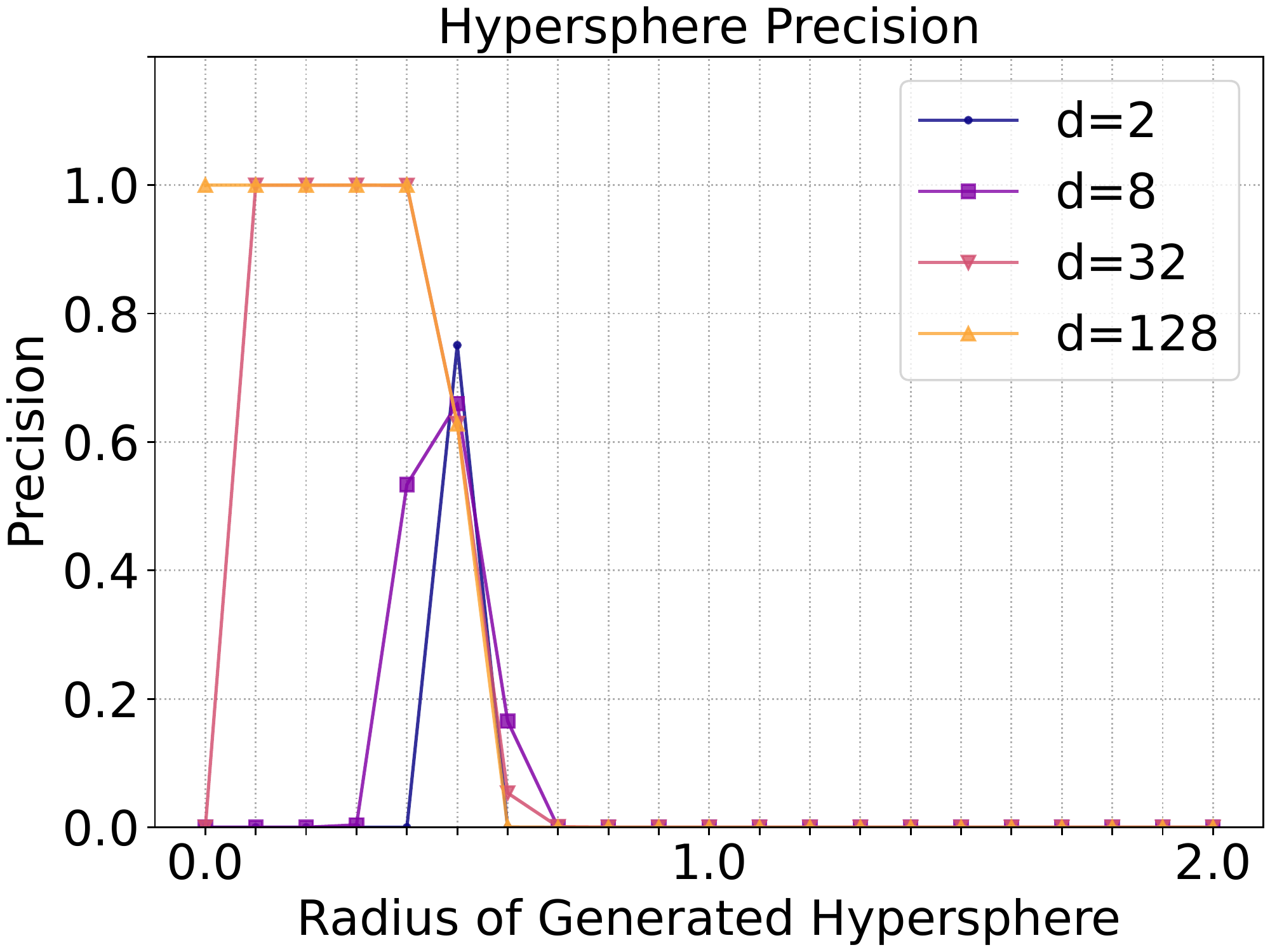}
        \label{fig:sphere_r05_pr_p_k1}}
    \subfloat[Recall (R=0.5)]{
        \centering
        \includegraphics[trim=64 0 0 25, clip, width=0.235\textwidth]{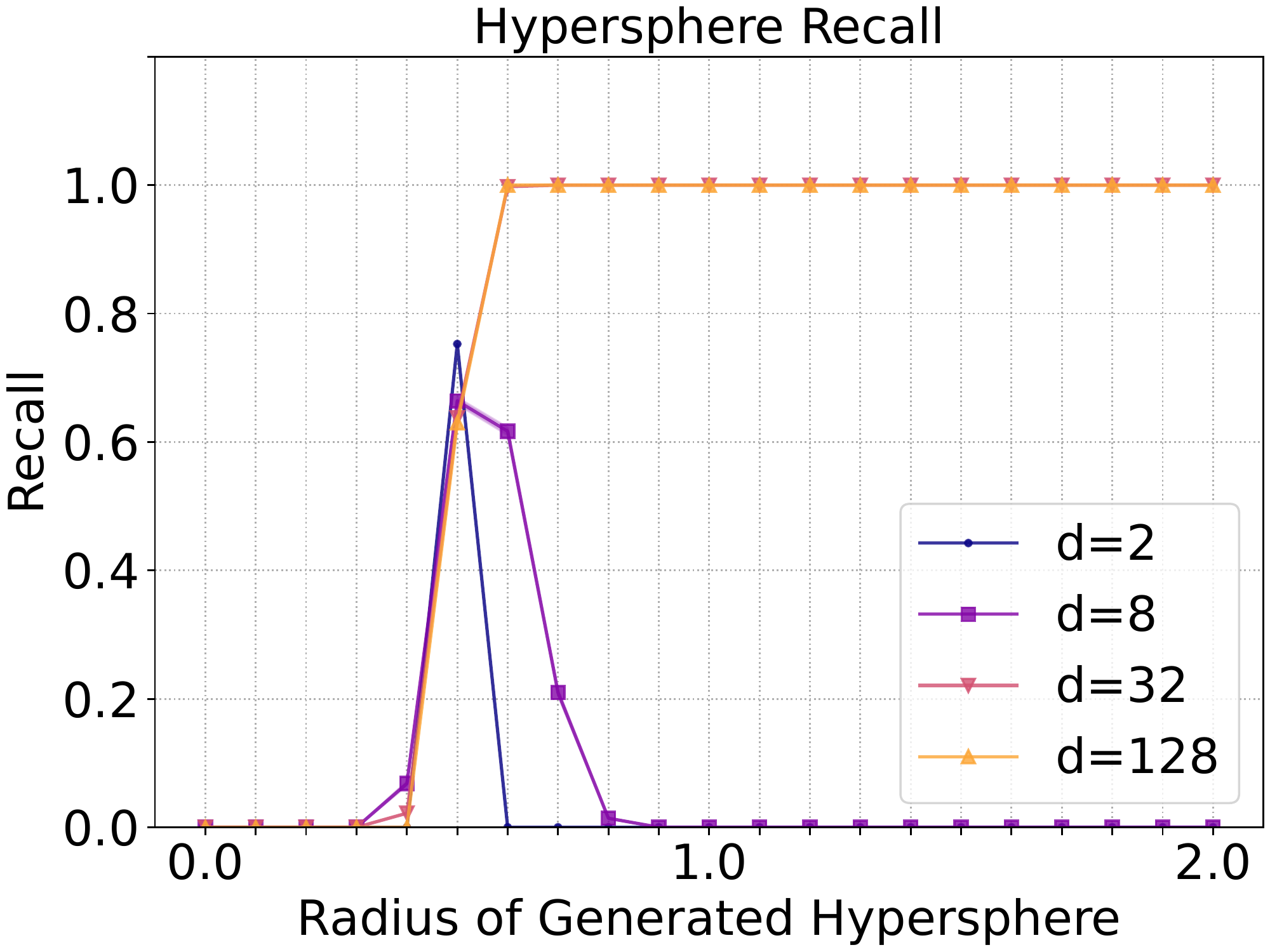}
        \label{fig:sphere_r05_pr_r_k1}}
    \subfloat[symPrecision (R=0.5)]{
        \centering
        \includegraphics[trim=64 0 0 25, clip, width=0.235\textwidth]{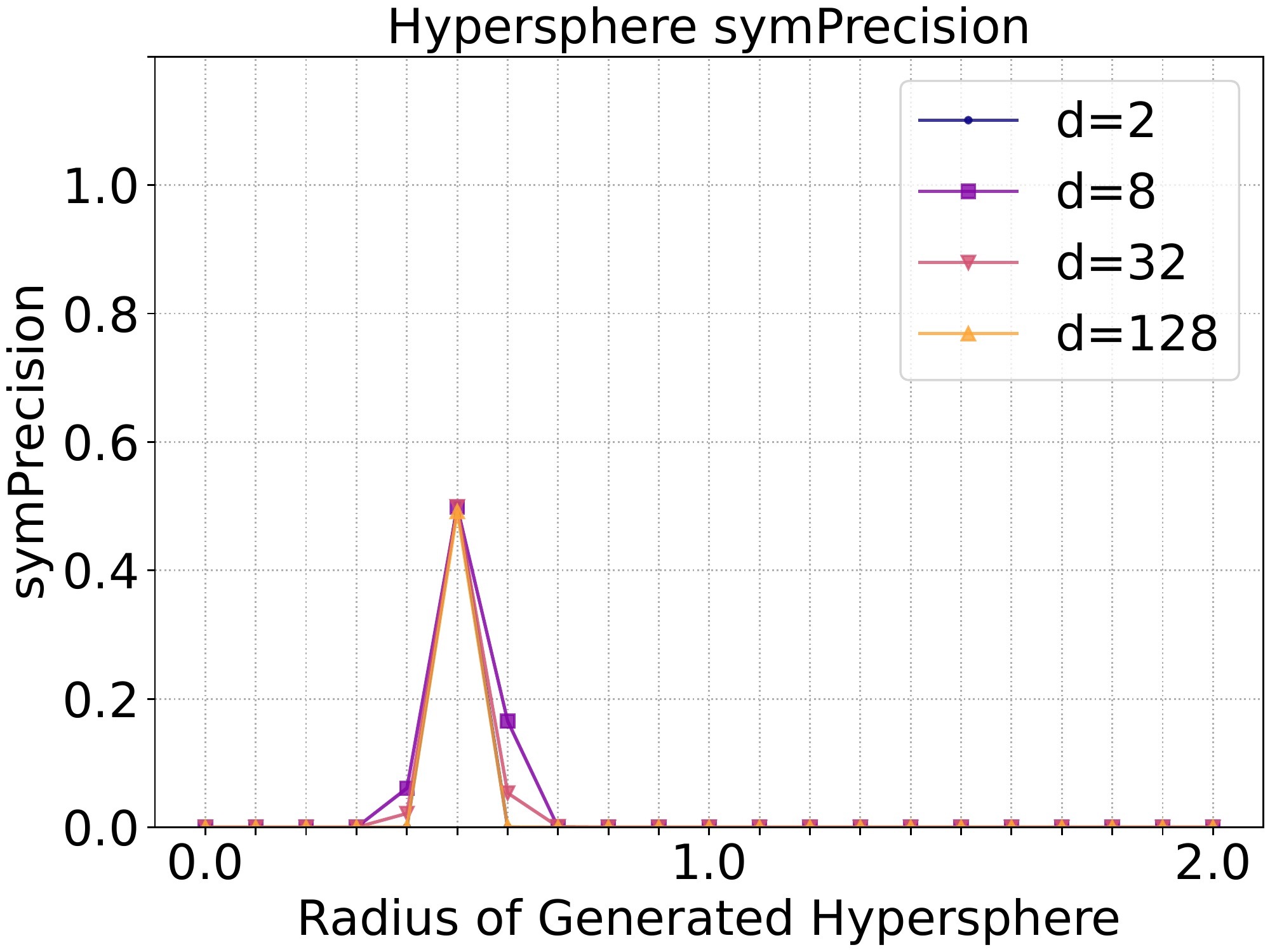}
        \label{fig:sphere_r05_pr_sp_k1}}
    \subfloat[symRecall (R=0.5)]{
        \centering
        \includegraphics[trim=64 0 0 25, clip, width=0.235\textwidth]{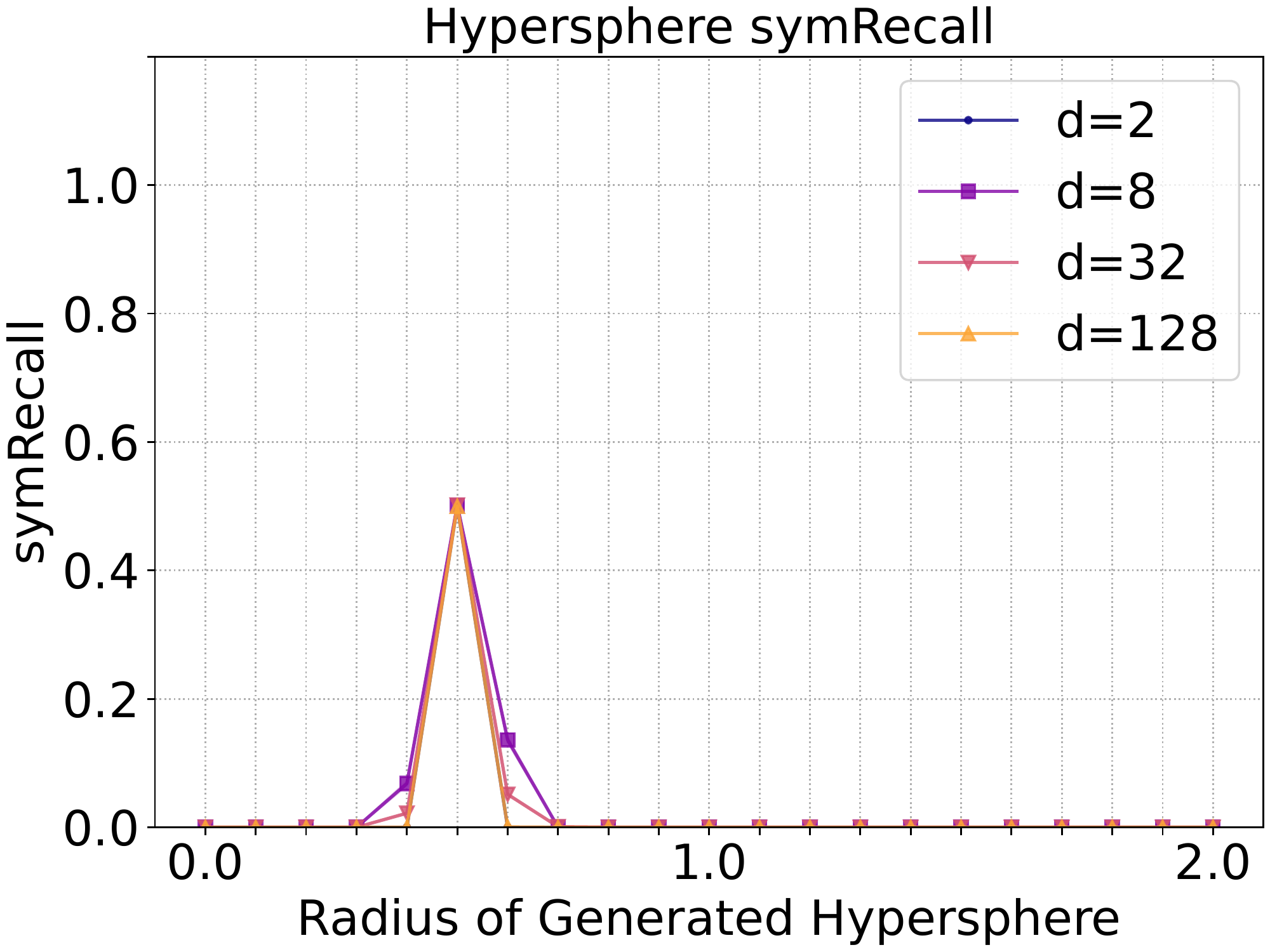}
        \label{fig:sphere_r05_pr_sr_k1}}\\
    \subfloat[Precision (R=2)]{
        \centering
        \includegraphics[trim=25 0 0 25, clip, width=0.253\textwidth]{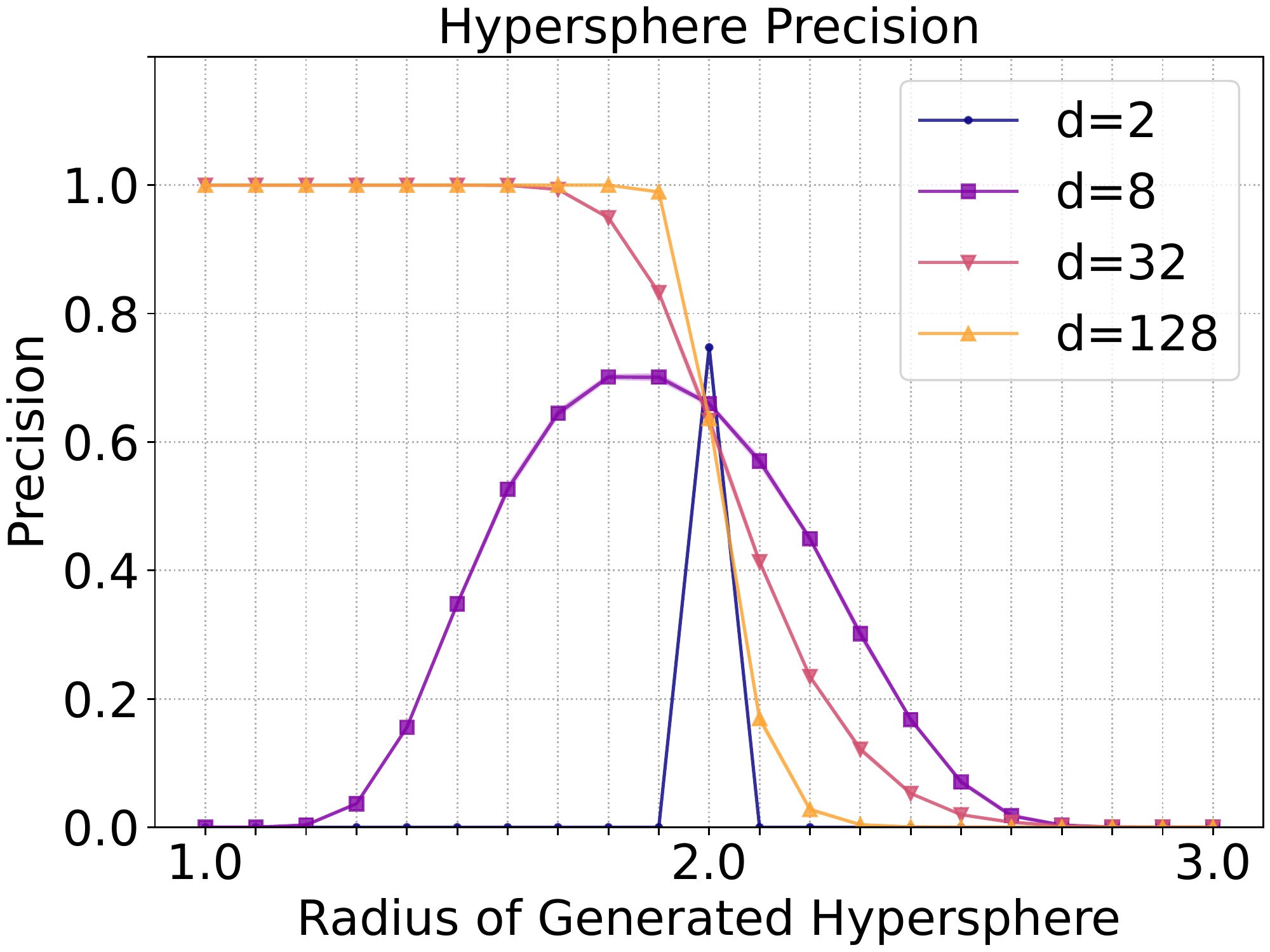}
        \label{fig:sphere_r2_pr_p_k1}}
    \subfloat[Recall (R=2)]{
        \centering
        \includegraphics[trim=64 0 0 25, clip, width=0.235\textwidth]{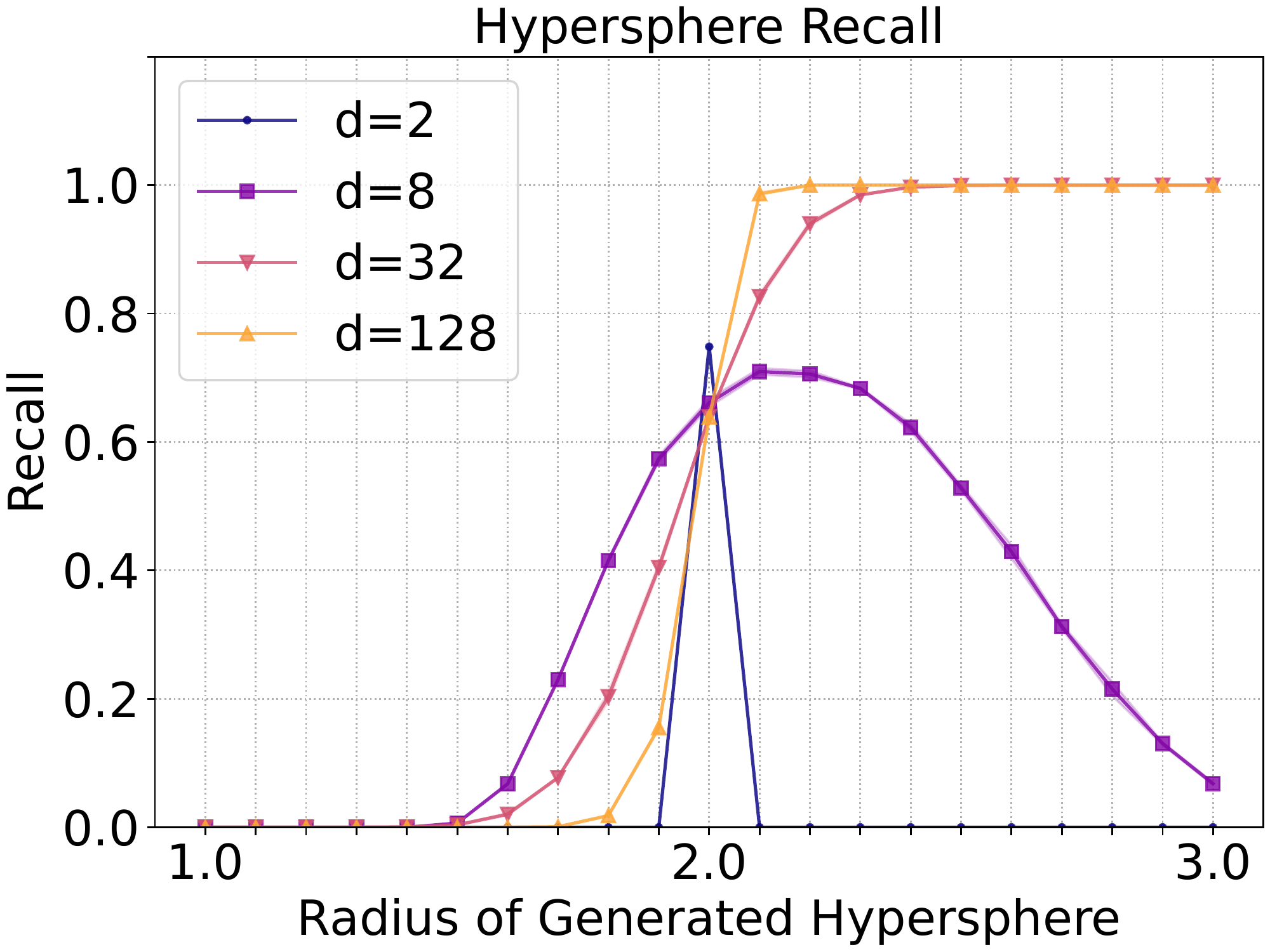}
        \label{fig:sphere_r2_pr_r_k1}}
    \subfloat[symPrecision (R=2)]{
        \centering
        \includegraphics[trim=64 0 0 25, clip, width=0.235\textwidth]{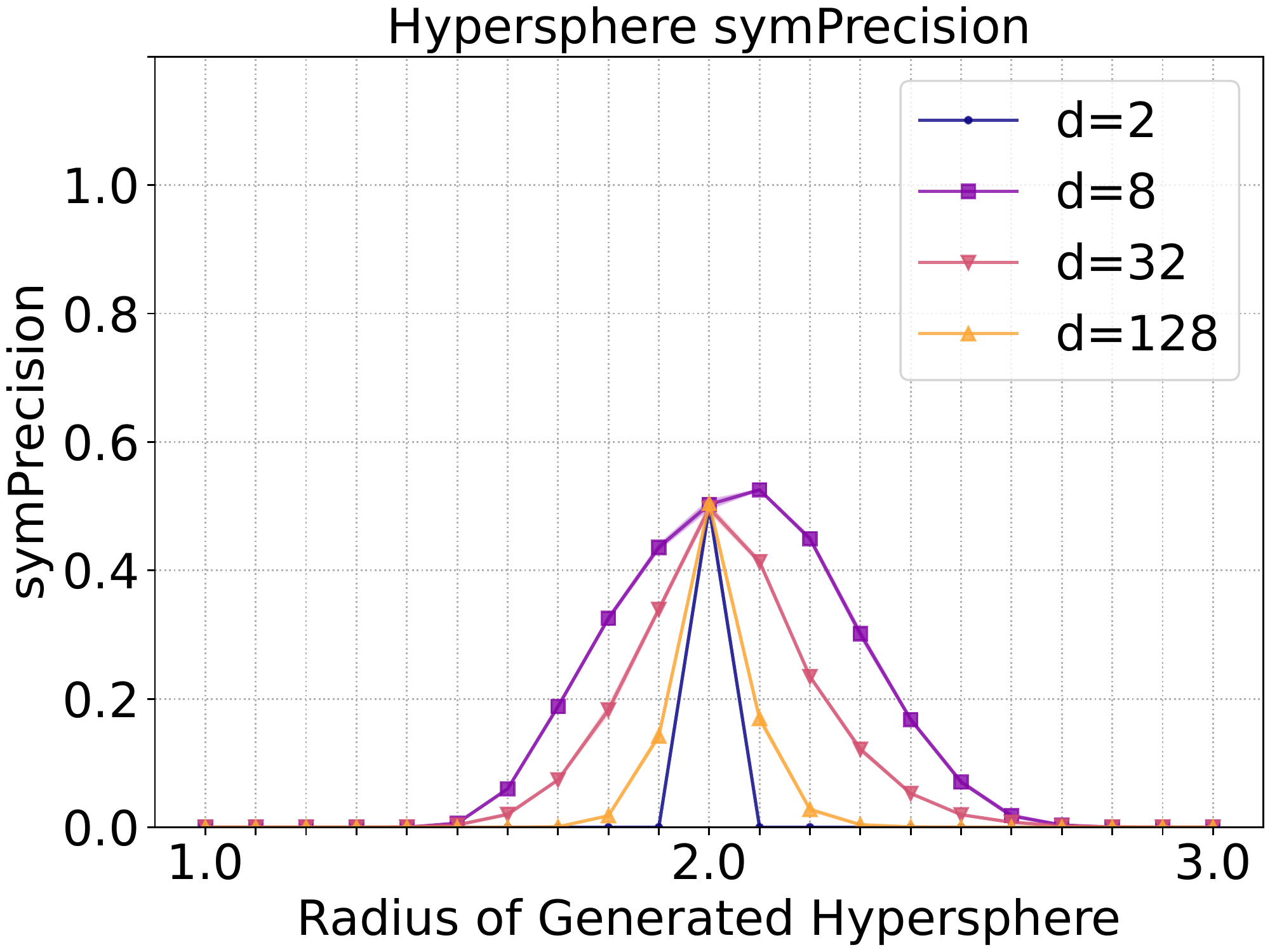}
        \label{fig:sphere_r2_pr_sp_k1}}
    \subfloat[symRecall (R=2)]{
        \centering
        \includegraphics[trim=64 0 0 25, clip, width=0.235\textwidth]{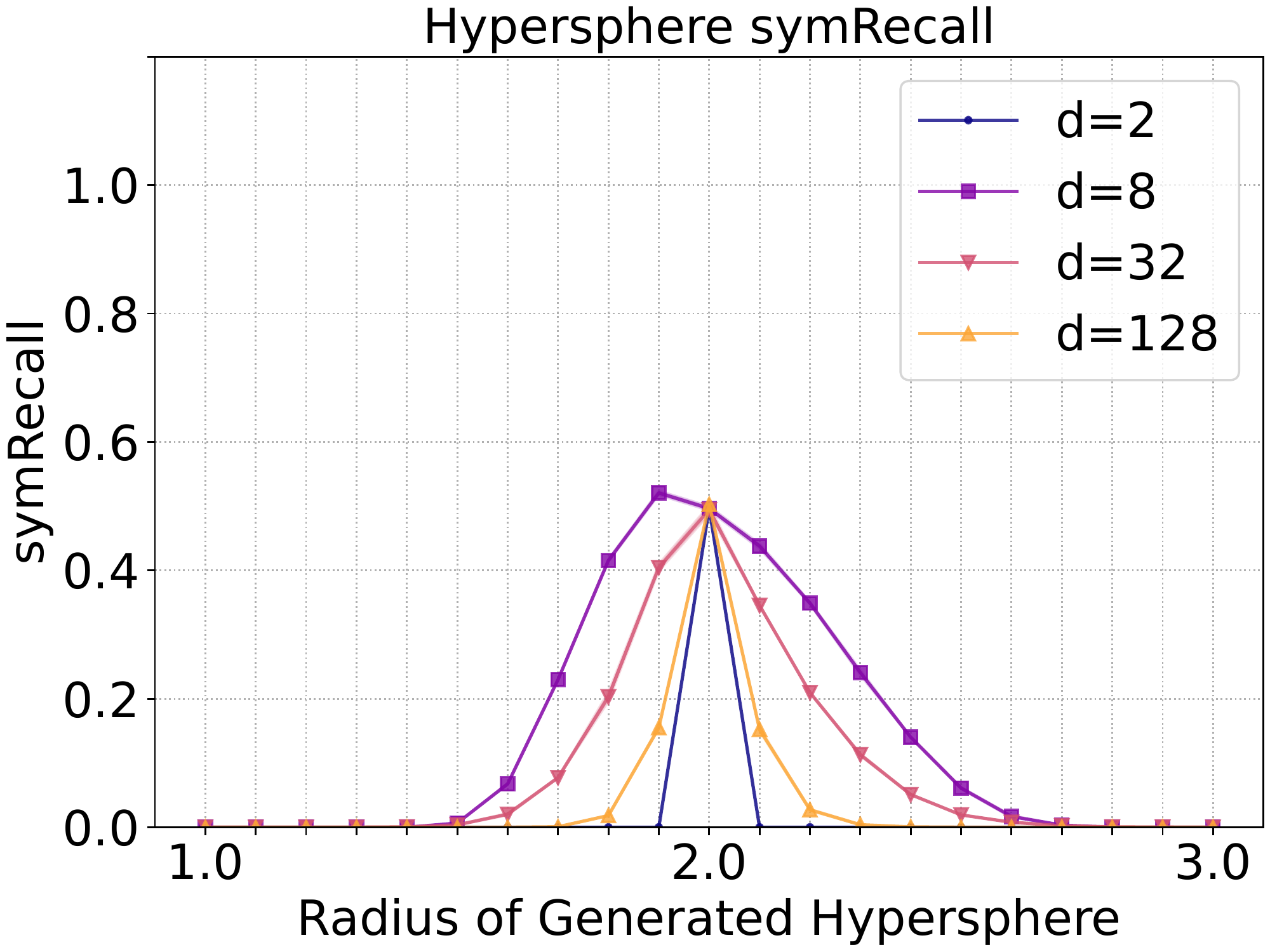}
        \label{fig:sphere_r2_pr_sr_k1}}\\
    \subfloat[Precision (R=8)]{
        \centering
        \includegraphics[trim=25 0 0 25, clip, width=0.253\textwidth]{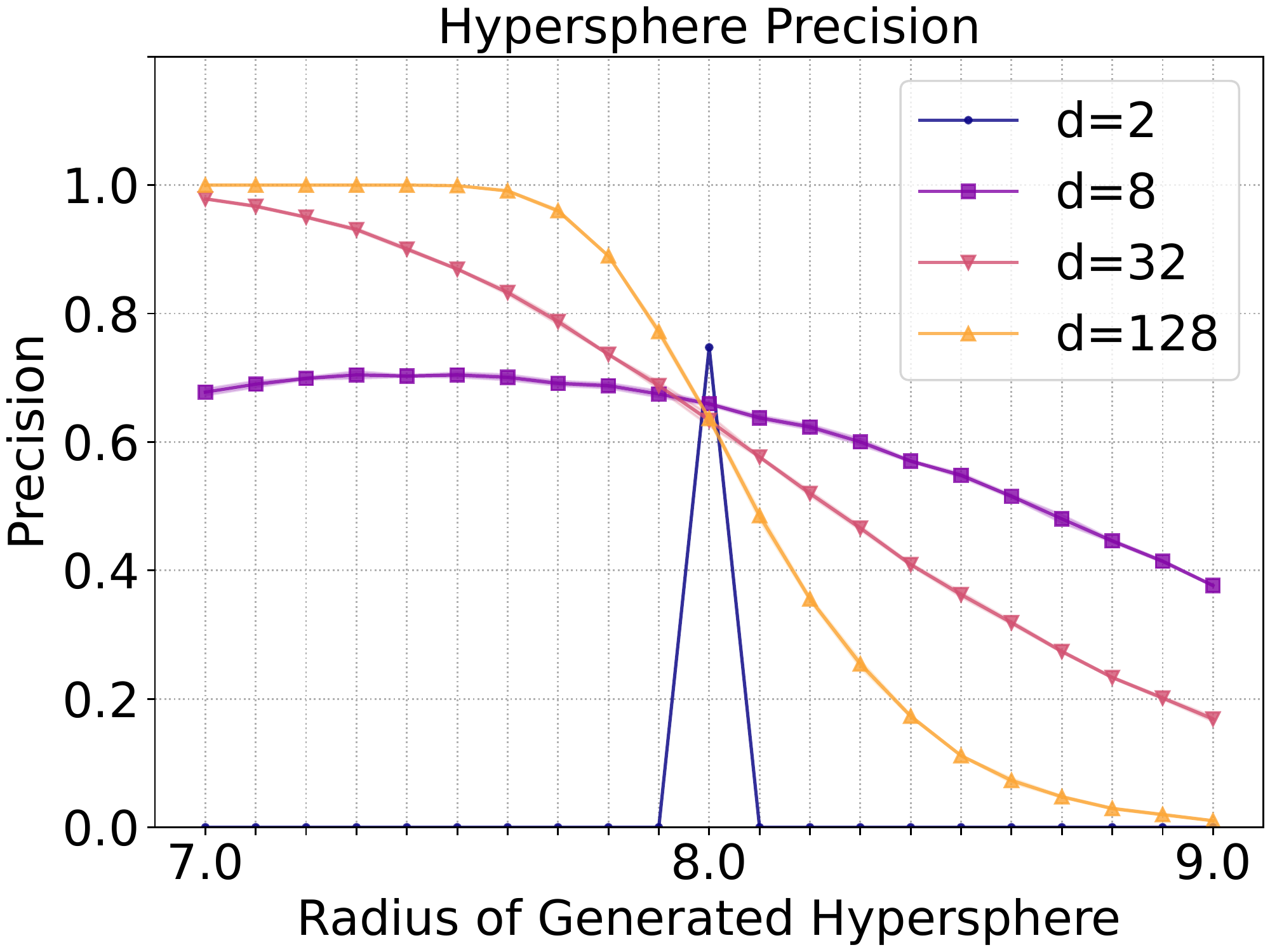}
        \label{fig:sphere_r8_pr_p_k1}}
    \subfloat[Recall (R=8)]{
        \centering
        \includegraphics[trim=64 0 0 25, clip, width=0.235\textwidth]{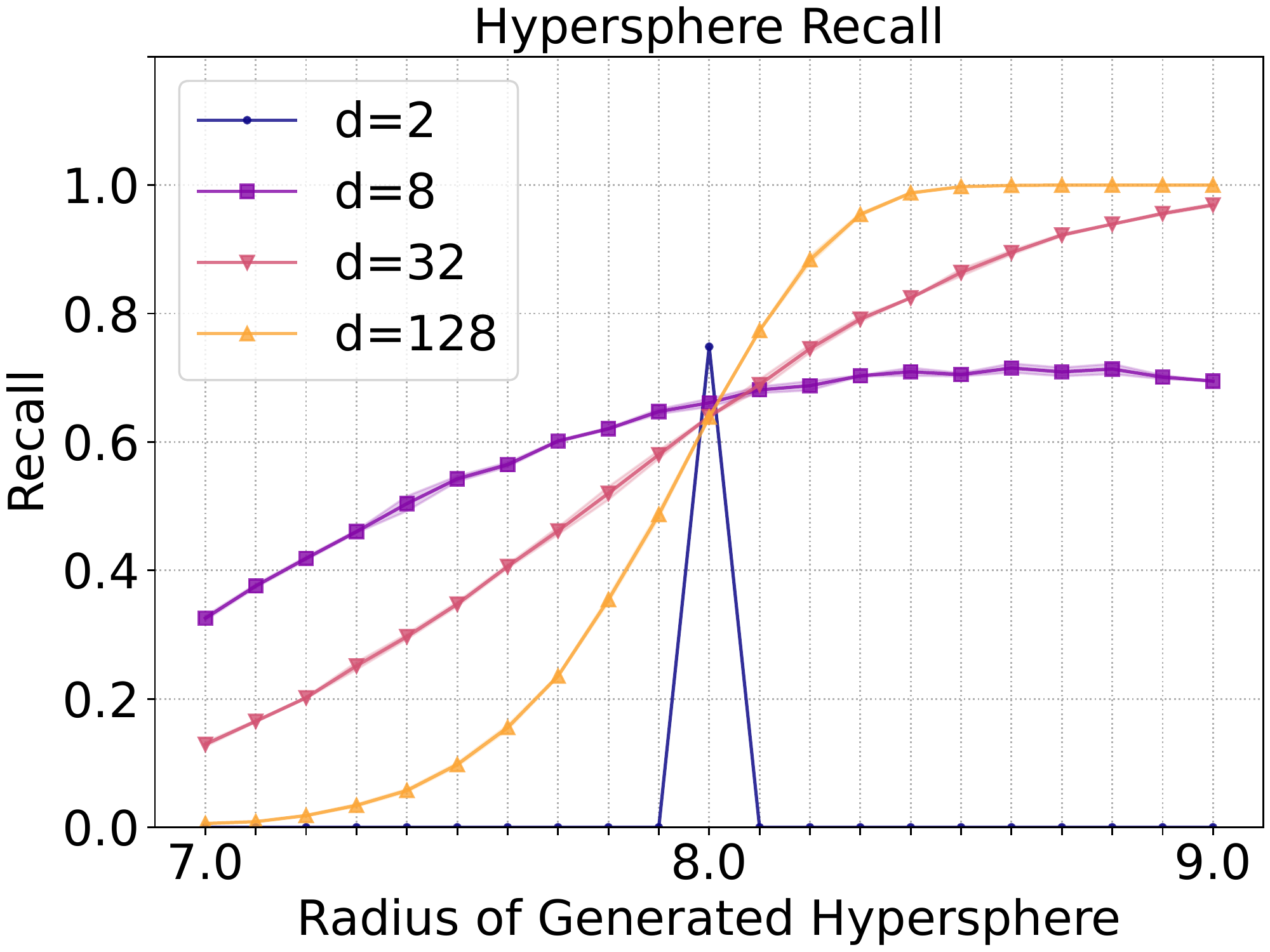}
        \label{fig:sphere_r8_pr_r_k1}}
    \subfloat[symPrecision (R=8)]{
        \centering
        \includegraphics[trim=64 0 0 25, clip, width=0.235\textwidth]{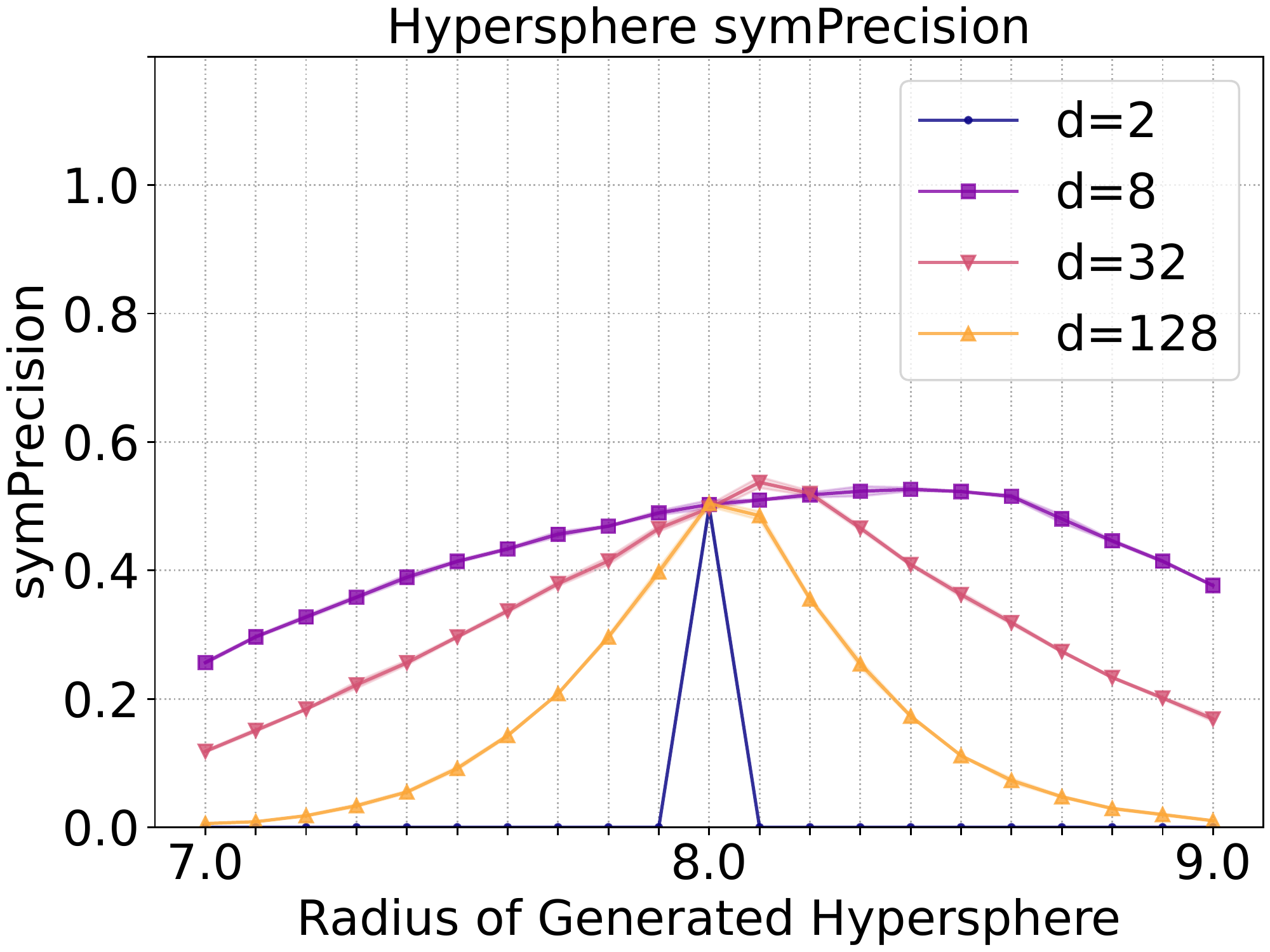}
        \label{fig:sphere_r8_pr_sp_k1}}
    \subfloat[symRecall (R=8)]{
        \centering
        \includegraphics[trim=64 0 0 25, clip, width=0.235\textwidth]{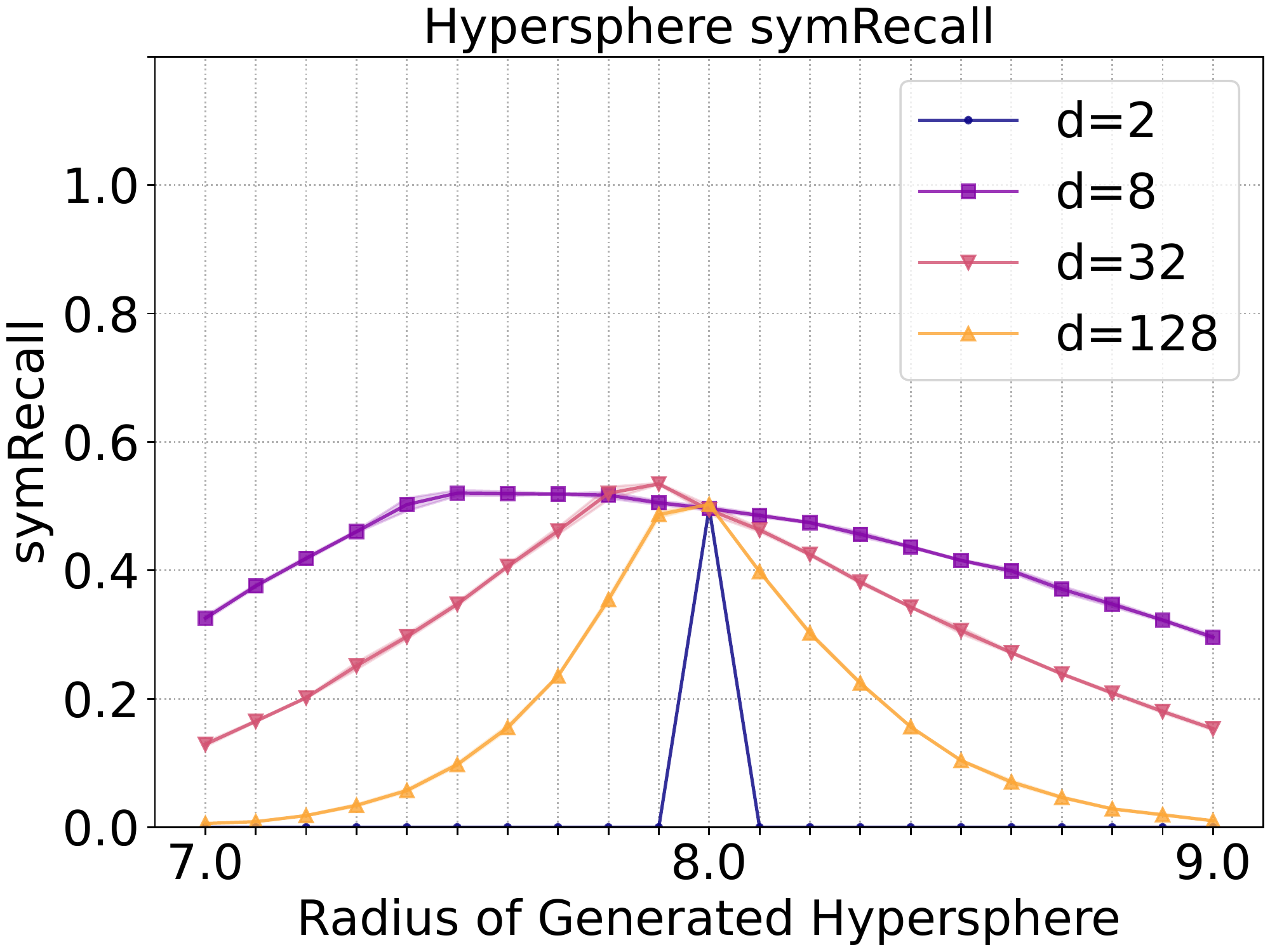}
        \label{fig:sphere_r8_pr_sr_k1}}
    \caption{Hyperspherical reference and generated supports of varying radii at $K=1$ of $K$-nearest-neighbors.}
    \label{fig:sphere_rm_k1}
\end{figure*}

\clearpage
\section{Experiments with Varying K of K-Nearest Neighbors}
\label{sec:varying_k}
\begin{figure*}[h!]
    \centering
    \subfloat[Precision (d=16)]{
        \centering
        \includegraphics[trim=25 0 0 25, clip, width=0.253\textwidth]{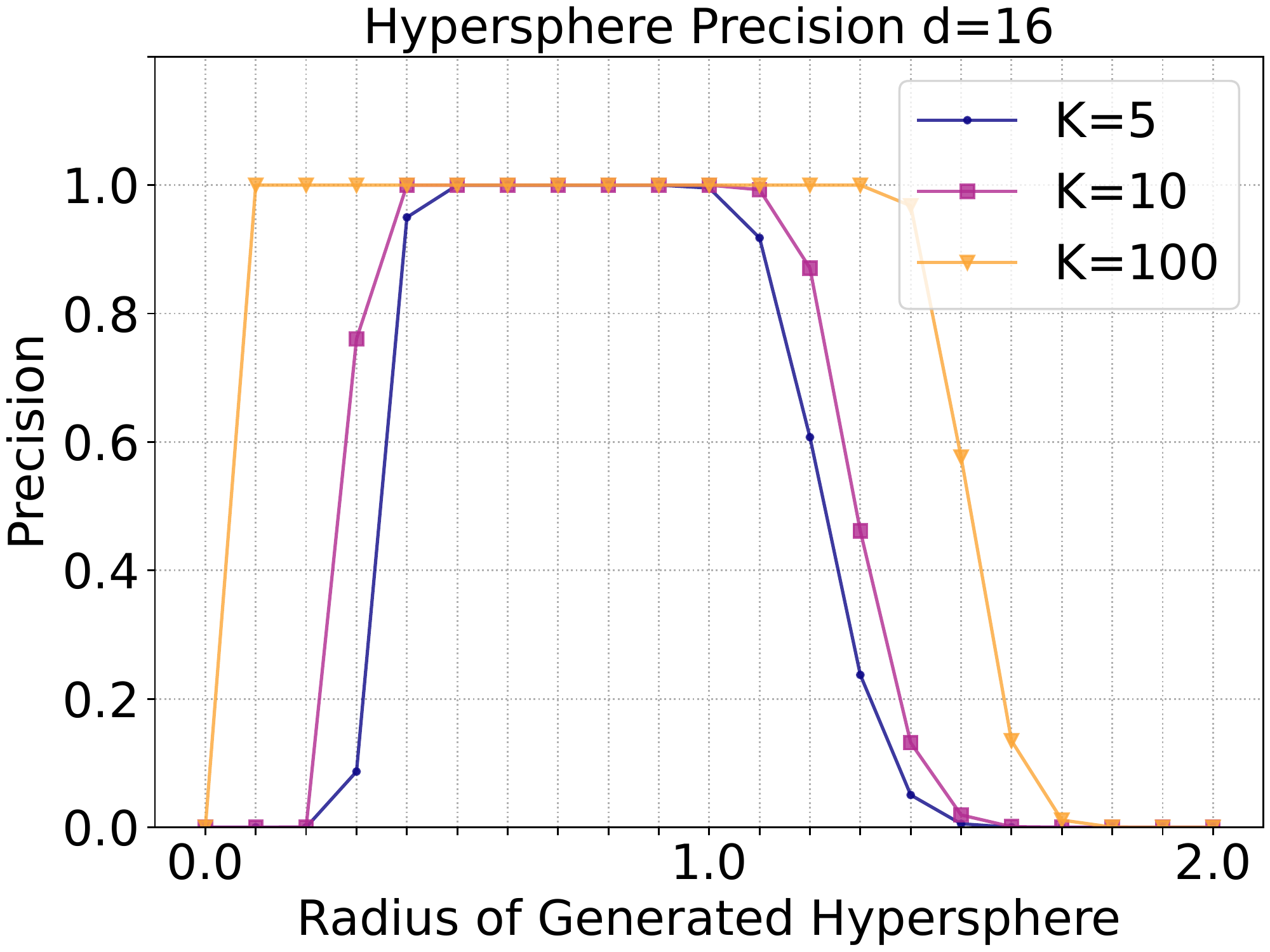}
        \label{fig:knn_sphere_dim16_pr_p}}
    \subfloat[Recall (d=16)]{
        \centering
        \includegraphics[trim=64 0 0 25, clip, width=0.235\textwidth]{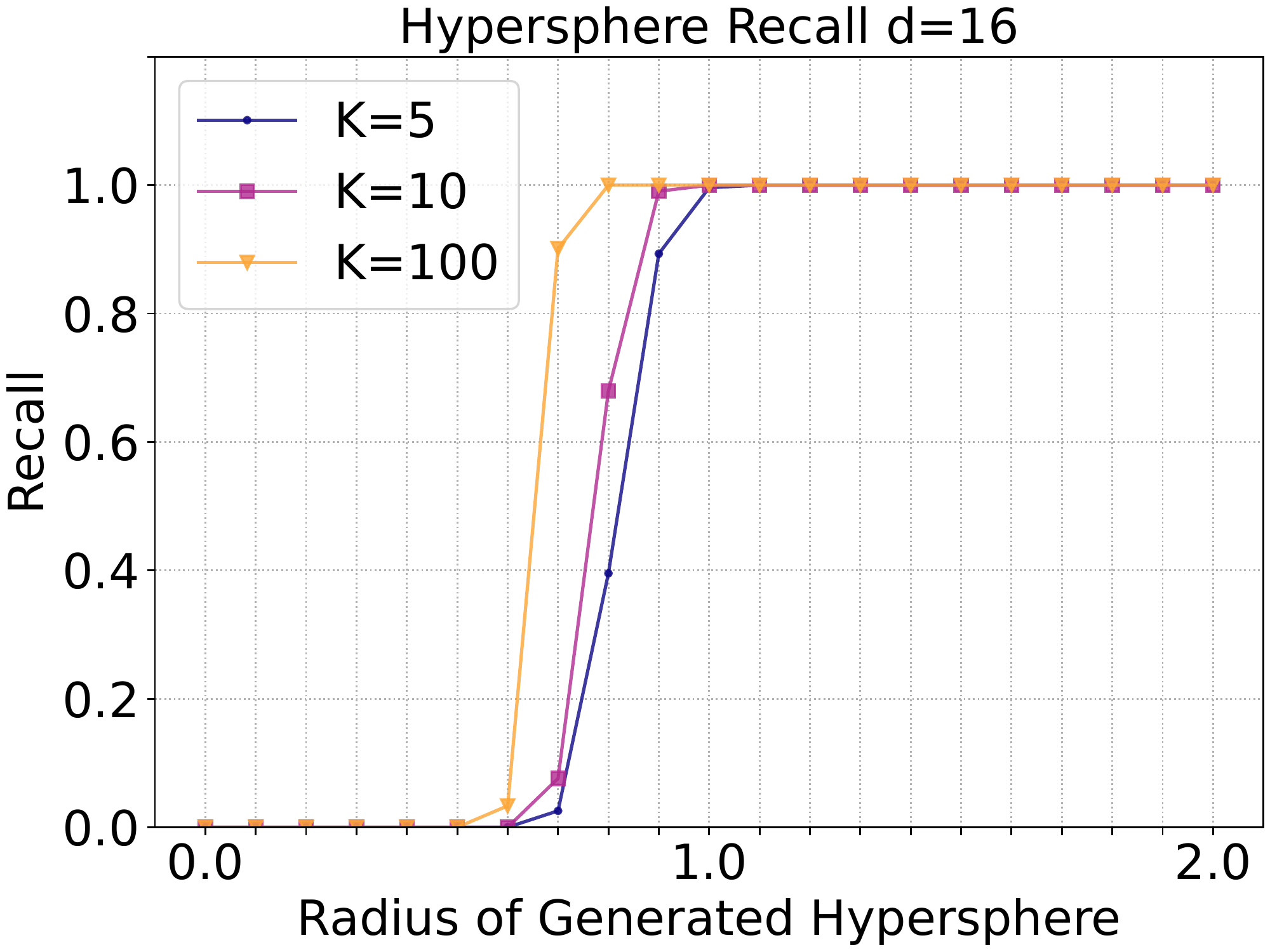}
        \label{fig:knn_sphere_dim16_pr_r}}
    \subfloat[symPrecision (d=16)]{
        \centering
        \includegraphics[trim=64 0 0 25, clip, width=0.235\textwidth]{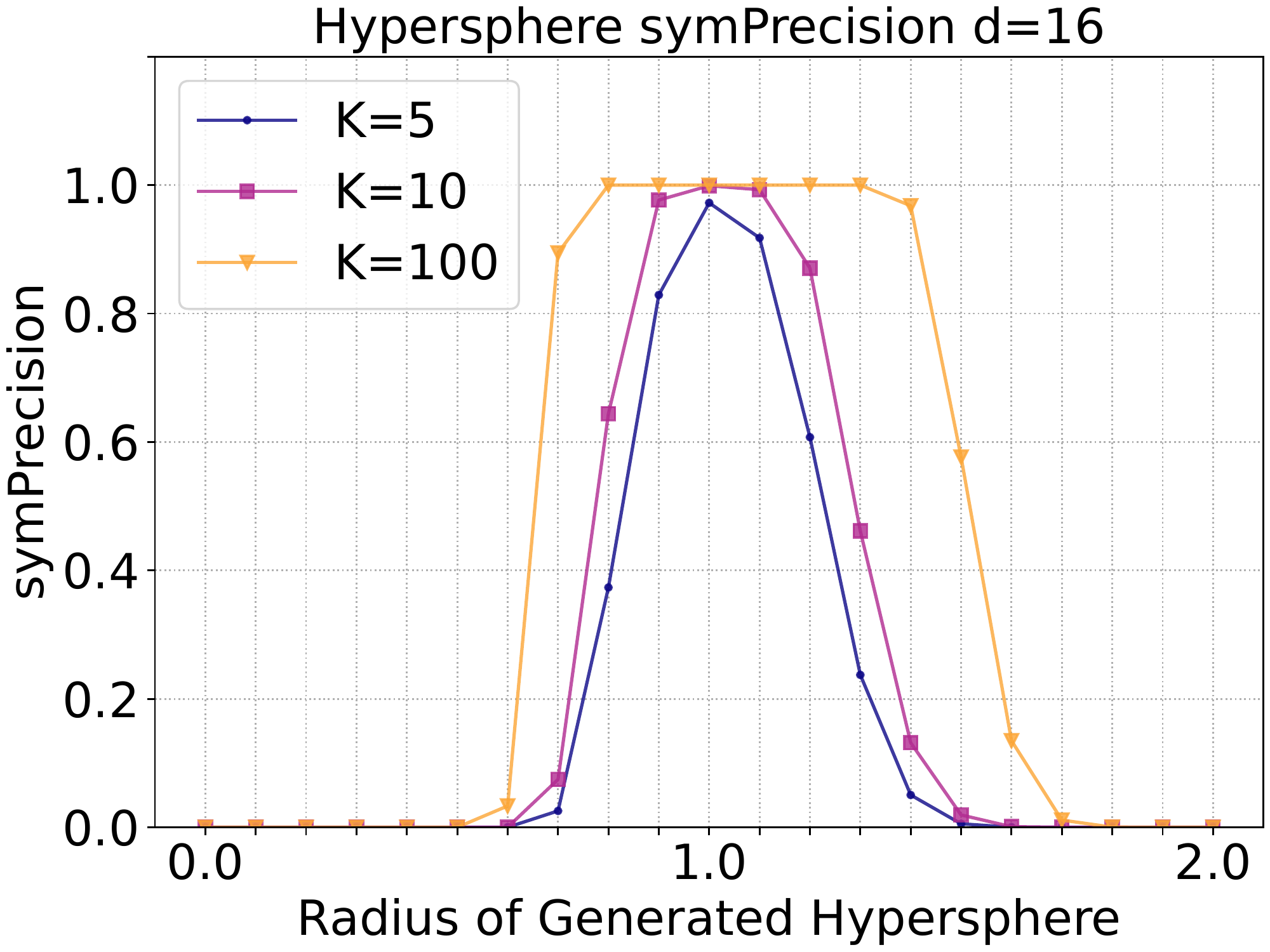}
        \label{fig:knn_sphere_dim16_pr_sp}}
    \subfloat[symRecall (d=16)]{
        \centering
        \includegraphics[trim=64 0 0 25, clip, width=0.235\textwidth]{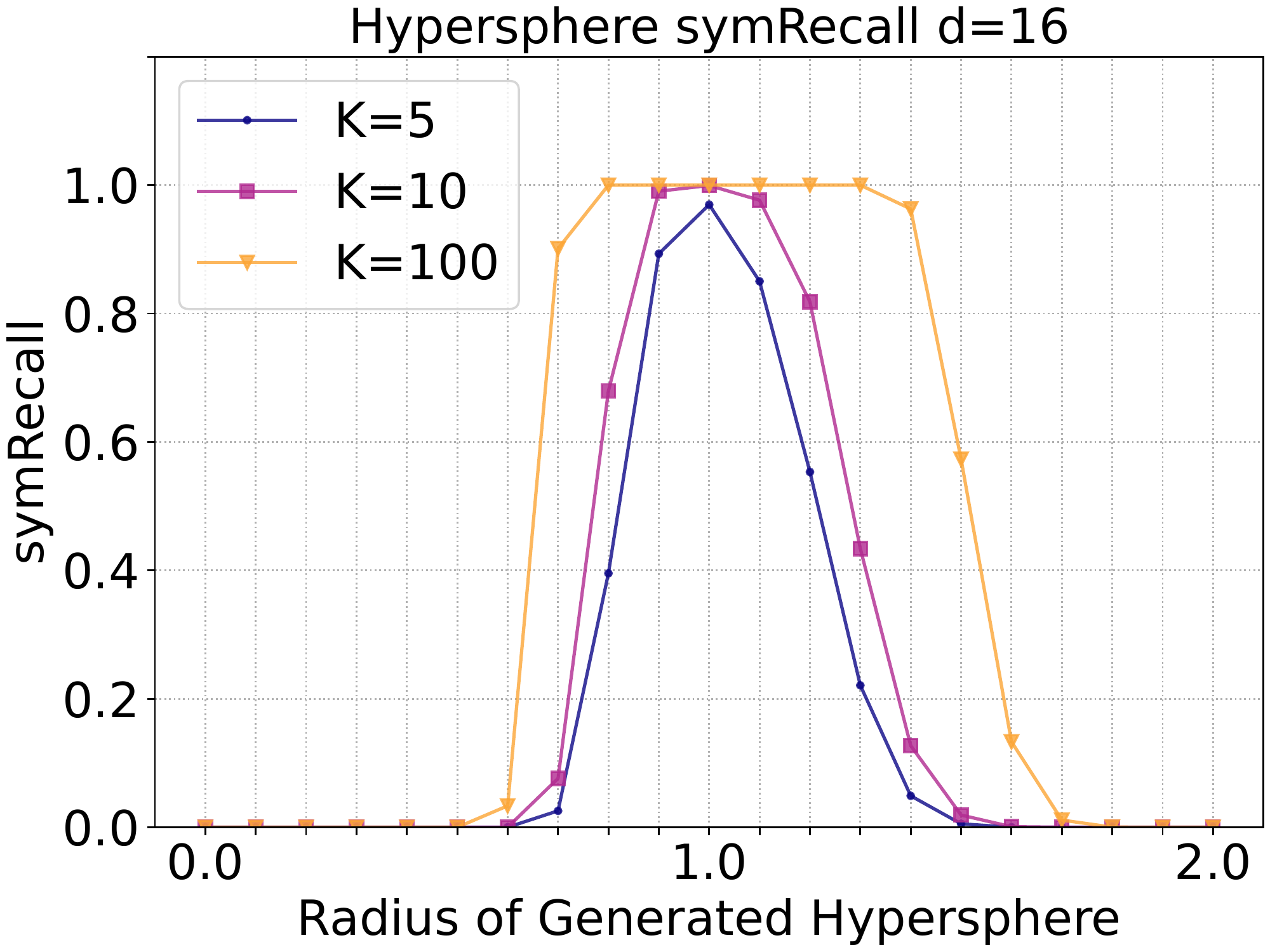}
        \label{fig:knn_sphere_dim16_pr_sr}}\\
    \subfloat[Precision (d=64)]{
        \centering
        \includegraphics[trim=25 0 0 25, clip, width=0.253\textwidth]{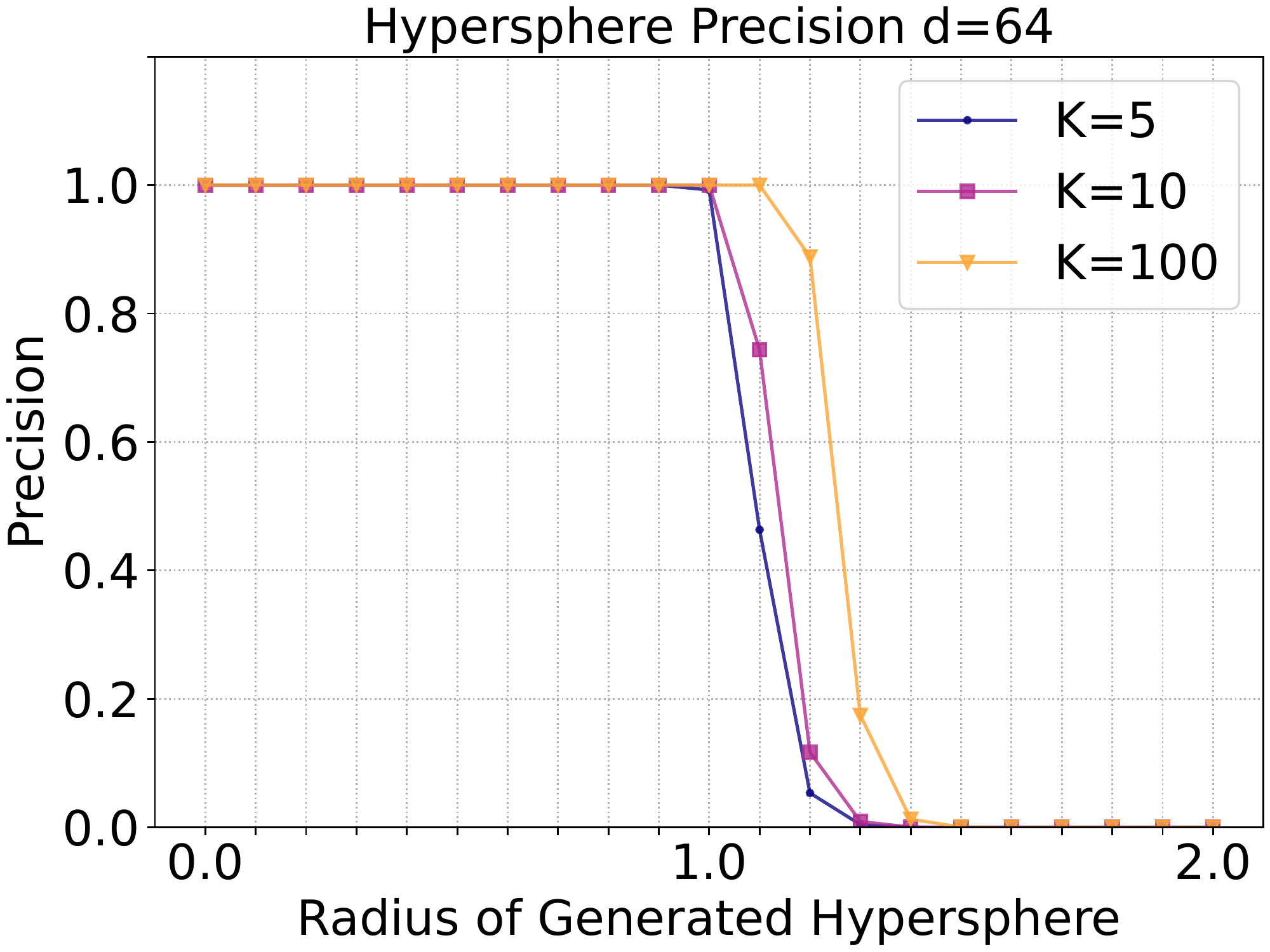}
        \label{fig:knn_sphere_dim64_pr_p}}
    \subfloat[Recall (d=64)]{
        \centering
        \includegraphics[trim=64 0 0 25, clip, width=0.235\textwidth]{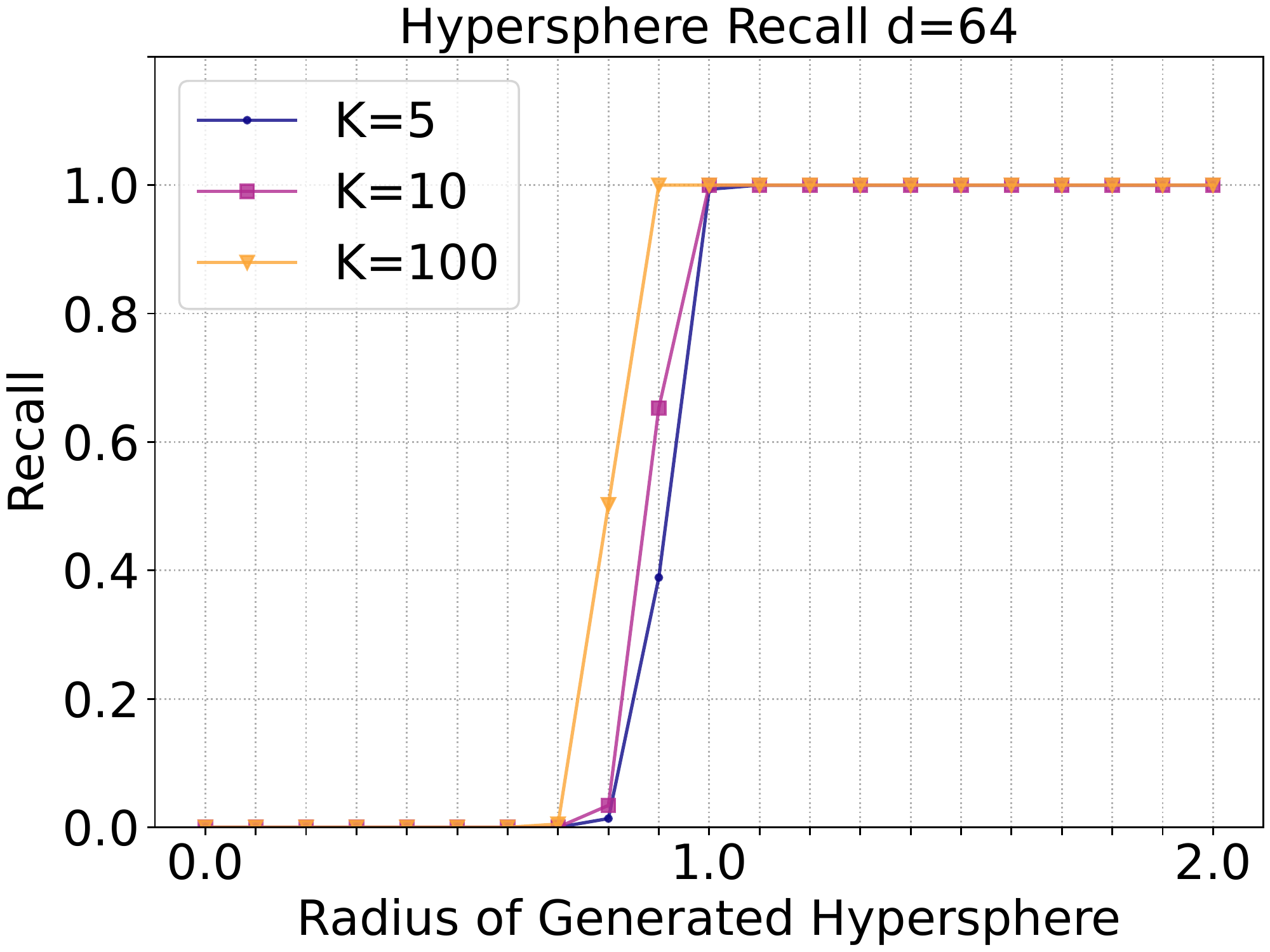}
        \label{fig:knn_sphere_dim64_pr_r}}
    \subfloat[symPrecision (d=64)]{
        \centering
        \includegraphics[trim=64 0 0 25, clip, width=0.235\textwidth]{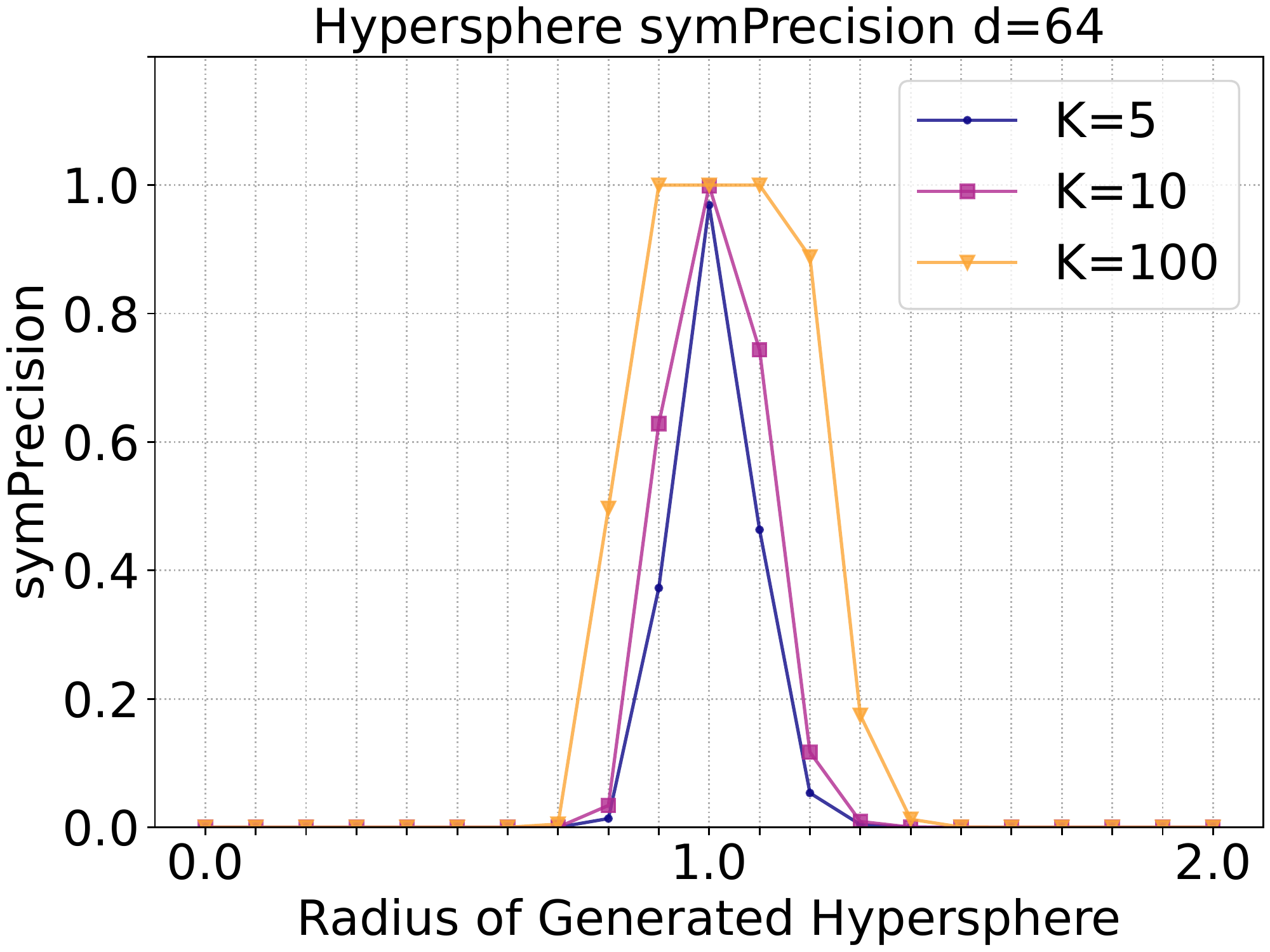}
        \label{fig:knn_sphere_dim64_pr_sp}}
    \subfloat[symRecall (d=64)]{
        \centering
        \includegraphics[trim=64 0 0 25, clip, width=0.235\textwidth]{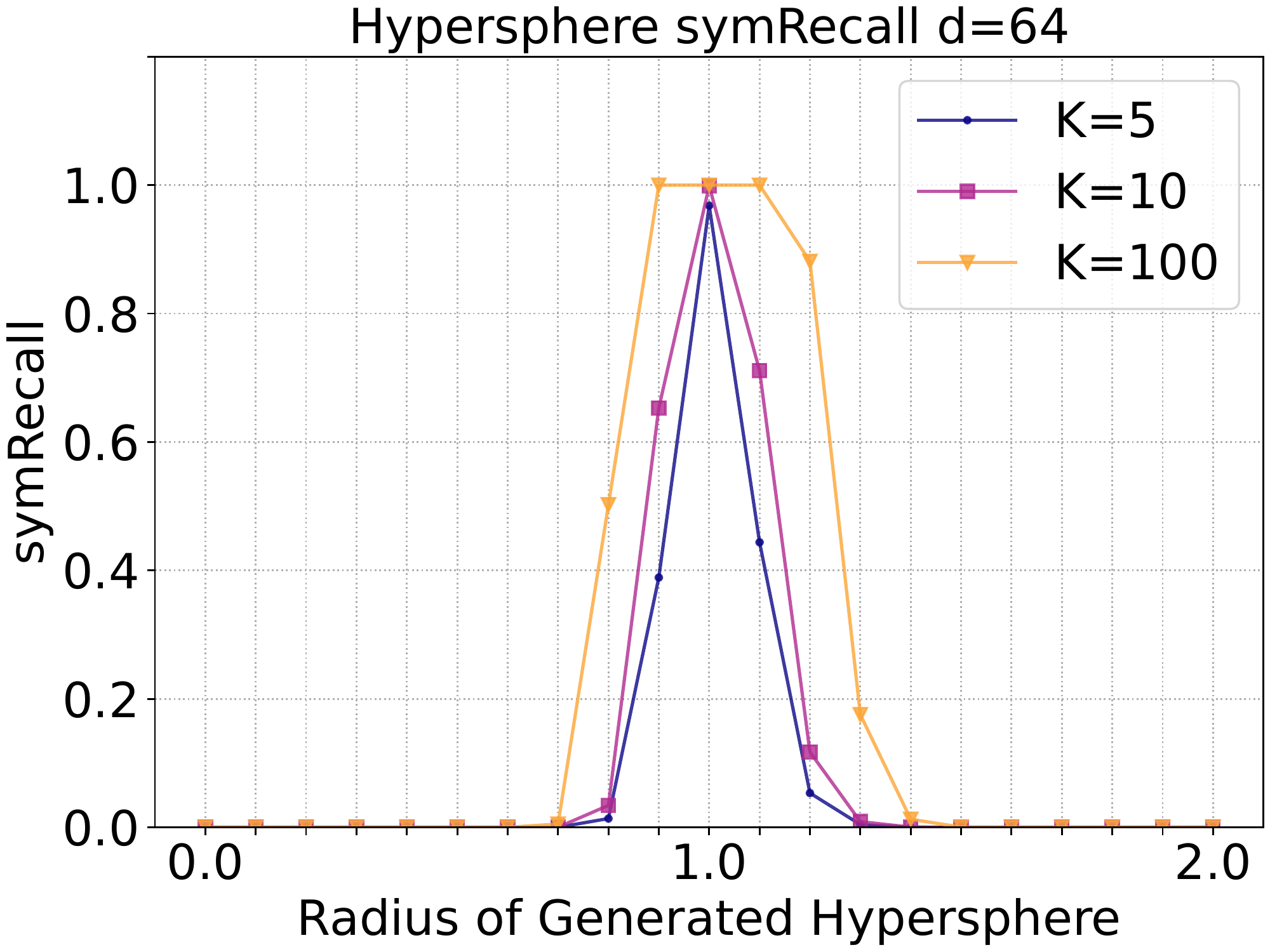}
        \label{fig:knn_sphere_dim64_pr_sr}}
    \caption{Hyperspherical reference and generated supports of varying $K$-nearest-neighbors approximations at radius 1. Larger $K$ results in loss of resolution in the manifold approximation, hence the saturated behavior near the reference support (near R=1). The asymmetry in Precision and Recall emerges regardless of $K$, consistent with the proposed theory.}
    \label{fig:knn_sphere}
\end{figure*}

\clearpage
\section{Experiments with Varying Number of Samples}
\label{sec:varying_num_samples}
\begin{figure*}[h!]
    \centering
    \subfloat[Precision (d=16)]{
        \centering
        \includegraphics[trim=25 0 0 25, clip, width=0.253\textwidth]{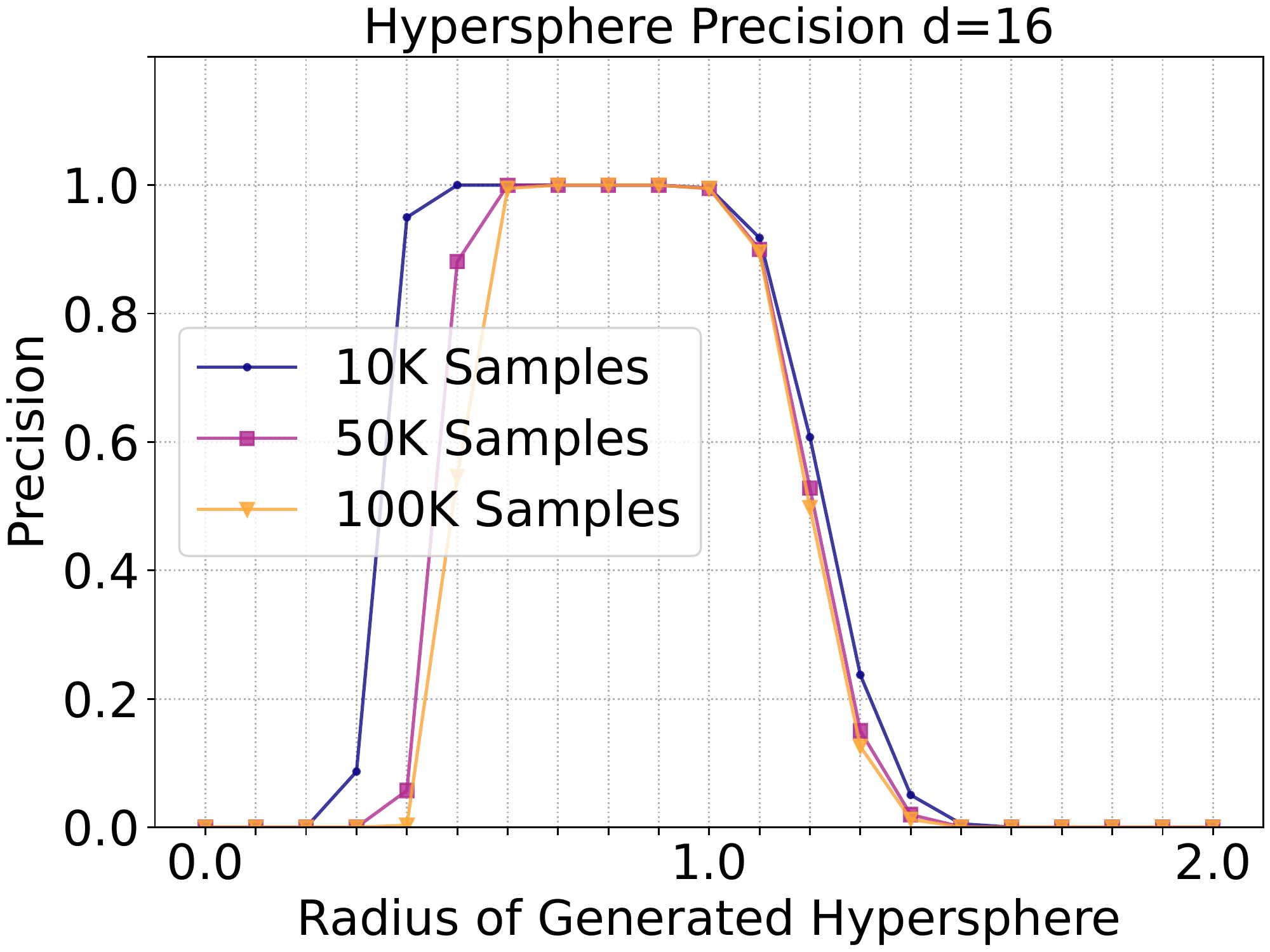}
        \label{fig:num_samples_sphere_dim16_pr_p}}
    \subfloat[Recall (d=16)]{
        \centering
        \includegraphics[trim=64 0 0 25, clip, width=0.235\textwidth]{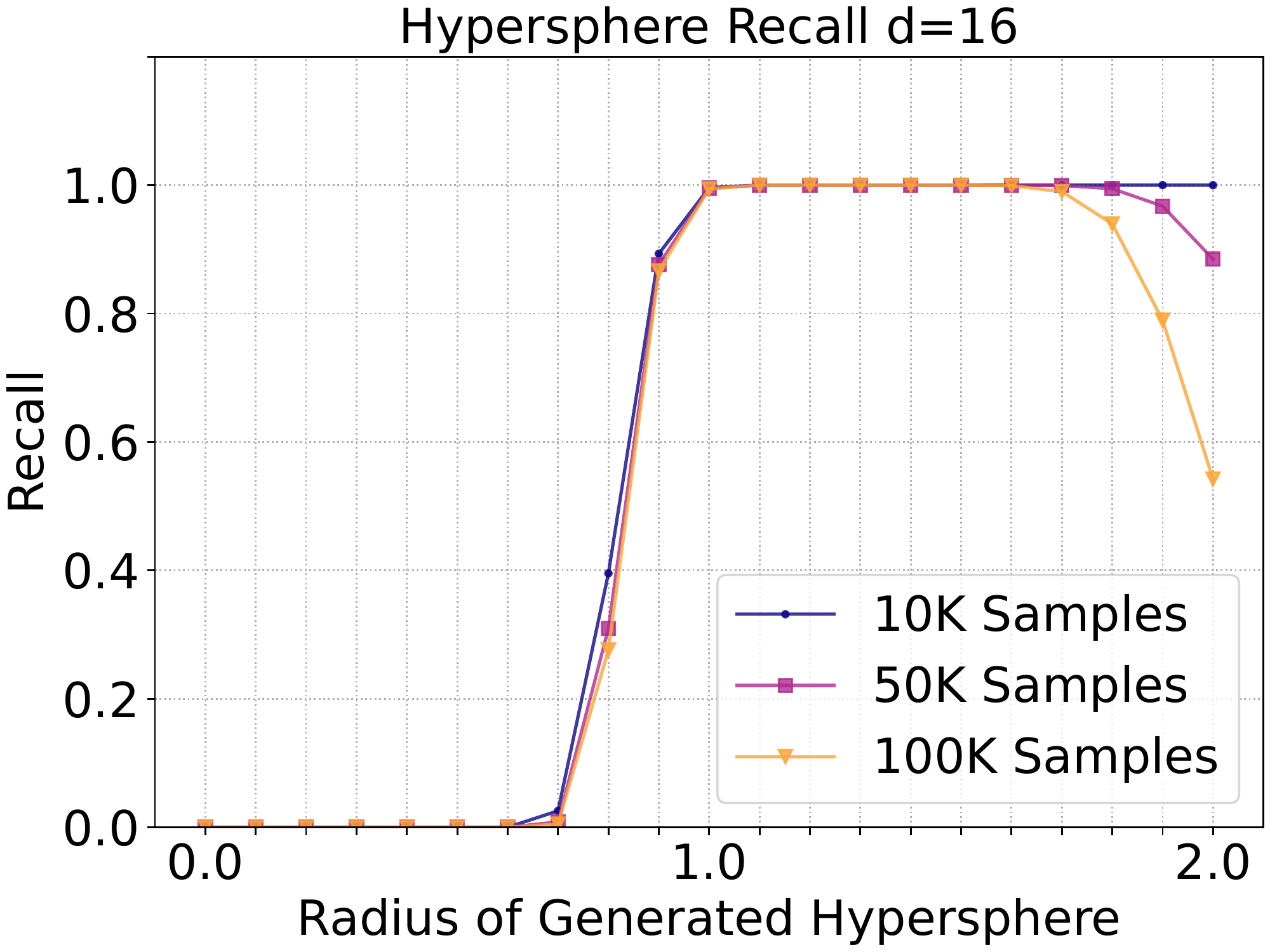}
        \label{fig:num_samples_sphere_dim16_pr_r}}
    \subfloat[symPrecision (d=16)]{
        \centering
        \includegraphics[trim=64 0 0 25, clip, width=0.235\textwidth]{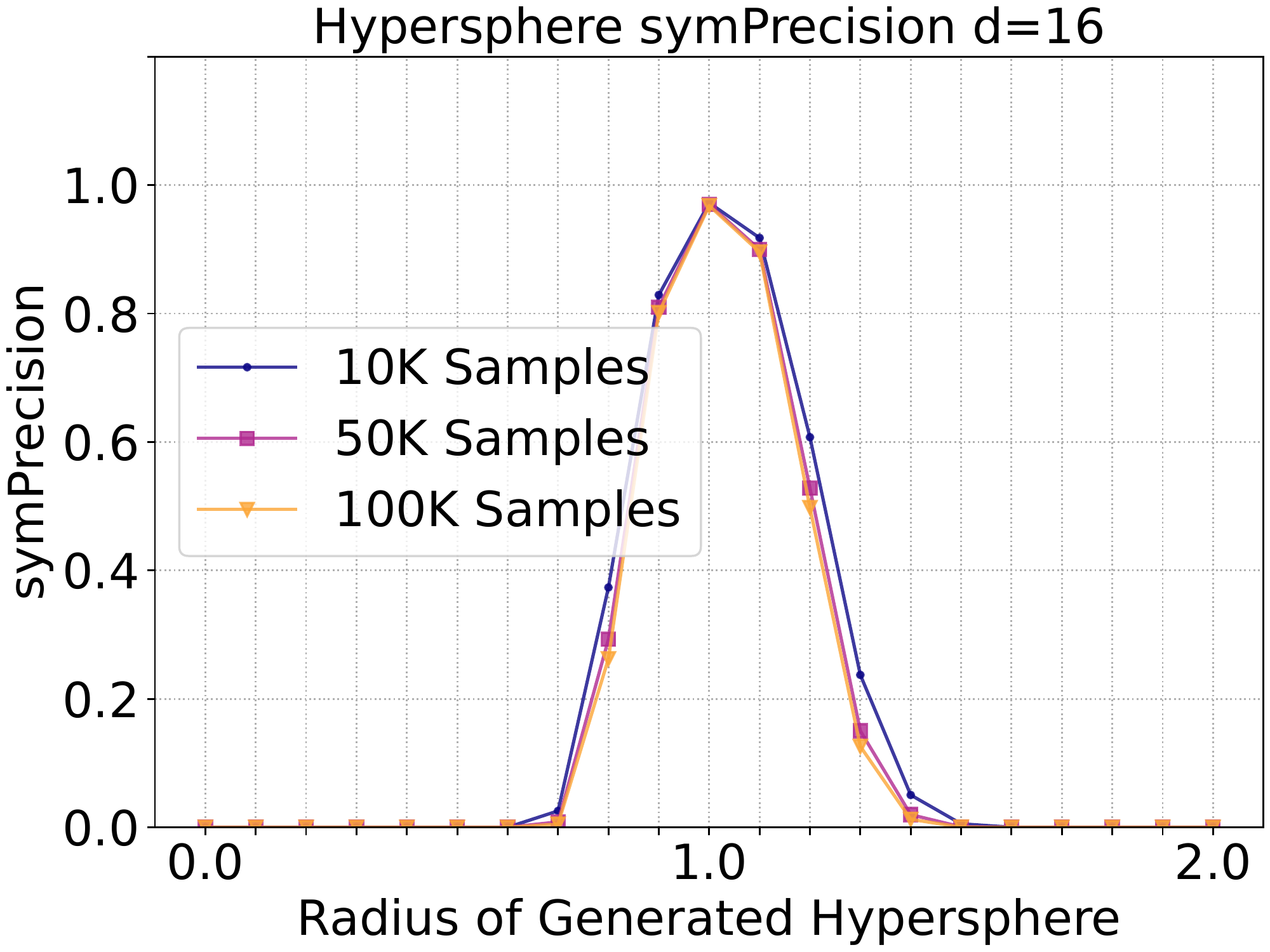}
        \label{fig:num_samples_sphere_dim16_pr_sp}}
    \subfloat[symRecall (d=16)]{
        \centering
        \includegraphics[trim=64 0 0 25, clip, width=0.235\textwidth]{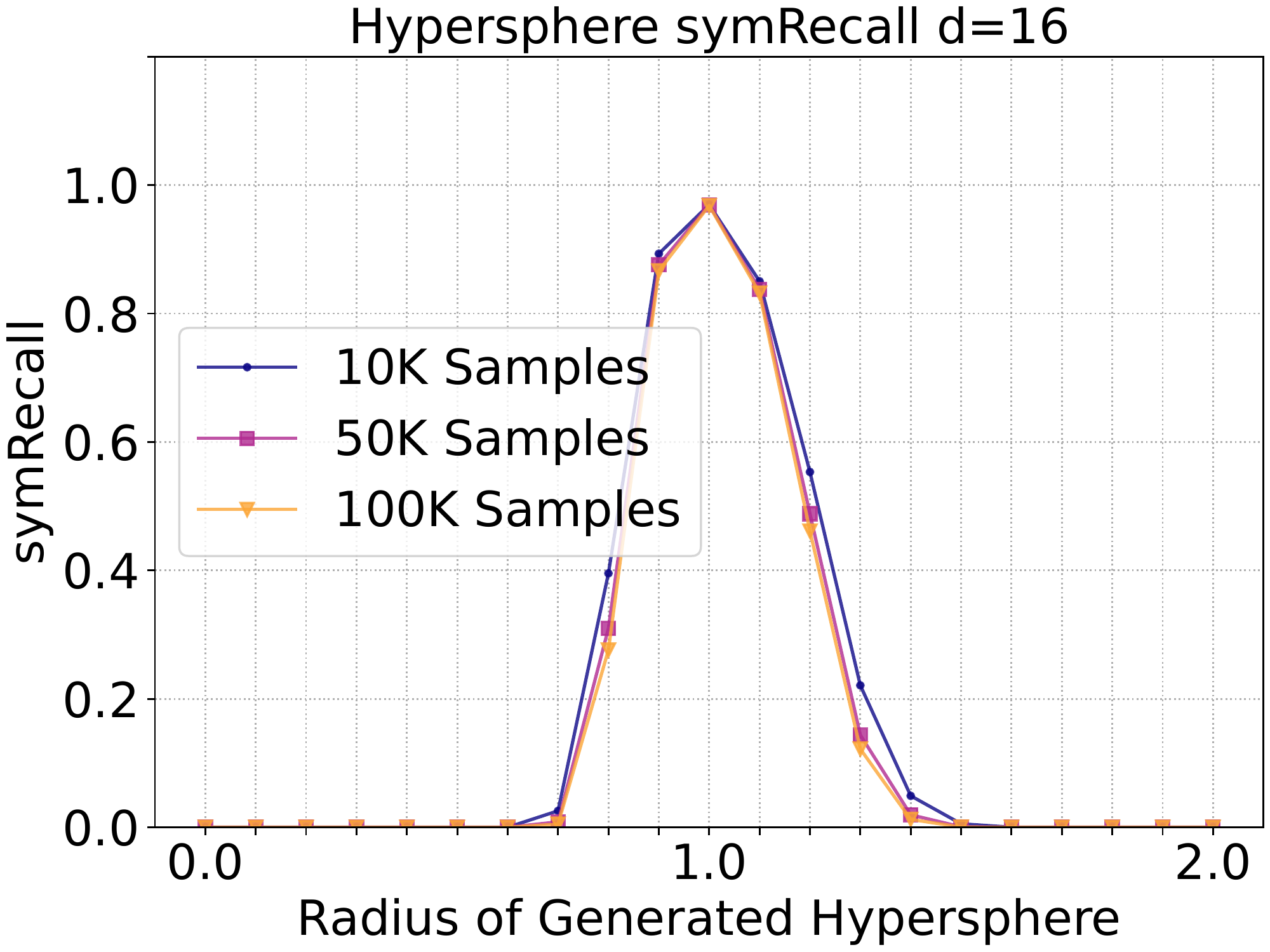}
        \label{fig:num_samples_sphere_dim16_pr_sr}}\\
    \subfloat[Precision (d=64)]{
        \centering
        \includegraphics[trim=25 0 0 25, clip, width=0.253\textwidth]{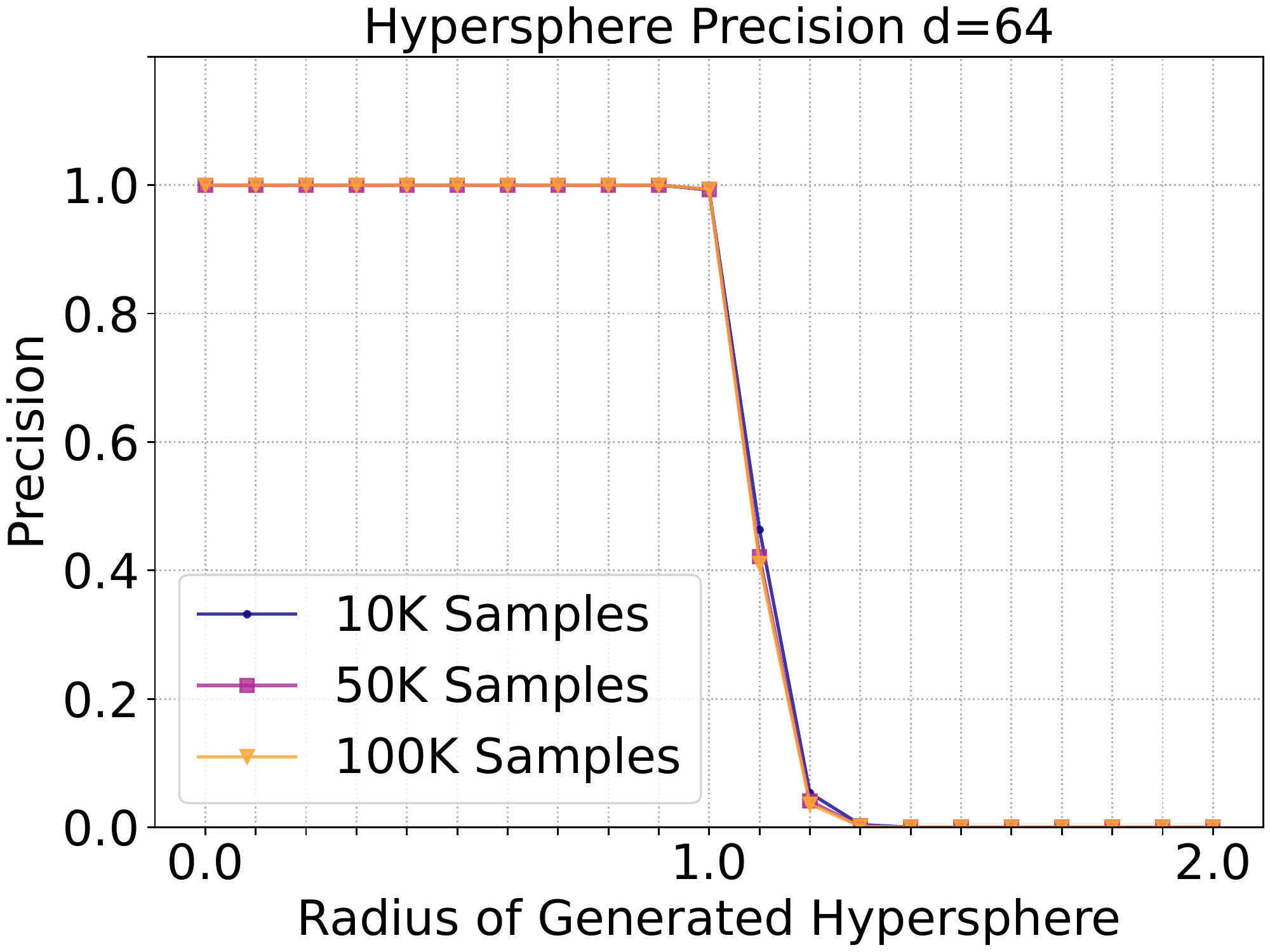}
        \label{fig:num_samples_sphere_dim64_pr_p}}
    \subfloat[Recall (d=64)]{
        \centering
        \includegraphics[trim=64 0 0 25, clip, width=0.235\textwidth]{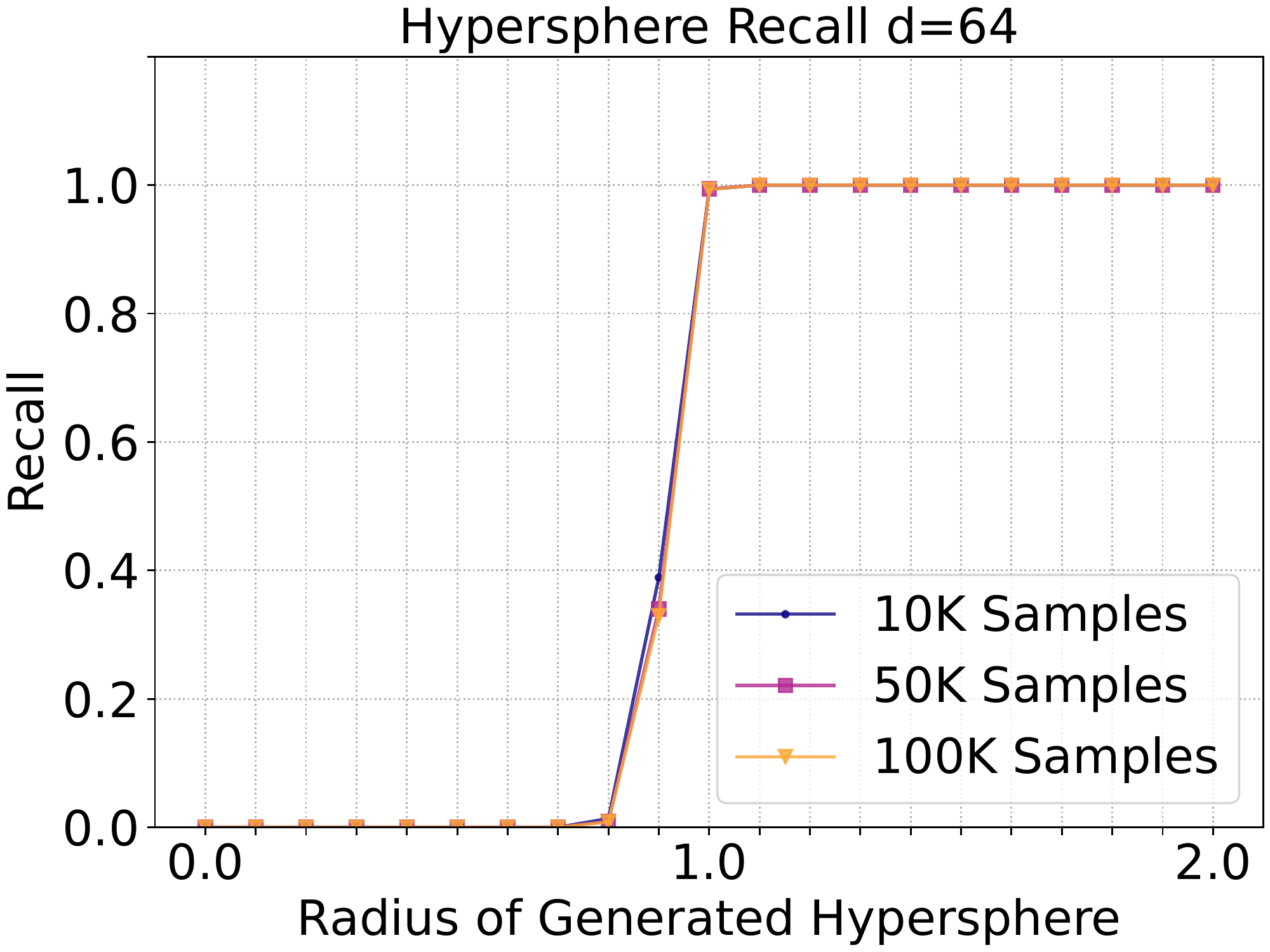}
        \label{fig:num_samples_sphere_dim64_pr_r}}
    \subfloat[symPrecision (d=64)]{
        \centering
        \includegraphics[trim=64 0 0 25, clip, width=0.235\textwidth]{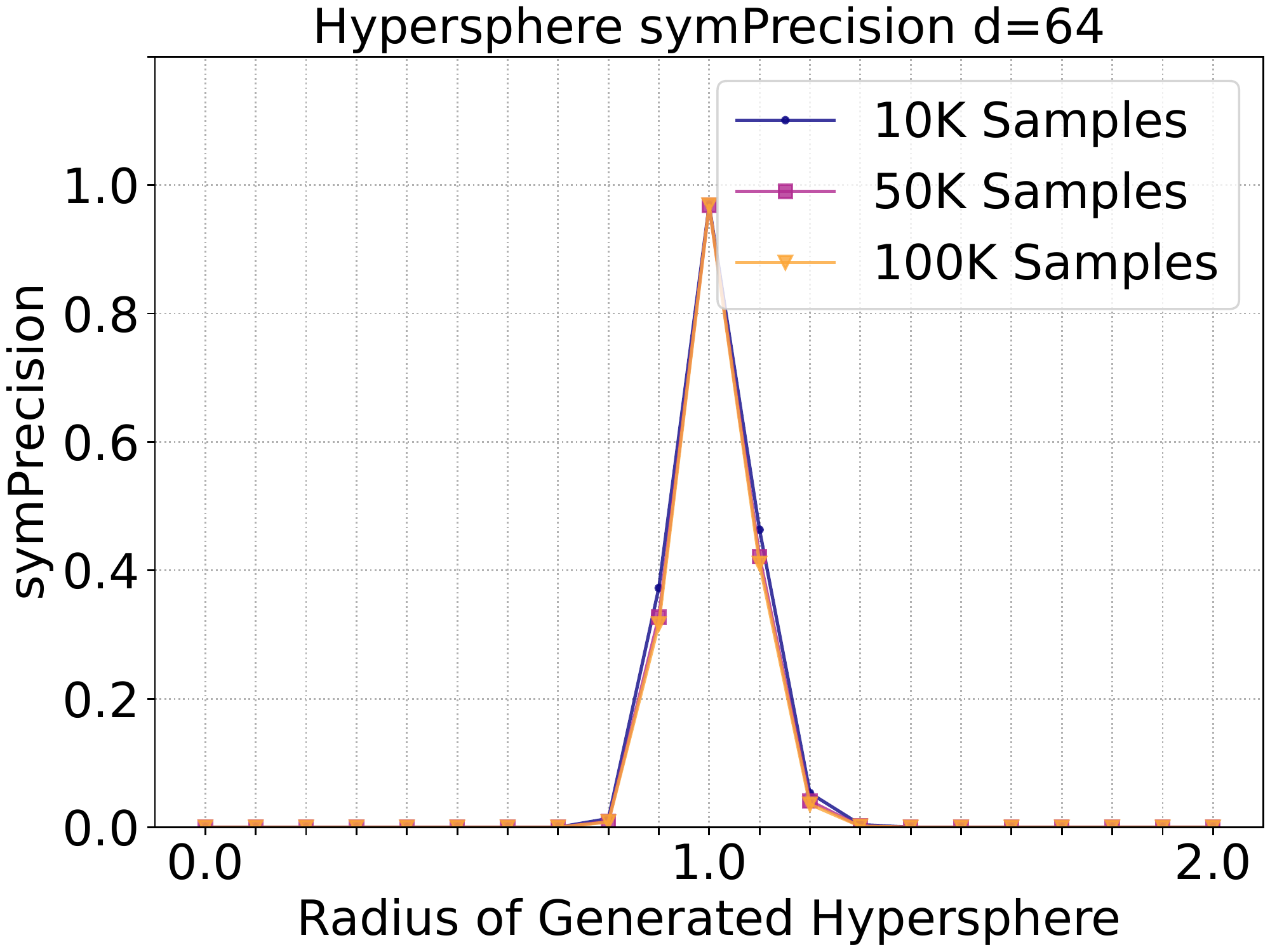}
        \label{fig:num_samples_sphere_dim64_pr_sp}}
    \subfloat[symRecall (d=64)]{
        \centering
        \includegraphics[trim=64 0 0 25, clip, width=0.235\textwidth]{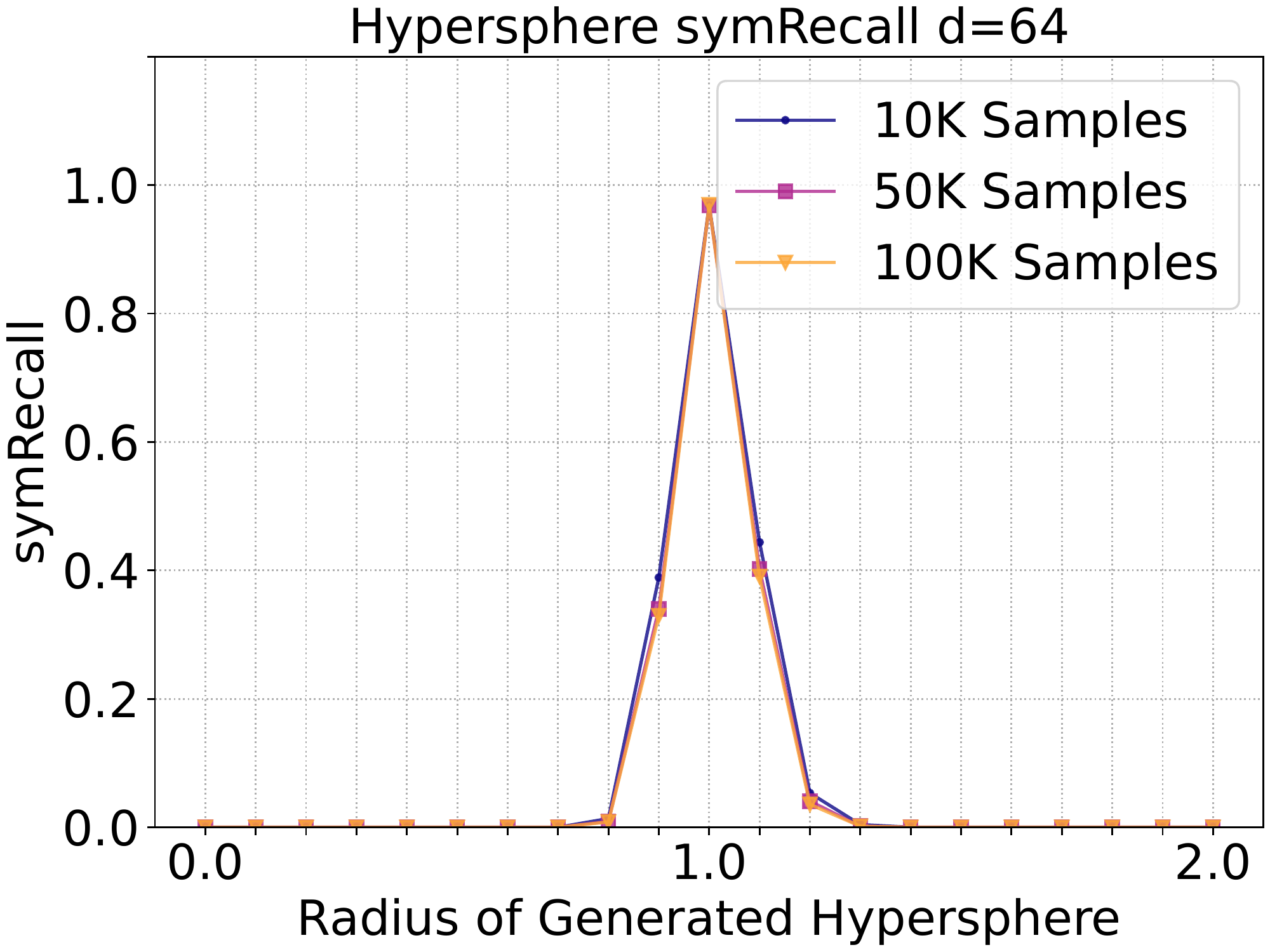}
        \label{fig:num_samples_sphere_dim64_pr_sr}}
    \caption{Hyperspherical reference and generated supports of varying number of samples at radius 1 and $K=5$. Larger number of samples in lower dimensions reduces the asymmetry (note the shrinking tails in $d=16$), however, in higher dimensions it has little to no effect (for $d>64$ the change becomes visually imperceptible in the plots). This is consistent with the proposed theory, which suggests the asymmetry emerges unless the number of samples grows at least exponentially in the number of dimensions.}
    \label{fig:num_samples_sphere}
\end{figure*}

\clearpage
\section{Experiments with Random Embedding for CelebA and CIFAR10}
\label{sec:random_feat}
\begin{figure*}[h!]
    \centering
    \subfloat[Fidelity (CelebA)]{
        \centering
        \includegraphics[trim=25 0 0 25, clip, width=0.253\textwidth]{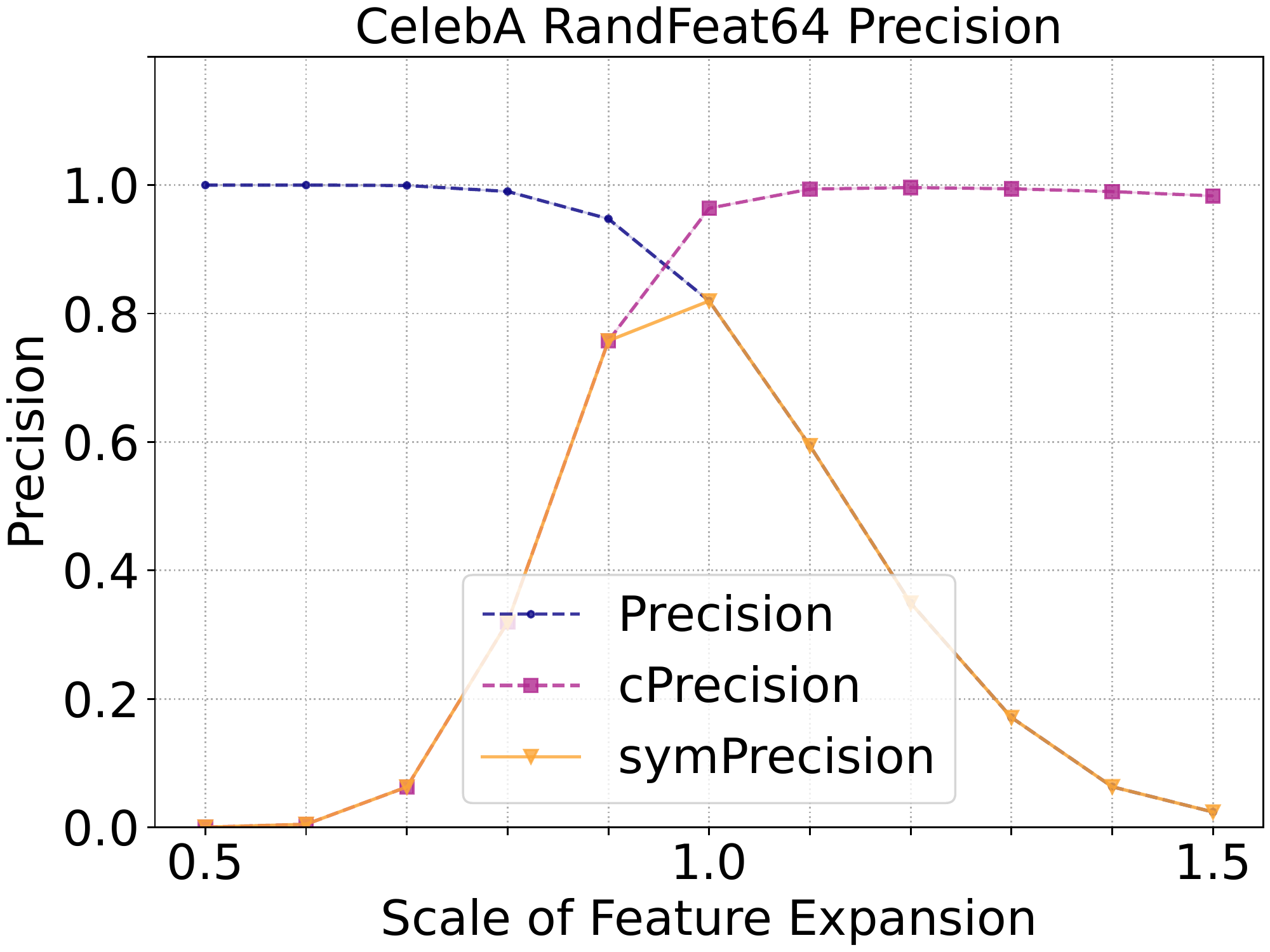}
        \label{fig:scale_embed_rand64_celeba_p}}
    \subfloat[Diversity (CelebA)]{
        \centering
        \includegraphics[trim=64 0 0 25, clip, width=0.235\textwidth]{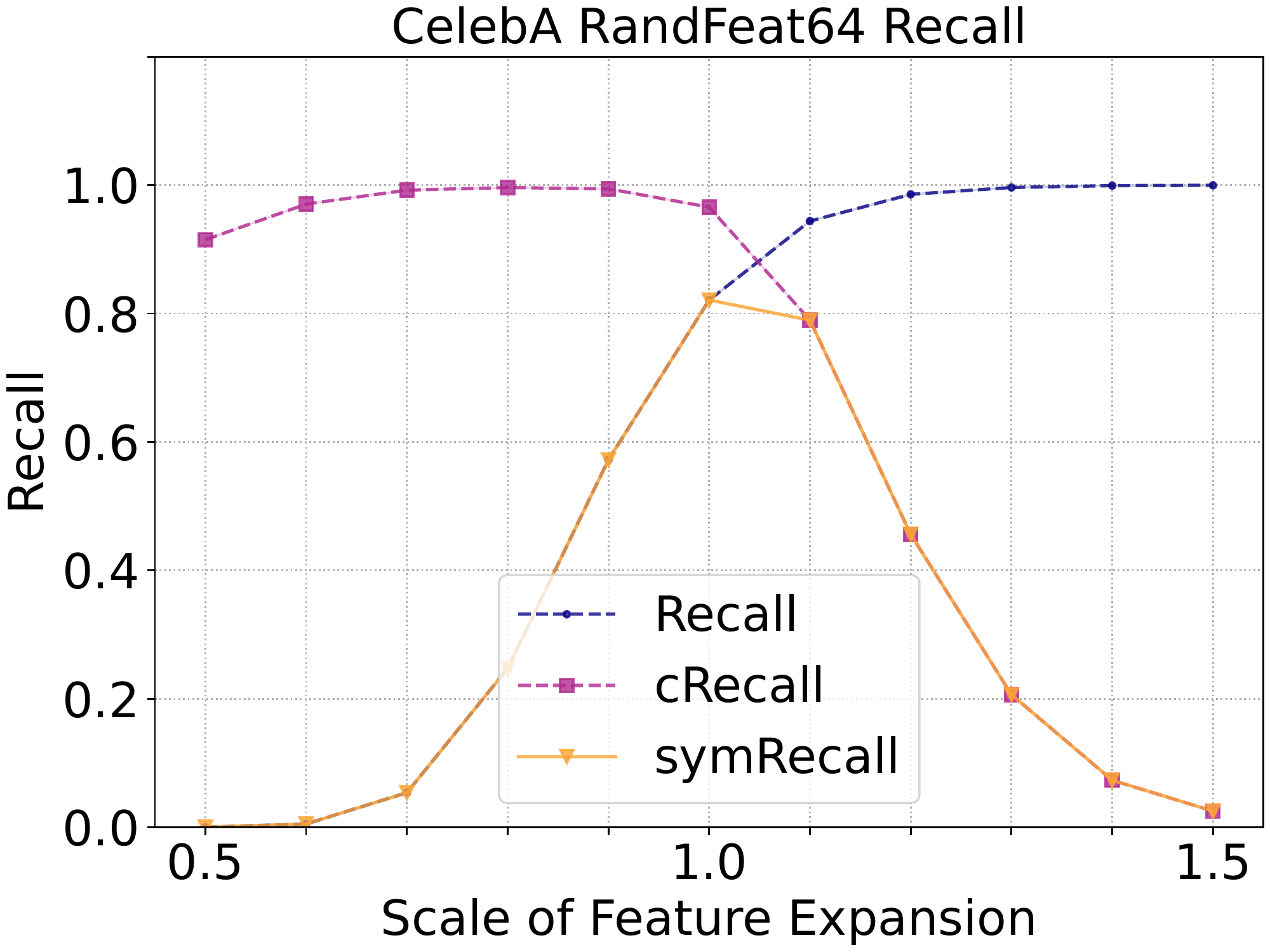}
        \label{fig:scale_embed_rand64_celeba_r}}
        \subfloat[Fidelity (CIFAR10)]{
        \centering
        \includegraphics[trim=64 0 0 25, clip, width=0.235\textwidth]{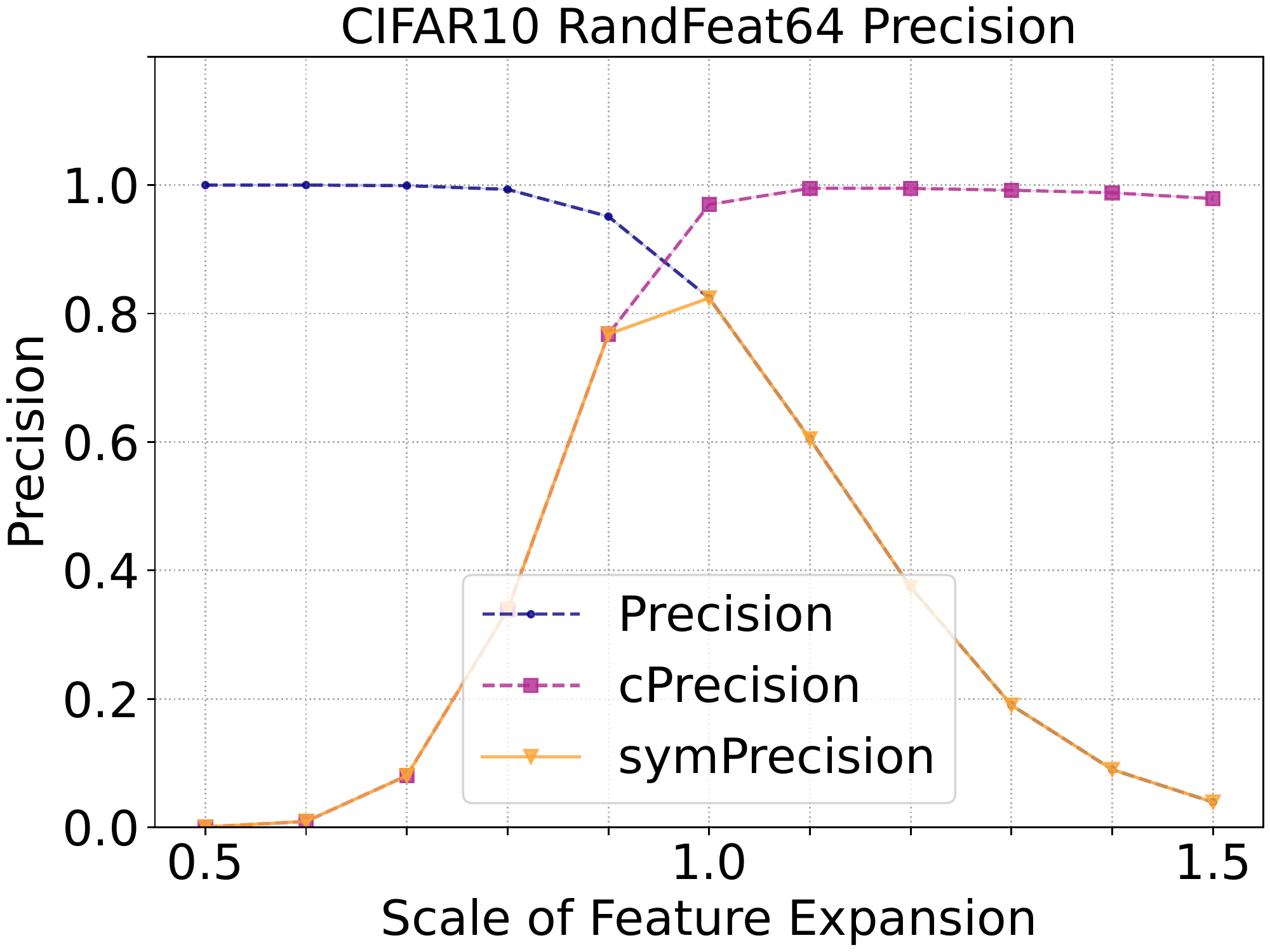}
        \label{fig:scale_embed_rand64_cifar_p}}
    \subfloat[Diversity (CIFAR10)]{
        \centering
        \includegraphics[trim=64 0 0 25, clip, width=0.235\textwidth]{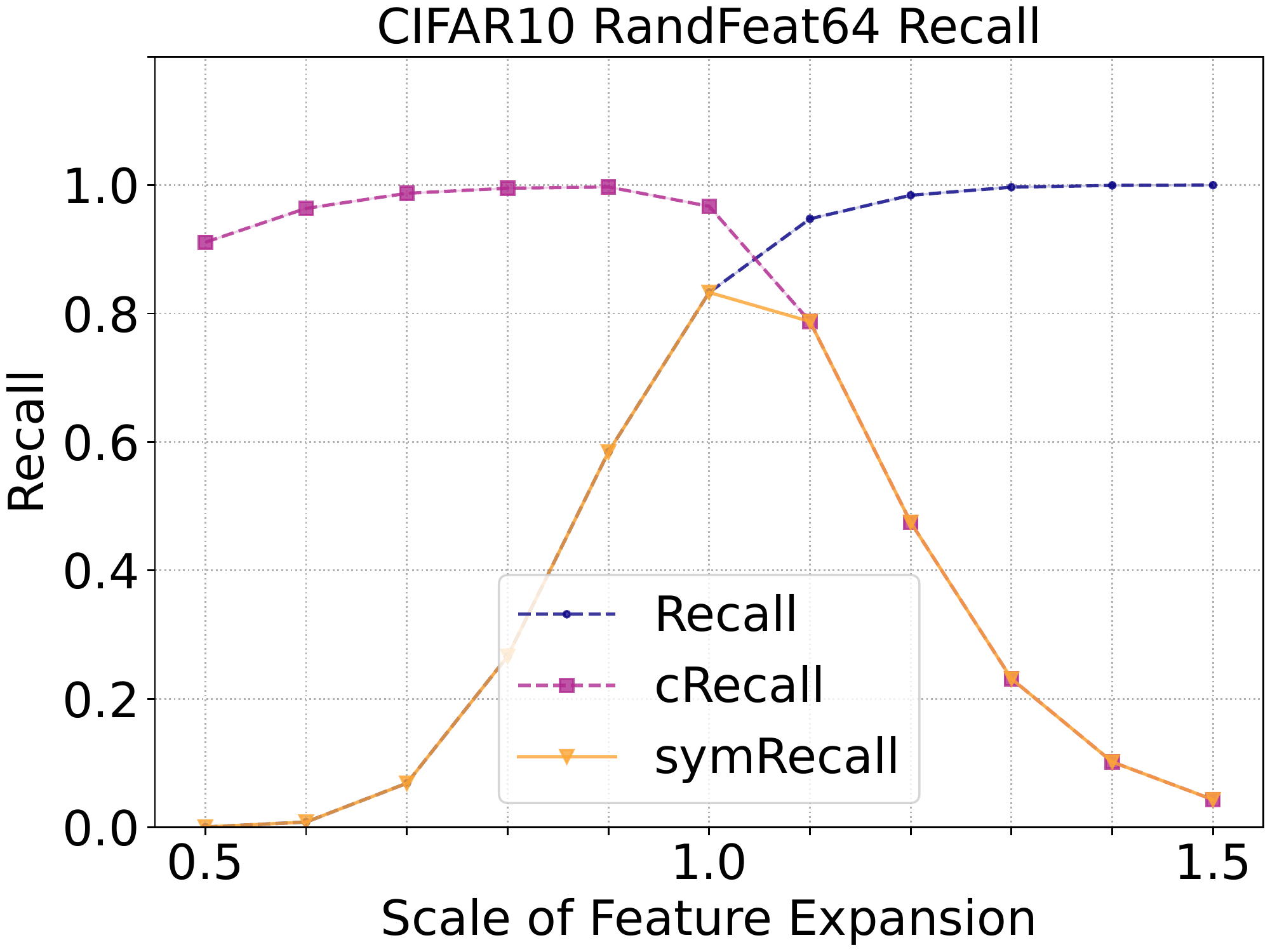}
        \label{fig:scale_embed_rand64_cifar_r}}
    \caption{Repeating the scaling feature space experiment of~\cref{sec:exp_scale} using the random R64 embedding of images as proposed by~\citet{naeem2020density-coverage} instead of the common pretrained VGG16~\cite{kynkaanniemi2019improved-precision-recall}. The results are consistent with~\cref{fig:scale_metrics}.}
    \label{fig:scale_embed_rand64}
\end{figure*}
\begin{figure*}[h!]
    \centering
    \subfloat[Fidelity (CelebA)]{
        \centering
        \includegraphics[trim=25 0 0 25, clip, width=0.253\textwidth]{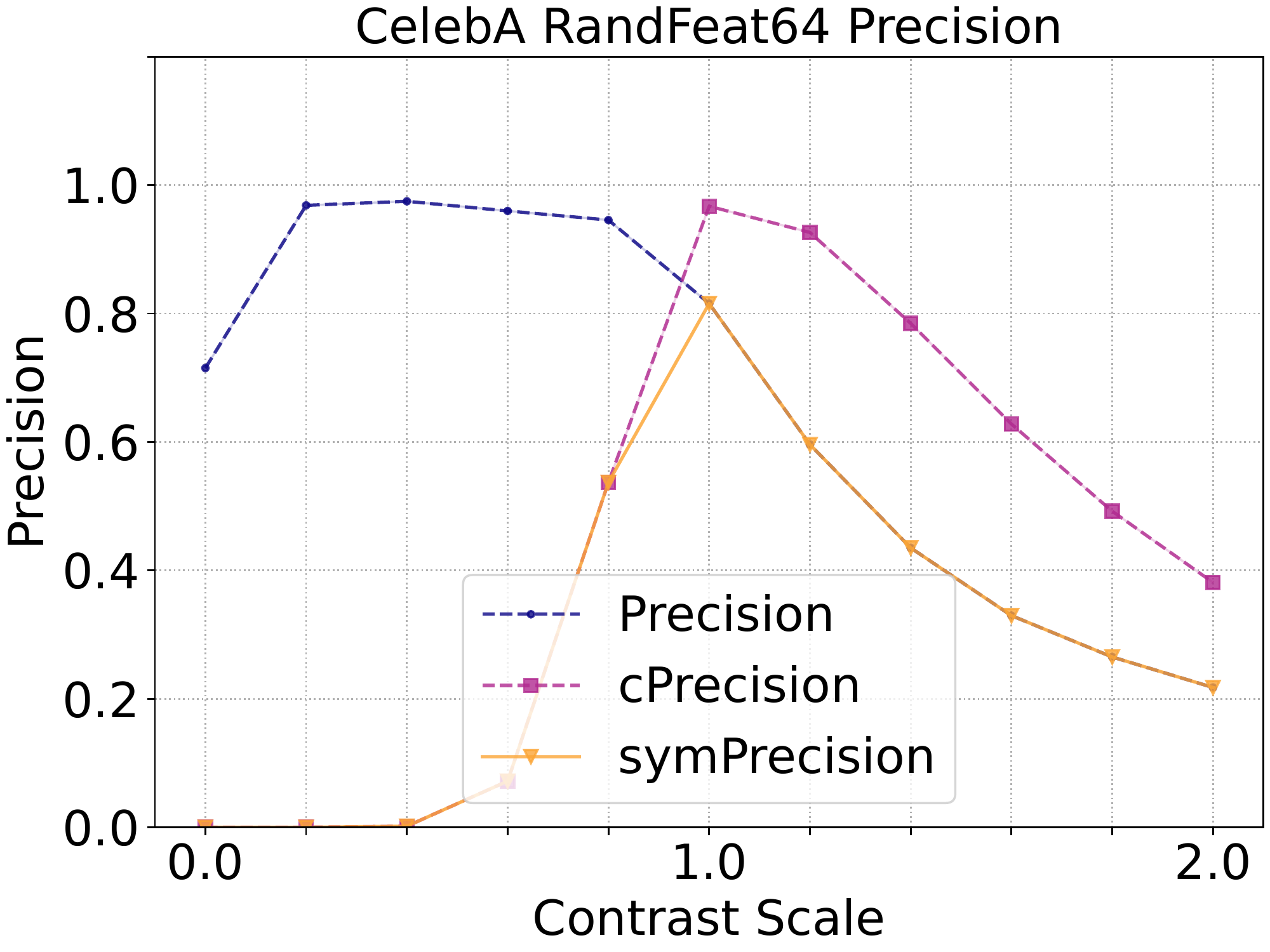}
        \label{fig:contrast_embed_rand64_celeba_p}}
    \subfloat[Diversity (CelebA)]{
        \centering
        \includegraphics[trim=64 0 0 25, clip, width=0.235\textwidth]{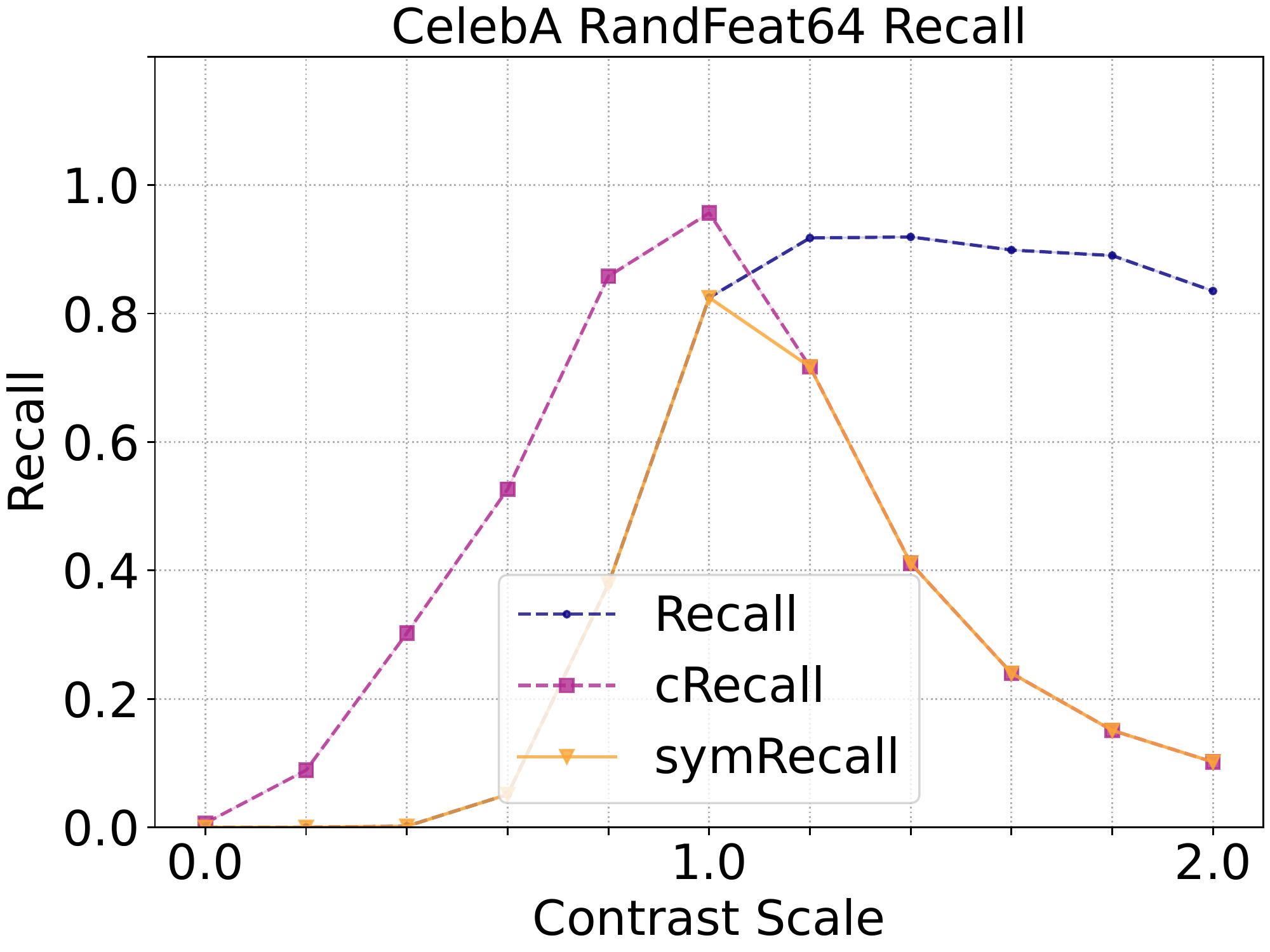}
        \label{fig:contrast_embed_rand64_celeba_r}}
        \subfloat[Fidelity (CIFAR10)]{
        \centering
        \includegraphics[trim=64 0 0 25, clip, width=0.235\textwidth]{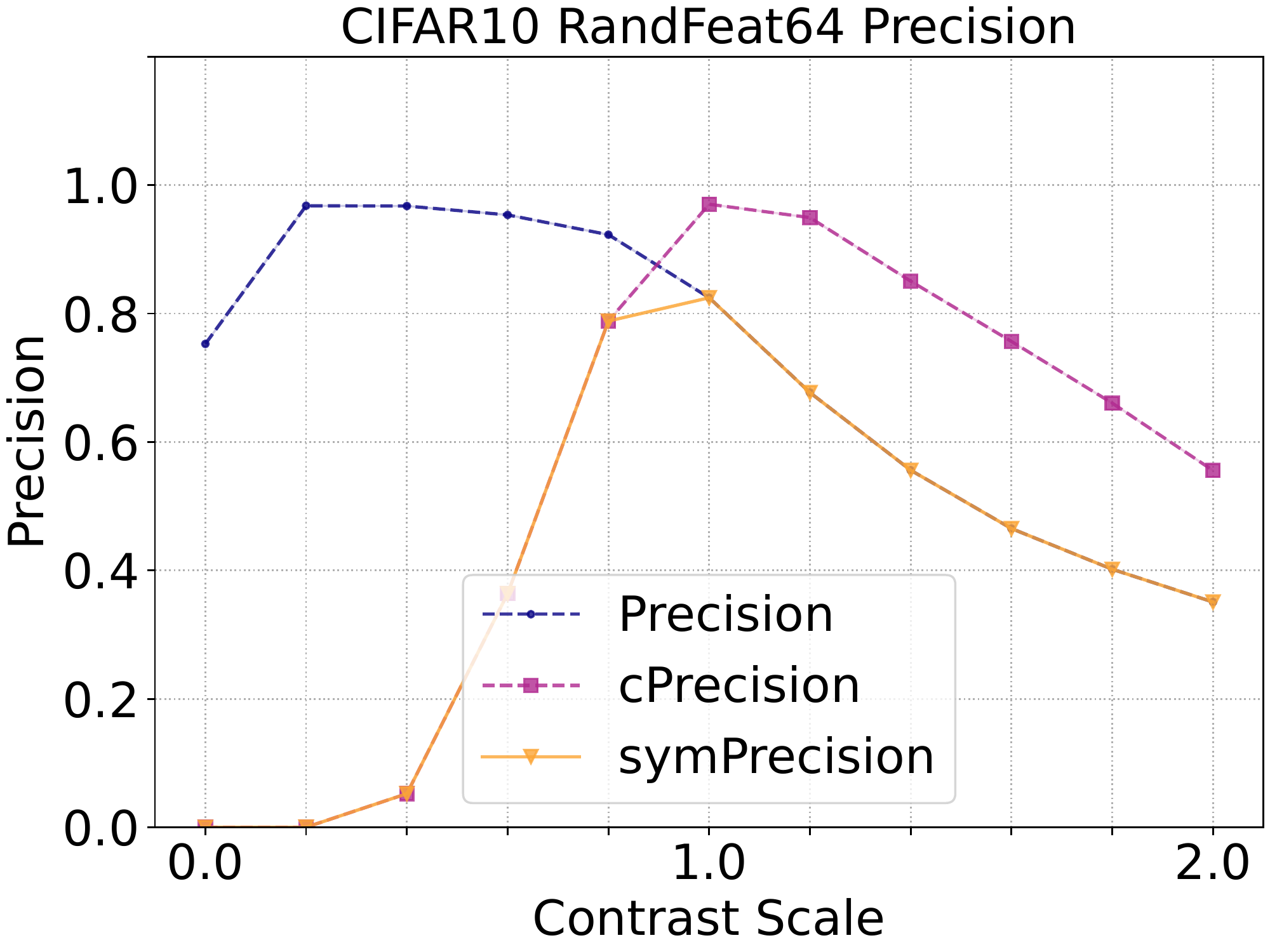}
        \label{fig:contrast_embed_rand64_cifar_p}}
    \subfloat[Diversity (CIFAR10)]{
        \centering
        \includegraphics[trim=64 0 0 25, clip, width=0.235\textwidth]{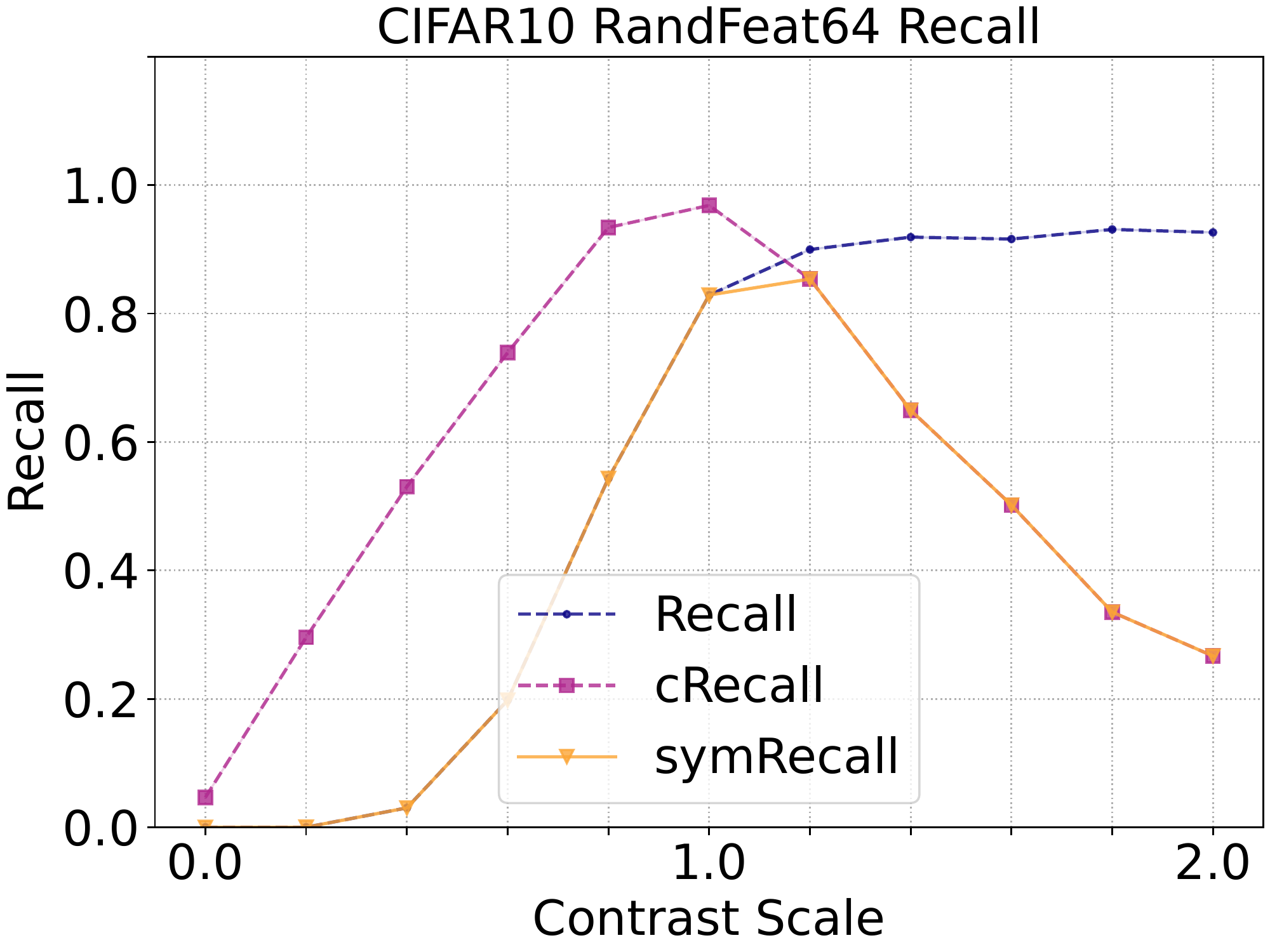}
        \label{fig:contrast_embed_rand64_cifar_r}}
    \caption{Repeating the varying contrast experiment of~\cref{sec:exp_contrast} using the random R64 embedding of images as proposed by~\citet{naeem2020density-coverage} instead of the common pretrained VGG16~\cite{kynkaanniemi2019improved-precision-recall}. The results are consistent with~\cref{fig:contrast_metrics}.}
    \label{fig:contrast_embed_rand64}
\end{figure*}

\end{document}